\documentclass{article}
\PassOptionsToPackage{numbers}{natbib}

\usepackage[main, final]{neurips_2026}

\usepackage[utf8]{inputenc}
\usepackage[T1]{fontenc}
\usepackage{amsmath}
\usepackage{float}
\usepackage{graphicx}
\usepackage{booktabs}
\usepackage{amsfonts}
\usepackage{nicefrac}
\usepackage{microtype}
\usepackage[dvipsnames]{xcolor}
\usepackage{tikz}
\usepackage{pgfplots}
\usepackage{pgf-pie}
\usepackage{subcaption}
\usepackage{makecell}
\usepackage[table]{xcolor}
\usepackage{xspace}
\usepackage{multirow}
\usepackage{pgfplots}
\usepackage{booktabs}
\usepackage{wrapfig}
\usepackage{orcidlink}
\usepackage{soul}
\usepackage{pifont}
\usepackage{hyperref}
\usepackage[framemethod=TikZ]{mdframed}
\usepackage{listings}

\usepackage{duckuments}

\definecolor{laucolor}{RGB}{0, 102, 204}
\definecolor{mariecolor}{RGB}{102, 204, 0}
\definecolor{laurencolor}{RGB}{255, 165, 0}
\definecolor{serenacolor}{RGB}{255, 77, 77}
\definecolor{extracolor}{RGB}{128, 0, 32}

\def\bench{NEST\xspace}
\def\FIND{FIND\xspace}
\newcommand{\crossmark}{\textcolor{red}{\ding{53}}}
\newcommand{\gcheckmark}{\textcolor{green!70!black}{\checkmark}}
\newcommand{\halfmark}{\checkmark\kern-1.1ex\raisebox{.7ex}{\rotatebox[origin=c]{125}{--}}}
\def\totalqs{14,024\xspace}
\def\totalvids{1,208\xspace}

\definecolor{deltapos}{HTML}{0a7f0a}
\definecolor{deltaneg}{HTML}{b22222}
\newcommand{\dpos}[1]{\textbf{\textcolor{deltapos}{#1}}}
\newcommand{\dneg}[1]{\textbf{\textcolor{deltaneg}{#1}}}

\title{Before Words, Beyond Speech: Evaluating Nonverbal Social Reasoning in Early Childhood}

\author{%
  Marie Amale Huynh \textsuperscript{*}\orcidlink{0009-0003-0994-1296} \And
  Laura Bravo-S\'anchez \textsuperscript{*}\orcidlink{0000-0003-3556-3391}
  \AND
  Lauren Klein
  \And
  Nick Haber\orcidlink{0000-0001-8804-7804} \And
  Philip Fisher\orcidlink{0000-0002-5399-7233} \And
  Serena Yeung-Levy\orcidlink{0000-0003-0529-0628} 
  \AND
  Stanford University \\
  \textsuperscript{*}Equal contribution
}

\begin{document}
\addtocontents{toc}{\protect\setcounter{tocdepth}{0}}
\maketitle
\begin{abstract}
Children's earliest years are a period of intense social learning, during which meaning is built through gaze, gesture, and physical contact. Yet, this critical aspect of social interaction remains under-evaluated in multimodal video understanding, as existing benchmarks prioritize the conversation-driven interactions of adults. We address this gap with \bench ($\textbf{N}$aturalistic $\textbf{E}$arly-childhood $\textbf{S}$ocial in$\textbf{T}$eractions), the first child-centric benchmark for nonverbal social reasoning. \bench comprises \totalvids manually annotated clips ($\sim$12s) curated for diversity across naturalistic settings, cultural contexts, and developmental stages. Beyond a lack of data, targeted model improvement in this domain is hindered by coarse evaluation schemas that conflate basic perception with higher-order reasoning. To resolve this, \bench introduces a compositional schema that builds atomic interaction cues into progressive levels of social reasoning: contextual, behavioral, interpersonal, and open-ended descriptions. Extensive analysis of state-of-the-art Vision-Language Models (VLMs) reveals a stark performance collapse: despite strong contextual recognition, models exhibit substantial gaps in decoding precise behavioral and interpersonal cues. We trace these failures to specific biases, including poor consequence grounding and temporal sensitivity. Finally, we ground \bench in real-world applications by evaluating VLMs using domain adaptation strategies. Ultimately, \bench provides a rigorous testbed for advancing socially grounded artificial intelligence in early childhood. Project page: \url{https://laubravo.github.io/nest_website/}.
\end{abstract}

\section{Introduction}
\label{sec:intro}

\begin{figure}[th]
    \centering
    \includegraphics[width=\linewidth]{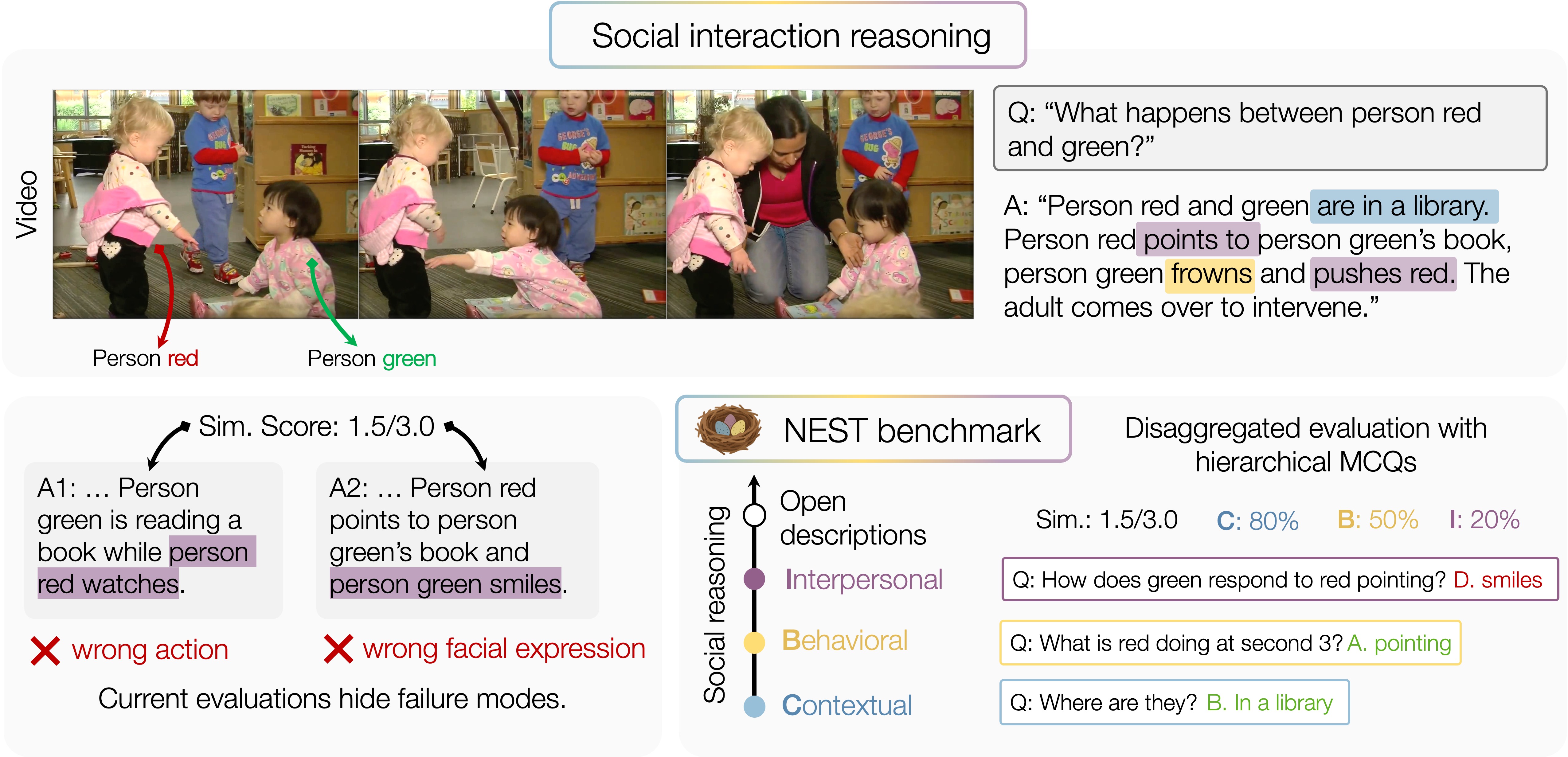}
    \caption{\textbf{Overview of the \bench benchmark.} We introduce the first dataset (\totalvids clips, 14k questions) for evaluating early-childhood non-verbal social reasoning. The example shows how an interaction description can be decomposed into \bench's four-level schema (contextual, behavioral, interpersonal, open) to isolate the fine-grained details that define the exchange. By mapping this schema to task-specific video VQA questions, \bench overcomes the diagnostic ambiguity of standard aggregated metrics and disentangles a model's underlying capabilities.}
    \label{fig:pull}
    \vspace{-8pt}
\end{figure}

Long before children say their first words, they navigate the social world using a sophisticated set of non-verbal tools, including gaze, gesture, and spatial proximity. Coding these behaviors is a core practice in the early childhood field, including child development research \citep{feldman2007parent, roggman2013parenting, ainsworth2015patterns}, interventions to support early development \citep{fisher2016promoting, vik2014tiny, juffer2023promoting}, developmental assessments \citep{esler2015autism, wetherby2002communication}, and classroom quality rating systems \citep{pianta2008classroom, harms1998early}. Despite rapid advances in computer vision and the importance of early childhood for lifelong outcomes, this task remains mostly manual, forcing researchers to trade behavior density against the scale and diversity of populations studied \citep{fukkink2008video, imhof2023improving, black2017early}. Interpreting subtle non-verbal signals in caregiver-child interaction remains a blind spot in modern video understanding. To ensure the scalability promised by recent machine learning advances reaches early childhood, we need tools to track model performance and guide ongoing work.


Existing video social reasoning benchmarks target adult, speech-centric interactions (Table~\ref{tab:related-work}). Large-scale datasets such as Social-IQ 2.0, Social Genome, and R$^3$-VQA are drawn from online media, where social meaning is dominated by dialogue. This contrasts with child-caregiver dynamics, where meaning emerges from a child's sub-second non-verbal cues integrated with a caregiver's one-sided speech \citep{eulau2024behavioral,tamis2014infant,yamamoto2019eye} — exchanges that demand compositional reasoning across interdependent cues. The few benchmarks that include children or non-verbal behavior (e.g., MMDB \citep{rehg2013decoding}, ChildPlay \citep{tafasca2023childplay}, MimeQA \citep{li2025mimeqa}) target narrow assessment protocols or scripted performances rather than diverse, naturalistic child-caregiver interaction. As a result, when models fail, it remains unclear whether errors stem from imprecise temporal grounding, missed social context, or participant-specific challenges.

To address this gap, we introduce \bench (\textbf{N}aturalistic \textbf{E}arly-childhood \textbf{S}ocial in\textbf{T}eractions), the first child-centric benchmark for compositional social reasoning in close, naturalistic video. The clips in \bench capture everyday moments like a child pointing across a dinner table at a cup her mother reaches for, or a caregiver crouching beside a flower to follow a toddler's gaze. \bench comprises \totalvids interaction segments ($\sim 12s$) paired with \totalqs structured questions, curated to maximize diversity across participants, developmental stages, and cultural contexts. Each segment is manually annotated using a unified, temporally grounded schema spanning four complementary axes (Fig.~\ref{fig:pull}): contextual attributes (e.g., setting, activity, age), behavioral cues (atomic actions), interpersonal dynamics (action-reaction events), and an open description of interactions. This dense schema resolves the diagnostic ambiguity inherent in aggregated metrics, decomposes model performance into distinct capabilities, and allows stratified evaluation across metadata. 

Extensive analysis of state-of-the-art video vision-language models (video-VLMs) and specialized models on \bench reveals a stark divergence: despite strong recognition of broad contextual cues, models exhibit substantial gaps in the higher-order reasoning required to decode precise behavioral and interpersonal cues. To diagnose \textit{why}, we identify three biases: models ground consequences less reliably than causes, systematically under-predict interactions in child-child dyads, and miss brief temporal cues. Finally, we examine how VLMs can be deployed in practice by comparing them against specialist social detection models, exploring fine-tuning, and showing that \bench's diagnostic findings persist in a real coaching intervention. Together with \bench, these analyses inform design guidelines for building socially grounded AI systems for early childhood interventions.

Our contributions can be summarized as follows:
\textit{1) The \bench benchmark:} the first child-centric dataset for nonverbal social reasoning in close, naturalistic interactions.
\textit{2) Compositional schema and evaluation:} a hierarchical schema over three reasoning levels — contextual, behavioral, interpersonal — plus an open description of interactions, that successfully disentangles basic perception from higher-order social reasoning, revealing a stark performance collapse in state-of-the-art VLMs.
\textit{3) Comprehensive evaluation of VLM failure modes:} We characterize gaps and test their practical implications on early-childhood interventions. We release annotations and metadata for the ChildPlay and CHILDES subsets, together with evaluation code. Appendix~\ref{app:privacy_release} details the released artifacts, access requirements, and reproducibility limits.

\begin{table*}[th]
\centering
\small
\setlength{\tabcolsep}{4pt}
\renewcommand{\arraystretch}{1.15}
\caption{\textbf{Comparison of \bench with the most directly related social and child-interaction video QA benchmarks.} \bench is the only benchmark that combines child-centric, non-verbal social interaction with naturalistic multi-source video, reported demographics, and a hierarchical reasoning schema applied uniformly at high density across all videos. \crossmark: absent, \gcheckmark: present, \halfmark: partial. \textit{Children}: extent of child–adult interaction content. \textit{Non-verbal}: whether the benchmark targets non-verbal social cues without relying on dialogue. \textit{Uniform schema}: whether the same fixed question schema is applied to every video (vs.\ questions generated per clip); \halfmark{} denotes a recurring task/dimension set that is not applied as a single dense schema. \textit{Q/Vid}: mean questions per video, indicating annotation density over the schema. \textit{Demog.}: whether participant demographics are reported.}
\resizebox{\linewidth}{!}{%
\begin{tabular}{lrrlcclcc}
\toprule
\textbf{Dataset} & \textbf{\# Videos} & \textbf{\# Q} & \textbf{Source} & \textbf{Children} & \textbf{Non-verbal} & \textbf{Uniform schema} & \textbf{Q/Vid} & \textbf{Demog.} \\
\midrule
Social-IQ 2.0~\cite{siq2}    & 1{,}000 & 6{,}000  & YouTube              & \crossmark    & \crossmark      & \crossmark   & 6.0  & \crossmark \\
MMDB$^\dagger$~\cite{rehg2013decoding} & 160   & --       & Lab study            & \gcheckmark   & \gcheckmark     & \gcheckmark  & --   & \gcheckmark \\
Social Genome~\cite{mathur2025social}    & 272     & 1{,}486  & YouTube              & \crossmark    & \crossmark      & \crossmark   & 5.5  & \crossmark \\
R$^3$-VQA~\cite{niu2025r}                & 316     & 5{,}156  & YouTube              & \crossmark    & \crossmark      & \crossmark   & 16.3 & \crossmark \\
SIV-Bench~\cite{kong2025siv}             & 2{,}792 & 8{,}792  & YouTube, TikTok      & \halfmark     & \crossmark      & \halfmark    & 3.1  & \crossmark \\
Motion2Mind~\cite{lee2025mind}    & 4{,}730 & --       & TV, movies           & \crossmark    & \gcheckmark       & \halfmark    & --   & \crossmark \\
MimeQA~\cite{li2025mimeqa}               & 101     & 806      & YouTube      & \crossmark    & \gcheckmark      & \gcheckmark  & 8.0  & \crossmark \\
\midrule
\textbf{NEST (ours)}  & \textbf{\totalvids} & \textbf{\totalqs} & \textbf{Home, lab, classroom, YouTube} & \textbf{\gcheckmark} & \textbf{\gcheckmark} & \textbf{\gcheckmark} & \textbf{11.6} & \textbf{\gcheckmark} \\
\bottomrule
\end{tabular}%
}
\\[2pt]
{\footnotesize $^\dagger$MMDB provides a uniform frame-level coding scheme rather than a question-answering schema.}
\label{tab:related-work}
\vspace{-5pt}
\end{table*}
\section{Related Work}
\label{sec:rw}




\noindent\textbf{Social video question answering.} Prior work \citep{zadeh2019social, siq2, guo2023desiq} established video question answering (VQA) over YouTube as the standard setup for evaluating social reasoning, focused on adult intentions and causal inference. Recent benchmarks add structural complexity through grounded reasoning chains \citep{mathur2025social}, mental-state inference \citep{niu2025r}, and relationship-aware evaluation \citep{kong2025siv}. These build on a broader line of causal VQA \citep{li2023intentqa,li2022representation,parmar2024causalchaos,zhou2026humanvbench}, whose causal question taxonomies (e.g., causal-why, causal-how) we adapt to the social-interaction setting. However, as Table~\ref{tab:related-work} summarizes, these benchmarks remain predominantly adult-centric and drawn from scripted or online media, where interactions are narratively framed and linguistically explicit. They also organize evaluation as sequential reasoning chains rather than across the multiple, complementary dimensions of a single interaction. \bench instead targets child-caregiver exchanges, where social meaning depends on the nonverbal coordination of these dimensions.

\noindent\textbf{Non-verbal benchmarks and structured action representations.} Two structural paradigms dominate non-verbal video understanding: task-decomposed evaluation schemas and graph-based action representations. MimeQA~\citep{li2025mimeqa} poses hierarchical questions over mime performances where dialogue is absent by construction, while Motion2Mind~\citep{lee2025mind} stages non-verbal Theory-of-Mind as detection, knowledge, and explanation over scripted-media clips. Action Genome~\citep{Ji_2020_CVPR} represents action structure as scene graphs of objects and their spatio-temporal relations, capturing the visual scaffolding of events but not the 
higher-order interaction dynamics they support. In contrast, \bench locates the diagnostic signal not in mental-state depth or scene structure but in composition: a child's gaze, a point, and a partner's answering look are each individually identifiable, but the social act—joint attention—exists only in their combination. Mirroring how developmental coding reads lower-order cues as assembling into social acts, \bench's contextual, behavioral, and interpersonal layers (Sec.~\ref{sec:deconstructing}) localize where in this assembly models break. 

\noindent\textbf{Non-verbal perception models.} A complementary line of work models social behavior using static or signal-level cues, treating nonverbal communication primarily as a perceptual-detection problem. Gaze localization benchmarks \citep{recasens2017following,chong2020detecting,tafasca2023childplay} frame social perception as identifying where a person is looking, while \citep{marin2019laeo,fan2018inferring,fan2019understanding,tafasca2024toward,vuillecard2026omnihead,cao2026gaze} extend this to mutual gaze, shared attention, and atomic gaze events such as joint attention. Other specialist models address adjacent capabilities: estimating demographics such as age \citep{mivolo2023}, while \citep{wei2024nonverbal} integrates gaze with other proxemics and haptics like gesture, posture, expression, and touch. We benchmark these specialist models against generalist VLMs on \bench to test whether general-purpose reasoning can substitute for purpose-built detection.

\noindent\textbf{Child-centric vision datasets.} A growing body of work targets child-centered visual learning. The Multimodal Dyadic Behavior dataset (MMDB) \citep{rehg2013decoding} pioneered structured assessment of child–adult interaction with frame-level behavioral annotations, but operates at single-lab scale and predates current VLM evaluation. ChildPlay \citep{tafasca2023childplay} provides naturalistic parent–child videos with 2D gaze annotations, but lacks shared-attention labels or links to higher-level social constructs. A separate line of work targets representation learning from infant-perspective video: BabyView \citep{long2024babyview} and SAYCam \citep{sullivan2021saycam} provide large-scale egocentric streams, and BabyVLM~\citep{wang2025babyvlm,wang2025babyvlm2} extends this toward developmentally grounded pretraining and benchmarking  of vision foundation models. Together, these resources target perceptual learning, isolated signals, or single protocol behavioral coding. \bench complements them by spanning multiple corpora, cultural contexts, and developmental stages.

\section{The \bench dataset}
\label{sec:dataset}
To bridge the gap in social reasoning, we introduce \bench: the first compositional benchmark designed to capture the nuanced, non-verbal social dynamics of early childhood. This section outlines the dataset's conceptual ideation (\ref{sec:deconstructing}), construction (\ref{sec:video-selection}) and description (\ref{sec:nest_statistics}).

\begin{figure*}[t]
\centering
\includegraphics[width=\linewidth]{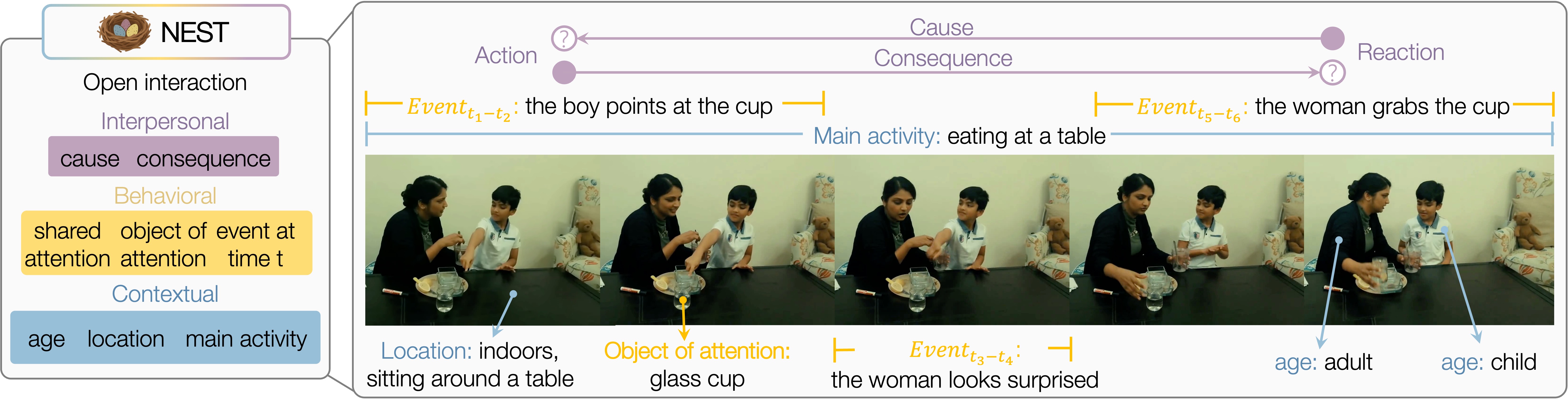}
\caption{\textbf{(Left)} \bench's hierarchical task ontology for non-verbal social cues. \textbf{(Right)} Example of dense per-task VQA annotations (avg. $\sim 12$ per clip) used to decompose model performance across the \bench schema.}
\label{fig:tasks}
\vspace{-2pt}
\end{figure*}

\subsection{Deconstructing social understanding}\label{sec:deconstructing}


To understand where state-of-the-art VLMs struggle with social reasoning, we cannot rely solely on overall judgments of broad concepts such as engagement or dyadic synchrony. These measures are debated even among experts, and a single accuracy score does not reveal which specific signal the model missed. Instead, diagnosis must break down evaluation into tasks that can be assessed independently \citep{rehg2013decoding, mathur2024advancing, niu2025r}. But which basic signals should guide this breakdown? We look to developmental psychology, where years of behavioral coding have identified a small set of social signals—gaze, posture, and proximity—that combine into more complex dyad-level concepts \citep{feldman2012parenting, perlman2016systematic, esler2015autism}. Rather than adopting a single schema, we design \bench around this principle with domain experts in early childhood interventions from the \FIND program \cite{fisher2016promoting}.

Concretely, \bench evaluates social reasoning at four levels of granularity (see Fig.~\ref{fig:tasks}). Contextual questions probe basic scene properties (age, setting, main activity). Behavioral questions test temporal signals between specific individuals (mutual gaze, shared attention, actions in a time frame). Interpersonal questions address the causal and reactive structure of a social exchange (what prompted a given reaction, how someone responded to an action directed at them). An open task asks for a free-form description of the interaction, spanning all three levels.

\subsection{Building \bench}\label{sec:video-selection}
\begin{figure}[th]
    \centering
    \includegraphics[width=\linewidth]{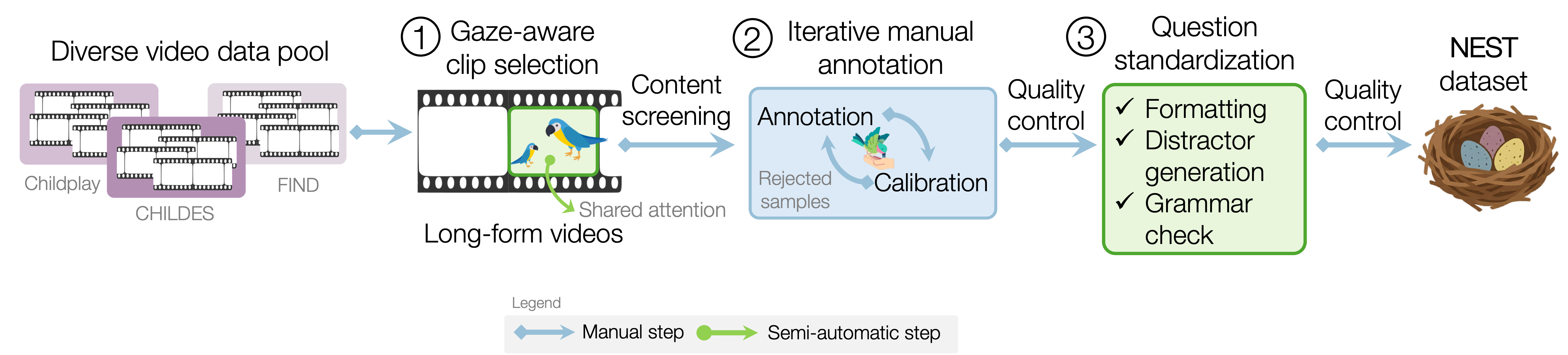}
    \caption{\textbf{Annotation procedure.} From a large pool of video datasets featuring early childhood interactions, we select videos that support diversity across settings, countries, activities, ages, etc. We emphasize manual annotation (blue arrows) and quality control with semi-automatic steps (green) to arrive at \bench.}
    \label{fig:process}
\end{figure}


NEST consists of short clips of naturalistic caregiver-child interaction, each capturing the brief, nonverbal exchanges that developmental coding relies on. Our curation strategy is guided by two goals: maximizing the density of social behavior within each clip, and maximizing diversity across settings, cultures, and developmental stages. Rather than stage new recordings, we draw on existing video, which lets us capture spontaneous interaction at a scale and breadth that controlled collection cannot reach, while spanning the range of real-world contexts in which these behaviors naturally occur. Concretely, we build NEST in three main stages (Fig.~\ref{fig:process}): (1) compiling a diverse video pool from multiple corpora and applying gaze-aware filtering to select clips with high social engagement, (2) iterative manual annotation of behavioral labels, and (3) conversion of these labels into multiple-choice questions. We describe each stage below.

\noindent\textbf{Compiling clips with high social engagement.} We source videos from three corpora of naturalistic adult-child interaction across cultural contexts, developmental stages, and recording conditions. Two are publicly available: ChildPlay~\citep{tafasca2023childplay}, unconstrained interactions sourced from YouTube; and CHILDES~\citep{macwhinney2000childes}, a longitudinal corpus of recordings collected across many studies, languages, and cultures. The third corpus is private and drawn from a real-world intervention: caregiver–child and classroom videos collected through \textit{Filming Interactions to Nurture Development} (FIND)~\citep{fisher2016promoting}, a positive-reinforcement coaching program that promotes supportive interactions between children and caregivers. Annotation and analysis were conducted under Stanford IRB-approved protocol 66624; consent, access conditions, and privacy safeguards
are detailed in Appendix~\ref{app:privacy}. From the long-form videos of the three corpora, we extract 12-second windows centered on frames with high social engagement.  We use gaze as the primary signal: it is visually grounded, unrecoverable from dialogue alone, and drives parent–child coordination from infancy \citep{feldman2007parent, 
fisher2016promoting}. Specifically, we restrict \bench to clips containing mutual gaze or shared focus on a common object, identified using estimated gaze points from \citep{tafasca2024sharingan} ($dist \leq 10$ pixels). To evaluate models on clips with limited social engagement, we also include 20\% of clips that do not meet these 
criteria. See Section~\ref{supp:segment_selection} for details.


\begin{figure*}[t]
\includegraphics[width=\linewidth]{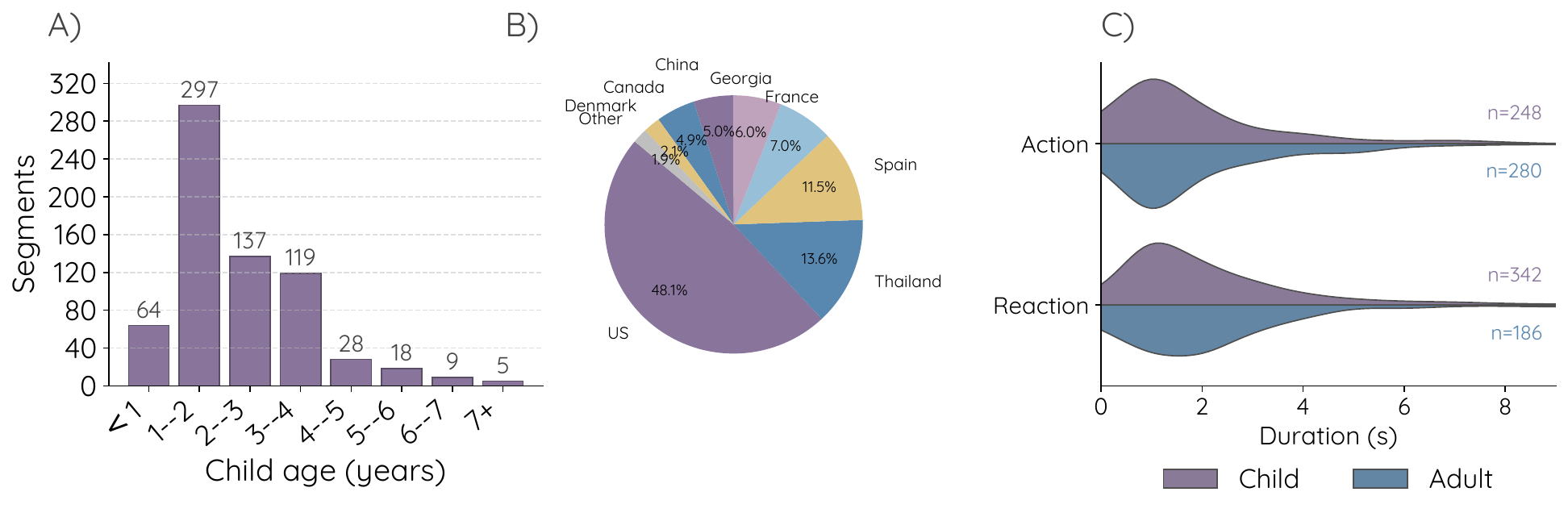}
\caption{\textbf{\bench Statistics.} \textbf{A)} Child age distribution (677/\totalvids clips with reported age). \textbf{B)} Recording country distribution (801/\totalvids with reported origin). \textbf{C)} Duration distributions for actions (a behavior directed at another participant) and reactions (the responding behavior), covering both adult$\rightarrow$child and child$\rightarrow$adult pairs.}
\label{fig:stats}
\end{figure*}

\noindent\textbf{Iterative manual annotation.}
Annotators labeled each clip using only the visual stream (no audio), ensuring the labels relied on observable cues. They first identified individuals and validated gaze events used for selection, then labeled contextual attributes, and finally described behavioral cues and interpersonal exchanges  — selecting up to two 
action-reaction pairs between participants, with temporal grounding of 
atomic actions. Annotators concluded with an open-ended description of 
the interaction. To ensure quality, 20\% of ChildPlay was annotated by 
all annotators and discussed before independent work proceeded; every 
annotation was later reviewed by a second annotator, with rejected 
labels reassigned. Full annotation protocols, exclusion criteria, and 
inter-rater agreement statistics are provided in Section~\ref{sec:annotation_protocol}.
 
\noindent\textbf{Generating multiple choice questions.}
To support quantitative evaluation, natural-language labels are reformatted into multiple-choice questions (MCQs). Informative MCQs require distractors that are plausible yet incorrect, as naïvely constructed distractors can artificially inflate performance by failing to meaningfully challenge models. To address this, we employ AutoConverter \citep{zhang2025automated} to generate distractors that are both visually grounded and linguistically consistent. For each MCQ, we construct a compact context $\mathcal{C}$ consisting of (i) five frames sampled from the corresponding video clip, (ii) the question text, and (iii) relevant human annotations, including the ground-truth answer and additional contextual labels. This context was provided to \textit{Qwen3-VL-32B-Instruct} \citep{qwen3technicalreport}, which generated candidate distractors conditioned jointly on visual evidence and linguistic context. By grounding distractor generation in both modalities, the resulting options remain semantically plausible while being clearly distinguishable from the correct answer.

\begin{table}[t]
  \caption{\textbf{Model performance on the public subset of \bench (ChildPlay and CHILDES)}. Open interactions are measured in similarity to ground truth (0-3) and recall (\%), MCQs
  in accuracy (\%) and balanced accuracy (\%) when categorical. \textbf{Best} model and \underline{second best} model. Darker color
  indicates higher performance.}
  \label{tab:main_results}
  \resizebox{\linewidth}{!}{%
  \begin{tabular}{lcccccccccccccccc}
  \toprule
  \multicolumn{1}{c}{\textbf{Model}} &  & \multicolumn{4}{c}{\textbf{Contextual}} & \multicolumn{4}{c}{\textbf{Behavioral}} &
  \multicolumn{4}{c}{\textbf{Interpersonal}} & \multicolumn{1}{c}{\textbf{All}} & \multicolumn{2}{c}{\textbf{Open Inter.}} \\
  \cmidrule(lr){3-6} \cmidrule(lr){7-10} \cmidrule(lr){11-14} \cmidrule(lr){15-15} \cmidrule(lr){16-17}
   & Size & \shortstack{age} & loc. & \shortstack{main\\act.} & Avg & \shortstack{object\\attn.} & \shortstack{shared\\attn.} &
  \shortstack{event\\at t} & Avg & \shortstack{react.} & cause & conseq. & Avg & Avg & \shortstack{sim.\\(0-3)} & recall \\
  \midrule
  Human & -- & 87.9 & 90.6 & 67.8 & 82.1 & 88.7 & 93.2 & 95.3 & 92.4 & 79.3 & 86.1 & 94.0 & 86.5 & 87.5 & -- & -- \\
  \midrule
  \multicolumn{17}{l}{\textit{Open-source}} \\
  \midrule
  LLaVA-NeXT-Video & 7B & \cellcolor[rgb]{0.841,0.696,0.770} 44.5 & \cellcolor[rgb]{0.742,0.619,0.724} 64.7 &
  \cellcolor[rgb]{0.832,0.687,0.766} 46.3 & \cellcolor[rgb]{0.807,0.665,0.754} 51.8 & \cellcolor[rgb]{0.809,0.667,0.756} 51.5 &
  \cellcolor[rgb]{0.867,0.724,0.783} 37.7 & \cellcolor[rgb]{0.901,0.769,0.804} 27.5 & \cellcolor[rgb]{0.863,0.719,0.781} 38.9 &
  \cellcolor[rgb]{0.841,0.696,0.770} 44.1 & \cellcolor[rgb]{0.852,0.707,0.776} 41.0 & \cellcolor[rgb]{0.863,0.719,0.781} 38.7 &
  \cellcolor[rgb]{0.852,0.707,0.776} 41.3 & \cellcolor[rgb]{0.841,0.696,0.770} 44.0 & \cellcolor[rgb]{0.899,0.766,0.803} 0.8 &
  \cellcolor[rgb]{0.949,0.866,0.860} 7.3 \\
  LLaVA-OV & 7B & \cellcolor[rgb]{0.718,0.605,0.712} 69.4 & \cellcolor[rgb]{0.642,0.564,0.665} 82.6 & \cellcolor[rgb]{0.733,0.614,0.720}
  66.1 & \cellcolor[rgb]{0.696,0.593,0.700} 72.7 & \cellcolor[rgb]{0.721,0.607,0.713} 68.4 & \cellcolor[rgb]{0.814,0.672,0.758} 50.2 &
  \cellcolor[rgb]{0.863,0.719,0.781} 38.4 & \cellcolor[rgb]{0.804,0.663,0.753} 52.3 & \cellcolor[rgb]{0.684,0.586,0.692} \underline{75.3} &
   \cellcolor[rgb]{0.814,0.672,0.758} 50.2 & \cellcolor[rgb]{0.875,0.734,0.788} 35.3 & \cellcolor[rgb]{0.799,0.659,0.751} 53.6 &
  \cellcolor[rgb]{0.769,0.637,0.737} 59.6 & \cellcolor[rgb]{0.881,0.741,0.791} 1.0 & \cellcolor[rgb]{0.935,0.833,0.839} 14.1 \\
  openPangu-VL & 7B & \cellcolor[rgb]{0.757,0.629,0.732} 61.8 & \cellcolor[rgb]{0.623,0.555,0.652} 85.8 &
  \cellcolor[rgb]{0.721,0.607,0.713} 68.7 & \cellcolor[rgb]{0.703,0.596,0.703} 72.1 & \cellcolor[rgb]{0.671,0.579,0.684} 77.5 &
  \cellcolor[rgb]{0.822,0.678,0.761} 48.8 & \cellcolor[rgb]{0.799,0.659,0.751} 53.8 & \cellcolor[rgb]{0.769,0.637,0.737} 60.0 &
  \cellcolor[rgb]{0.727,0.610,0.717} 67.9 & \cellcolor[rgb]{0.802,0.661,0.752} 53.4 & \cellcolor[rgb]{0.802,0.661,0.752} 52.8 &
  \cellcolor[rgb]{0.777,0.643,0.741} 58.1 & \cellcolor[rgb]{0.748,0.623,0.727} 63.4 & \cellcolor[rgb]{0.910,0.785,0.812} 0.7 &
  \cellcolor[rgb]{0.947,0.860,0.857} 8.2 \\
  InternVL3.5 & 4B & \cellcolor[rgb]{0.910,0.785,0.812} 24.2 & \cellcolor[rgb]{0.724,0.608,0.715} 68.2 & \cellcolor[rgb]{0.832,0.687,0.766}
   46.4 & \cellcolor[rgb]{0.832,0.687,0.766} 46.3 & \cellcolor[rgb]{0.736,0.616,0.721} 65.7 & \cellcolor[rgb]{0.896,0.761,0.800} 29.2 &
  \cellcolor[rgb]{0.913,0.790,0.814} 23.0 & \cellcolor[rgb]{0.861,0.717,0.780} 39.3 & \cellcolor[rgb]{0.841,0.696,0.770} 44.0 &
  \cellcolor[rgb]{0.857,0.712,0.778} 40.2 & \cellcolor[rgb]{0.871,0.729,0.785} 36.2 & \cellcolor[rgb]{0.857,0.712,0.778} 40.1 &
  \cellcolor[rgb]{0.850,0.705,0.775} 41.9 & \cellcolor[rgb]{0.896,0.761,0.800} 0.9 & \cellcolor[rgb]{0.945,0.855,0.853} 10.0 \\
  InternVL3.5 & 8B & \cellcolor[rgb]{0.887,0.749,0.794} 31.9 & \cellcolor[rgb]{0.696,0.593,0.700} 72.9 & \cellcolor[rgb]{0.809,0.667,0.756}
   51.3 & \cellcolor[rgb]{0.807,0.665,0.754} 52.0 & \cellcolor[rgb]{0.754,0.627,0.730} 62.3 & \cellcolor[rgb]{0.829,0.685,0.765} 46.9 &
  \cellcolor[rgb]{0.904,0.774,0.807} 26.4 & \cellcolor[rgb]{0.837,0.691,0.768} 45.2 & \cellcolor[rgb]{0.871,0.729,0.785} 36.1 &
  \cellcolor[rgb]{0.829,0.685,0.765} 46.6 & \cellcolor[rgb]{0.857,0.712,0.778} 40.5 & \cellcolor[rgb]{0.852,0.707,0.776} 41.1 &
  \cellcolor[rgb]{0.832,0.687,0.766} 46.1 & \cellcolor[rgb]{0.890,0.754,0.797} 0.9 & \cellcolor[rgb]{0.942,0.849,0.849} 10.9 \\
  InternVL3.5 & 38B & \cellcolor[rgb]{0.814,0.672,0.758} 50.2 & \cellcolor[rgb]{0.652,0.569,0.672} 80.7 &
  \cellcolor[rgb]{0.739,0.618,0.723} 65.1 & \cellcolor[rgb]{0.739,0.618,0.723} 65.4 & \cellcolor[rgb]{0.703,0.596,0.703} 72.0 &
  \cellcolor[rgb]{0.796,0.657,0.750} 54.2 & \cellcolor[rgb]{0.850,0.705,0.775} 42.1 & \cellcolor[rgb]{0.788,0.651,0.746} 56.1 &
  \cellcolor[rgb]{0.829,0.685,0.765} 46.5 & \cellcolor[rgb]{0.769,0.637,0.737} 59.5 & \cellcolor[rgb]{0.829,0.685,0.765} 46.5 &
  \cellcolor[rgb]{0.812,0.669,0.757} 50.9 & \cellcolor[rgb]{0.780,0.645,0.742} 57.4 & \cellcolor[rgb]{0.910,0.785,0.812} 0.7 &
  \cellcolor[rgb]{0.949,0.866,0.860} 7.7 \\
  MiniCPM-V 4.5 & 8B & \cellcolor[rgb]{0.658,0.572,0.676} 80.0 & \cellcolor[rgb]{0.592,0.539,0.626} 91.5 &
  \cellcolor[rgb]{0.677,0.582,0.688} 76.4 & \cellcolor[rgb]{0.642,0.564,0.665} 82.6 & \cellcolor[rgb]{0.655,0.571,0.674} 80.3 &
  \cellcolor[rgb]{0.785,0.649,0.745} 56.4 & \cellcolor[rgb]{0.809,0.667,0.756} 51.2 & \cellcolor[rgb]{0.754,0.627,0.730} 62.6 &
  \cellcolor[rgb]{0.760,0.631,0.733} 61.5 & \cellcolor[rgb]{0.760,0.631,0.733} 61.7 & \cellcolor[rgb]{0.817,0.674,0.759} 49.9 &
  \cellcolor[rgb]{0.780,0.645,0.742} 57.7 & \cellcolor[rgb]{0.727,0.610,0.717} 67.6 & \cellcolor[rgb]{0.881,0.741,0.791} 1.0 &
  \cellcolor[rgb]{0.940,0.844,0.845} 11.8 \\
  Molmo2 & 8B & \cellcolor[rgb]{0.671,0.579,0.684} 77.7 & \cellcolor[rgb]{0.608,0.547,0.639} 88.9 & \cellcolor[rgb]{0.674,0.581,0.686} 77.1
   & \cellcolor[rgb]{0.652,0.569,0.672} 81.2 & \cellcolor[rgb]{0.646,0.566,0.668} 81.7 & \cellcolor[rgb]{0.774,0.641,0.740} 58.9 &
  \cellcolor[rgb]{0.769,0.637,0.737} 60.0 & \cellcolor[rgb]{0.730,0.612,0.718} 66.9 & \cellcolor[rgb]{0.763,0.633,0.734} 60.7 &
  \cellcolor[rgb]{0.757,0.629,0.732} 61.8 & \cellcolor[rgb]{0.771,0.639,0.738} \underline{59.3} & \cellcolor[rgb]{0.763,0.633,0.734} 60.6 &
   \cellcolor[rgb]{0.715,0.603,0.710} 69.6 & \cellcolor[rgb]{0.885,0.746,0.793} 1.0 & \cellcolor[rgb]{0.935,0.833,0.839} 14.3 \\
  Keye-VL 1.5 & 8B & \cellcolor[rgb]{0.829,0.685,0.765} 47.2 & \cellcolor[rgb]{0.642,0.564,0.665} 82.5 & \cellcolor[rgb]{0.804,0.663,0.753}
   52.6 & \cellcolor[rgb]{0.763,0.633,0.734} 60.8 & \cellcolor[rgb]{0.754,0.627,0.730} 62.5 & \cellcolor[rgb]{0.788,0.651,0.746} 56.2 &
  \cellcolor[rgb]{0.902,0.771,0.805} 27.3 & \cellcolor[rgb]{0.822,0.678,0.761} 48.7 & \cellcolor[rgb]{0.703,0.596,0.703} 72.2 &
  \cellcolor[rgb]{0.802,0.661,0.752} 52.8 & \cellcolor[rgb]{0.859,0.714,0.779} 39.7 & \cellcolor[rgb]{0.794,0.655,0.749} 54.9 &
  \cellcolor[rgb]{0.794,0.655,0.749} 54.8 & \cellcolor[rgb]{0.852,0.707,0.776} \textbf{1.3} & \cellcolor[rgb]{0.928,0.816,0.829} 17.3 \\
  Qwen3-VL & 4B & \cellcolor[rgb]{0.890,0.754,0.797} 31.0 & \cellcolor[rgb]{0.696,0.593,0.700} 73.2 & \cellcolor[rgb]{0.754,0.627,0.730}
  62.2 & \cellcolor[rgb]{0.788,0.651,0.746} 55.5 & \cellcolor[rgb]{0.721,0.607,0.713} 68.6 & \cellcolor[rgb]{0.814,0.672,0.758} 50.4 &
  \cellcolor[rgb]{0.848,0.703,0.773} 42.3 & \cellcolor[rgb]{0.799,0.659,0.751} 53.8 & \cellcolor[rgb]{0.718,0.605,0.712} 68.9 &
  \cellcolor[rgb]{0.794,0.655,0.749} 54.6 & \cellcolor[rgb]{0.822,0.678,0.761} 48.6 & \cellcolor[rgb]{0.780,0.645,0.742} 57.4 &
  \cellcolor[rgb]{0.788,0.651,0.746} 55.5 & \cellcolor[rgb]{0.863,0.719,0.781} 1.2 & \cellcolor[rgb]{0.928,0.816,0.829} 17.4 \\
  Qwen3-VL & 8B & \cellcolor[rgb]{0.794,0.655,0.749} 54.7 & \cellcolor[rgb]{0.642,0.564,0.665} 82.5 & \cellcolor[rgb]{0.739,0.618,0.723}
  64.9 & \cellcolor[rgb]{0.727,0.610,0.717} 67.4 & \cellcolor[rgb]{0.693,0.591,0.698} 73.6 & \cellcolor[rgb]{0.780,0.645,0.742} 57.4 &
  \cellcolor[rgb]{0.807,0.665,0.754} 51.9 & \cellcolor[rgb]{0.760,0.631,0.733} 60.9 & \cellcolor[rgb]{0.852,0.707,0.776} 41.6 &
  \cellcolor[rgb]{0.794,0.655,0.749} 54.6 & \cellcolor[rgb]{0.812,0.669,0.757} 51.0 & \cellcolor[rgb]{0.820,0.676,0.760} 49.1 &
  \cellcolor[rgb]{0.771,0.639,0.738} 59.1 & \cellcolor[rgb]{0.850,0.705,0.775} \textbf{1.3} & \cellcolor[rgb]{0.924,0.808,0.824} 19.1 \\
  Qwen3-VL & 32B & \cellcolor[rgb]{0.709,0.600,0.707} 70.8 & \cellcolor[rgb]{0.620,0.553,0.649} 86.3 & \cellcolor[rgb]{0.690,0.589,0.696}
  74.3 & \cellcolor[rgb]{0.674,0.581,0.686} 77.1 & \cellcolor[rgb]{0.655,0.571,0.674} 80.3 & \cellcolor[rgb]{0.788,0.651,0.746} 55.8 &
  \cellcolor[rgb]{0.760,0.631,0.733} 61.2 & \cellcolor[rgb]{0.736,0.616,0.721} 65.8 & \cellcolor[rgb]{0.829,0.685,0.765} 46.9 &
  \cellcolor[rgb]{0.730,0.612,0.718} 67.1 & \cellcolor[rgb]{0.791,0.653,0.747} 55.3 & \cellcolor[rgb]{0.785,0.649,0.745} 56.4 &
  \cellcolor[rgb]{0.733,0.614,0.720} 66.4 & \cellcolor[rgb]{0.892,0.756,0.798} 0.9 & \cellcolor[rgb]{0.918,0.798,0.819} 21.2 \\
  Qwen3.5 & 4B & \cellcolor[rgb]{0.627,0.556,0.654} 85.5 & \cellcolor[rgb]{0.586,0.536,0.621} \underline{92.9} &
  \cellcolor[rgb]{0.668,0.577,0.682} 78.2 & \cellcolor[rgb]{0.627,0.556,0.654} \underline{85.5} & \cellcolor[rgb]{0.627,0.556,0.654} 85.4 &
   \cellcolor[rgb]{0.788,0.651,0.746} 56.0 & \cellcolor[rgb]{0.760,0.631,0.733} 61.5 & \cellcolor[rgb]{0.727,0.610,0.717} 67.6 &
  \cellcolor[rgb]{0.791,0.653,0.747} 55.4 & \cellcolor[rgb]{0.736,0.616,0.721} 65.8 & \cellcolor[rgb]{0.829,0.685,0.765} 47.1 &
  \cellcolor[rgb]{0.788,0.651,0.746} 56.1 & \cellcolor[rgb]{0.715,0.603,0.710} 69.7 & \cellcolor[rgb]{0.883,0.744,0.792} 1.0 &
  \cellcolor[rgb]{0.945,0.855,0.853} 9.9 \\
  Qwen3.5 & 9B & \cellcolor[rgb]{0.623,0.555,0.652} 85.8 & \cellcolor[rgb]{0.586,0.536,0.621} 92.2 & \cellcolor[rgb]{0.671,0.579,0.684}
  77.7 & \cellcolor[rgb]{0.627,0.556,0.654} 85.2 & \cellcolor[rgb]{0.623,0.555,0.652} \underline{85.8} & \cellcolor[rgb]{0.763,0.633,0.734}
   60.6 & \cellcolor[rgb]{0.760,0.631,0.733} 61.7 & \cellcolor[rgb]{0.718,0.605,0.712} \underline{69.4} &
  \cellcolor[rgb]{0.681,0.584,0.690} \textbf{75.6} & \cellcolor[rgb]{0.715,0.603,0.710} 69.7 & \cellcolor[rgb]{0.788,0.651,0.746} 55.6 &
  \cellcolor[rgb]{0.730,0.612,0.718} \underline{67.0} & \cellcolor[rgb]{0.690,0.589,0.696} \underline{73.9} &
  \cellcolor[rgb]{0.877,0.736,0.789} 1.0 & \cellcolor[rgb]{0.942,0.849,0.849} 10.8 \\
  Qwen3.5 & 27B & \cellcolor[rgb]{0.614,0.550,0.644} \textbf{87.3} & \cellcolor[rgb]{0.573,0.529,0.610} \textbf{94.9} &
  \cellcolor[rgb]{0.636,0.561,0.661} \textbf{83.4} & \cellcolor[rgb]{0.608,0.547,0.639} \textbf{88.5} & \cellcolor[rgb]{0.592,0.539,0.626}
  \textbf{91.7} & \cellcolor[rgb]{0.757,0.629,0.732} \textbf{62.0} & \cellcolor[rgb]{0.727,0.610,0.717} \textbf{67.3} &
  \cellcolor[rgb]{0.693,0.591,0.698} \textbf{73.7} & \cellcolor[rgb]{0.754,0.627,0.730} 62.4 & \cellcolor[rgb]{0.674,0.581,0.686}
  \textbf{76.7} & \cellcolor[rgb]{0.774,0.641,0.740} 58.3 & \cellcolor[rgb]{0.736,0.616,0.721} 65.8 & \cellcolor[rgb]{0.681,0.584,0.690}
  \textbf{76.0} & \cellcolor[rgb]{0.850,0.705,0.775} \textbf{1.3} & \cellcolor[rgb]{0.916,0.795,0.817} \underline{21.9} \\
  \midrule
  \multicolumn{17}{l}{\textit{Closed-source (API)}} \\
  \midrule
  Gemini 3.1 Flash Lite & -- & \cellcolor[rgb]{0.620,0.553,0.649} 86.5 & \cellcolor[rgb]{0.620,0.553,0.649} 86.1 & \cellcolor[rgb]{0.674,0.581,0.686} 76.9 & \cellcolor[rgb]{0.639,0.563,0.663} 83.2 & \cellcolor[rgb]{0.630,0.558,0.656} 84.6 & \cellcolor[rgb]{0.757,0.629,0.732} \textbf{62.0} & \cellcolor[rgb]{0.760,0.631,0.733} 61.5 & \cellcolor[rgb]{0.718,0.605,0.712} \underline{69.4} & \cellcolor[rgb]{0.848,0.703,0.773} 42.5 & \cellcolor[rgb]{0.703,0.596,0.703} \underline{72.3} & \cellcolor[rgb]{0.794,0.655,0.749} 54.8 & \cellcolor[rgb]{0.785,0.649,0.745} 56.5 & \cellcolor[rgb]{0.715,0.603,0.710} 69.7 & \cellcolor[rgb]{0.867,0.724,0.783} 1.1 & \cellcolor[rgb]{0.929,0.819,0.831} 17.1 \\
  GPT-5 mini & -- & \cellcolor[rgb]{0.617,0.551,0.647} \underline{86.9} & \cellcolor[rgb]{0.655,0.571,0.674} 80.4 & \cellcolor[rgb]{0.655,0.571,0.674} \underline{80.4} & \cellcolor[rgb]{0.642,0.564,0.665} 82.6 & \cellcolor[rgb]{0.633,0.559,0.659} 84.0 & \cellcolor[rgb]{0.763,0.633,0.734} 60.7 & \cellcolor[rgb]{0.751,0.625,0.729} \underline{63.0} & \cellcolor[rgb]{0.718,0.605,0.712} 69.3 & \cellcolor[rgb]{0.684,0.586,0.692} 75.2 & \cellcolor[rgb]{0.712,0.601,0.708} 70.5 & \cellcolor[rgb]{0.769,0.637,0.737} \textbf{59.6} & \cellcolor[rgb]{0.721,0.607,0.713} \textbf{68.4} & \cellcolor[rgb]{0.696,0.593,0.700} 73.4 & \cellcolor[rgb]{0.777,0.643,0.741} \textbf{1.7} & \cellcolor[rgb]{0.841,0.696,0.770} \textbf{43.8} \\
  \bottomrule
  \end{tabular}
  }
  \end{table}
\begin{figure*}[th]
    \centering
    \includegraphics[width=\linewidth]{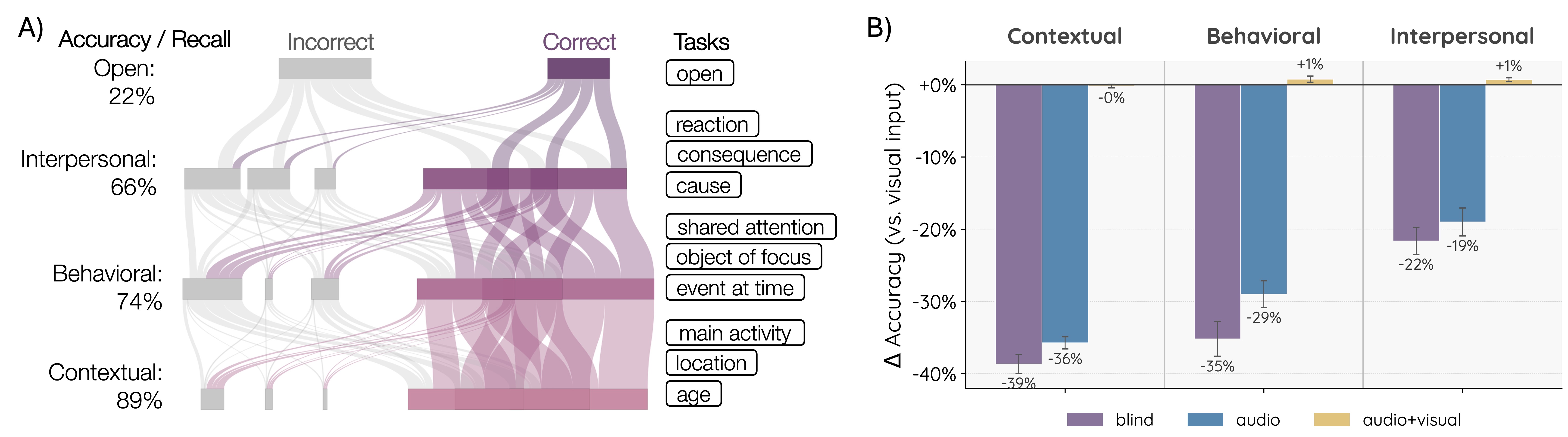}
    \caption{\textbf{A)} Sankey plot of the best open-source model Qwen 3.5-27B, showing how correct (purple) and incorrect (gray) questions change across \bench's levels. \textbf{B)} Model performance by input type relative to the visual input, the results highlight the non-verbal and visual nature of \bench.}
    \label{fig:sankey_baselines}
\end{figure*}

\subsection{Dataset description} \label{sec:nest_statistics}
\bench comprises \totalvids segments and \totalqs questions drawn from 20+ studies, 150+ participants, and 10 countries, with developmental stages ranging from infancy to school age (Figure~\ref{fig:stats}).

\noindent\textbf{Diversity of clips.} We prioritize real-world data to reflect the heterogeneity of early-childhood interactions. Clips vary substantially across demographics, recording location, and acquisition conditions: recording settings are predominantly naturalistic homes (60\%+), with the remainder drawn from labs and classrooms, and most interactions are dyadic, reflecting the structure of early child-caregiver exchange. (Fig.~\ref{fig:people_setting_datasource}) This diversity supports evaluation of model robustness beyond a single collection setting (Fig.~\ref{fig:stats} A and B and Sec.~\ref{supp:stats}). The dataset also captures a wide spectrum of interaction dynamics (Fig.~\ref{fig:stats} C), ranging from brief, subtle responses (such as rapid gaze shifts or small posture adjustments) to longer, sustained actions such as reaching toward or passing objects. These patterns occur in both child-led and adult-led contexts, reflecting the asymmetric and temporally varied nature of natural social exchanges.

\noindent\textbf{Dense compositional annotations.} Each segment is paired with an average of 11.6 questions spanning all four annotation axes (contextual, behavioral, interpersonal, and open-ended) yielding dense, multi-dimensional coverage of every interaction. The resulting density of annotations enables per-segment diagnostic analysis: a model's performance on contextual recognition can be compared directly against its performance on behavioral and interpersonal reasoning for the same clip, supporting the stratified evaluation used throughout our experiments. Table~\ref{tab:supp_question_counts} reports the full distribution of 
question types per segment.

\noindent\textbf{No-reaction segments.} To test for false positives in social reasoning, where models predict a plausible reaction from prior expectations rather than from what the clip actually shows, we include \emph{no-reaction} segments. These clips contain either (i) an action by one participant that elicits no observable response from the other, or (ii) no salient action and no reaction. For example, a caregiver may hold out a toy toward the child while the child continues looking elsewhere, showing no gaze shift, reach, or change in posture. No-reaction segments account for approximately 20\% of the dataset (\(n=265\)) and serve as explicit negative controls, requiring models to distinguish genuine action–reaction dynamics from prior expectations about what should follow.
\section{Experiments}
\label{sec:experiments}

\vspace{-3pt}

\begin{figure*}[t]
\centering
\includegraphics[width=\linewidth]{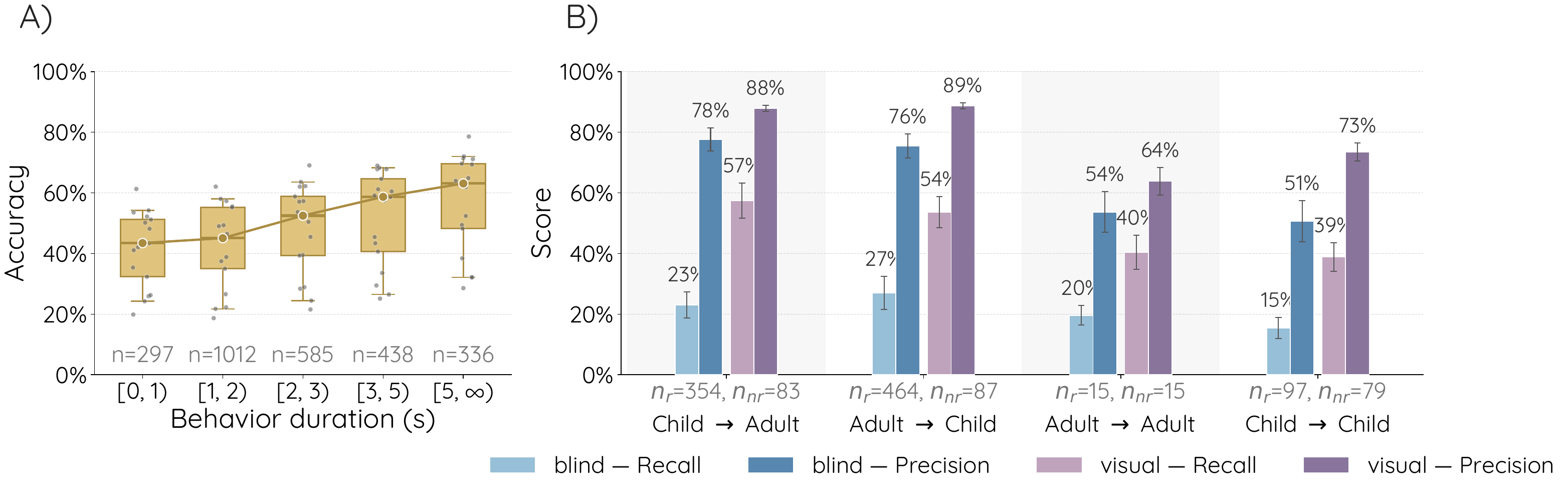}
\caption{
\textbf{A)} Per-model accuracy on \emph{event-at-time-$t$} questions
binned by the question's action duration (seconds). Gray dots are individual models. $n$ = unique questions per bin.
\textbf{B)} Recall and precision on (\emph{Is Reaction?}) questions by aggregated
initiator $\rightarrow$ responder dyad, averaged across all models ($\pm$ SEM). $n_r$ = number of reaction segments; $n_{nr}$ = number of no-reaction segments.}
\label{fig:diagnostics}
\vspace{-10pt}
\end{figure*}

\subsection{Experimental setup}
\noindent\textbf{Model selection.}
We target our evaluation on video-VLMs capable of direct video processing with raw video as input. To ensure a fair comparison given variations in native context windows, we standardize the evaluation by setting the maximum sequence length to 32,768 tokens and a maximum image side of 512 pixels to preserve aspect ratio. Rather than imposing a restrictive fixed frame rate that would bottleneck performance to the least capable model, this approach allows higher-capacity models to use their full temporal resolution for capturing fine-grained details essential in this context. We evaluate both open-source (VLLM~\citep{kwon2023efficient}) and closed-source models (through their APIs). 

\noindent\textbf{Metrics.} We assess model performance with accuracy (\%) for MCQs and balanced accuracy (\%) on classification questions. To account for varying levels of detail in open-ended responses, we follow the LLM-as-Judge paradigm from \citet{liu2024mmbench} with a 0-3 similarity score, prioritizing semantic meaning over exact phrasing. However, VLMs often favor generic summaries (e.g., 'toys' vs. 'blue car'), causing holistic scores to obscure missing details. To enforce granular rigor, we introduce a complementary claim-level recall metric (\%). Inspired by existing literature \citep{min2023factscore, ostmeier-etal-2024-green}, the judge decomposes ground-truth narratives into atomic interaction claims and scores binary matches in the generated answer.

\subsection{Main results}\label{sec:main_results}

Tables~\ref{tab:main_results} and~\ref{tab:main_find} report the performance of state-of-the-art video VLMs across \bench's reasoning levels on our public and \FIND subsets, respectively.\footnote{Due to the private nature of the \FIND dataset, closed-source models served by third-party companies were not evaluated on this subset. To promote a fair comparison, Table~\ref{tab:main_results} shows results of the public subset of NEST across all tested models, while open-source performance on \FIND is reported in the supplementary (Table~\ref{tab:main_find}).} Across both subsets, models reliably identify contextual attributes but performance collapses as tasks shift toward fine-grained behavioral and interpersonal cues (Fig.~\ref{fig:sankey_baselines}A), with the difficulty of interpersonal reasoning compounded by the precise temporal grounding it demands (e.g., shared attention, event at time t). The same pattern persists for open-ended descriptions: similarity to ground truth remains moderate (0.7-1.7 on a 0-3 scale) and recall stays below 50\%, indicating partially aligned outputs that fail to compose behavioral cues into coherent interaction-level descriptions. 


\noindent\textbf{Asymmetric reasoning across causal direction.} 
Action–reaction pairs can be probed in two directions: predicting the consequence given the action, or inferring the cause given the reaction (see Fig~\ref{fig:tasks}). Across models (Tab.~\ref{tab:main_results}), consequence prediction is significantly harder than causal attribution (Wilcoxon signed-rank, $p=1.5\times 10^{-5}$; Fig.~\ref{fig:cause_conseq_init}). We hypothesize that this stems from information embedded in the reaction itself: a question like \textit{"why did the child reach towards the toy?"} is strongly constrained by the preceding action (the caregiver offered it), whereas the converse (predicting how a child will respond to an offered toy) admits many equally plausible reactions.

\noindent\textbf{Difficulty in temporal grounding.}
We examine how performance varies across event-at-time-t questions as a function of the action's duration (Figure~\ref{fig:diagnostics}A). Accuracy increases monotonically with action duration, rising from a median of $\sim$44\% for actions under 1 second to $\sim$60\% for actions of 5 seconds or longer. This pattern suggests that brief, fast-occurring actions are systematically harder to localize in time than longer, more sustained ones. We hypothesize that shorter action spans involve fewer sampled video frames, leaving models with less visual evidence to anchor the event temporally. 


\subsection{Diagnostic analyses}

\noindent\textbf{\bench is non-verbal and visual in nature.} To ensure \bench is not solvable via language priors or transcribed speech, we evaluate three ablations all of which receive the text prompt: blind (text-only), audio (with Whisper large-v3 transcriptions~\cite{radford2023robust}), and audio+visual. Figure~\ref{fig:sankey_baselines}B compares these against the standard visual input. This experiment reveals that \bench is fundamentally visual and non-verbal in nature: blind and audio variants degrade performance across all levels (avg. $-32\%$ and $-28\%$). Unlike prior adult-centric benchmarks where speech dominates, combining audio and visual inputs in \bench yields only marginal benefits (avg. $+1\%$). Fig.~\ref{fig:audio_comparison} illustrates cases where audio can be helpful, particularly for activity recognition and consequence reasoning in multilingual settings.

\noindent\textbf{Bias in child-centric scenes.} Do models default to assuming an interaction is occurring just because multiple people are present? To test this underexplored condition, we evaluate negative instances defined as "no-reaction" segments, where there is low social engagement between the people (Sec.~\ref{sec:nest_statistics}). Specifically, we ask: \textit{After Person A performs ⟨action$\rangle$, does Person B respond?} This formulation explicitly tests whether models hallucinate false positives in passive scenarios. VLMs exhibit two systematic biases (Fig.~\ref{fig:diagnostics} B). Without visual input, models default to predicting no reaction: precision reaches 51-78\%, but recall collapses to 15-27\%. With visual input, recall recovers asymmetrically. Same-dyad interactions (adult$\to$adult, child$\to$child) remain harder (39-40\%) than cross-dyad interactions (54-57\%), suggesting that VLMs lean on coarse demographic cues to parse interaction roles.

\noindent\textbf{Performance across video quality and demographics.} 
Additional analyses (Appendix~\ref{sec:additional_exp}) show that accuracy is broadly stable across video-quality bins, and event-at-time-$t$ performance varies little with child age. Differences across recording continents are modest, while setting differences depend on the task: lab clips show higher object-attention and reaction accuracy than home and classroom clips. These setting comparisons are descriptive, since setting is largely nested within source corpus.

\subsection{Towards VLM-assisted child interaction analysis}
Current VLMs cannot automate child interaction analysis on their own, but they can play a useful role in larger systems that speed up workflows for early-childhood researchers and practitioners. To guide the design of such systems, we compare VLMs to specialist models, test their learning capacity through finetuning, and check if \bench performance transfers to real parent-coaching footage.

\begin{table}[t]
\caption{Specialist models compared to top-performing VLMs on the public subsets of NEST (ChildPlay and CHILDES). \textbf{Best} and \underline{second best}.}
\label{tab:specialist_vs_vlm}
\vspace{0.5em}
\centering
\resizebox{\linewidth}{!}{%
\begin{minipage}[t]{0.49\textwidth}
\centering
\textbf{Age Estimation}\\[4pt]
\begin{tabular}{lcc}
\toprule
Model & Params & Acc. (\%) \\
\midrule
MiVOLO v2 \cite{mivolo2023} & 0.1B & 62.2 \\
\midrule
Qwen3.5 & 4B & 85.5 \\
Qwen3.5 & 27B & \textbf{87.3} \\
Gemini 3.1 Flash Lite & - & \underline{86.5} \\
\bottomrule
\end{tabular}
\end{minipage}\hfill
\begin{minipage}[t]{0.49\textwidth}
\centering
\textbf{Shared Attention}\\[4pt]
\begin{tabular}{lcc}
\toprule
Model & Params & Acc. (\%) \\
\midrule
Gazelle \cite{ryan2025gaze} & 0.3B & \textbf{64.7} \\
SemGaze \cite{tafasca2024toward} & 0.2B & \underline{62.7} \\
\midrule
Qwen3.5 & 4B & 56.0 \\
Qwen3.5 & 27B & 62.0 \\
Gemini 3.1 Flash Lite & - & 62.0 \\

\bottomrule
\end{tabular}
\end{minipage}\hfill
\begin{minipage}[t]{0.49\textwidth}
\centering
\textbf{Object of Attention}\\[4pt]
\begin{tabular}{lcc}
\toprule
Model & Params & Acc. (\%) \\
\midrule
SemGaze \cite{tafasca2024toward} & 0.2B & 42.0 \\
\midrule
Qwen3.5 & 4B & \underline{85.4} \\
Qwen3.5 & 27B & \textbf{91.7} \\
Gemini 3.1 Flash Lite & - & 84.6 \\
\bottomrule
\end{tabular}
\end{minipage}
}
\vspace{-5pt}
\end{table}

\noindent\textbf{The value of specialist models.}
We evaluate the trade-offs between generalist VLMs and specialist models across three \bench tasks (Table~\ref{tab:specialist_vs_vlm}). While VLMs demonstrate strong zero-shot capabilities in age estimation and flexibility across understudied tasks (e.g., object of attention), specialist gaze models achieve superior accuracy on shared attention with a fraction of the parameters. Consequently, given sufficient annotated data, specialists deliver targeted precision that makes them ideal modular tools or priors within larger systems.

\begin{figure*}
    \centering
      \includegraphics[width=\linewidth]{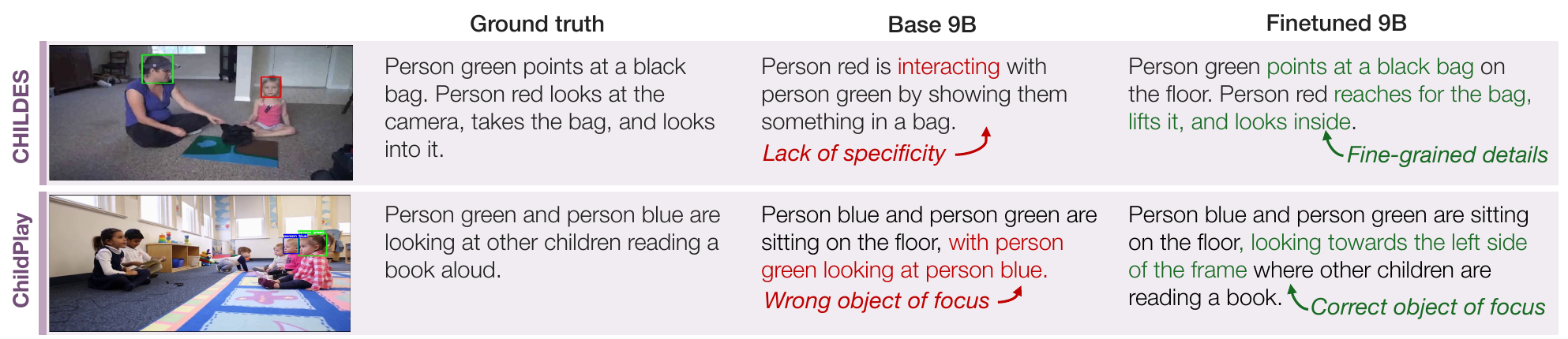}
      \captionof{figure}{\textbf{Qwen3.5-VL 9B predictions before vs.\ after finetuning} for two held-out \bench{} items (one ChildPlay, one CHILDES). Each row shows the keyframe, human ground truth, and predictions from the base and finetuned models.}
      \label{fig:finetune_examples}
      \vspace{-10pt}
\end{figure*}

\textbf{Supervised finetuning improves the language of non-verbal interaction.} Open-ended evaluation reveals a sharp degradation: free-form descriptions collapse to generic phrases (\emph{``interacting''}, \emph{``playing with toys''}) with recall below 30\% across the evaluated open-source models, even on clips where the model answers the corresponding closed-form questions correctly. This raises the question of whether VLMs lack exposure to the language of non-verbal interaction or whether the gap reflects deeper perceptual limits. We test this by finetuning Qwen3.5-\{4B, 9B, 27B\} with LoRA~\cite{hu2022lora} on a separate pool of CHILDES and \FIND clips outside the \bench benchmark (train: 485 CHILDES; val: 137 \FIND). We evaluate on \bench-filtered ($N=683$ clips), removing any source videos that overlap with the finetuning to prevent data leakage.

Finetuning yields $+0.15$ in similarity and $+12\%$ in recall on the 9B model, bringing it to parity with the 27B base (sim: $1.21$ vs.\ $1.26$; rec: $0.23$ vs.\ $0.24$). Gains shrink at 4B (sim: $+0.11$, rec: $+8\%$) and reverse at 27B (sim: $-0.08$, rec: $-0.5\%$) (Table~\ref{tab:finetune}). A possible explanation is that the 27B base is already strong on this vocabulary, and LoRA overfits the small training set. Figure~\ref{fig:finetune_examples} shows the finetuned 9B, which replaces generics with concrete action sequences, corrects attention targets, and adds fine-grained physical details. This points to one tractable component of the gap: the language of non-verbal interaction is learnable from modest task-specific data. The deeper perceptual failures \bench surfaces are not addressed here and remain open research targets.

\begin{table}[h]
\centering
\caption{\textbf{Finetuning impact on the NEST-no-overlap open-interaction held-out set (n = 683).} Subscript \emph{before} denotes the un-finetuned base model; $\Delta$ columns report the change after finetuning (finetuned $-$ base). We report similarity ($0$-$3$) and recall ($0$-$1$). A change of +0.12 corresponds to a gain of 12 percentage points.}

\small
\setlength{\tabcolsep}{8pt}
\begin{tabular}{lrrrr}
\toprule
\textbf{Model} & \textbf{Sim$_\mathrm{before}$} & \textbf{$\Delta$Sim} & \textbf{Rec$_\mathrm{before}$} & \textbf{$\Delta$Rec} \\
\midrule
Qwen3.5-VL 4B  & 1.00 & \dpos{+0.11} & 0.11 & \dpos{+0.08} \\
Qwen3.5-VL 9B  & 1.06 & \dpos{+0.15} & 0.11 & \dpos{+0.12} \\
Qwen3.5-VL 27B & 1.26 & \dneg{-0.08} & 0.24 & \dneg{-0.01} \\
\bottomrule
\end{tabular}
\label{tab:finetune}
\vspace{-0.5cm}
\end{table}

\textbf{Case-study: Parent-coaching videos.} A direct application of \bench is in interventional programs, where  computer vision could reduce experts' manual labeling efforts. We study FIND, a parent-coaching program in which coaches show caregivers or educators video examples of their developmentally supportive interactions. Each example pairs a clip with a description to help caregivers recognize and reinforce the behavior. Currently, selecting examples and annotating descriptions is done manually by a trained editor. These editor hours are the main bottleneck to scaling the program to more families and educators. This case study aims to assess the feasibility of using VLMs to help editors generate descriptions for selected clips.

However, as NEST shows, fully open description generation by VLMs lacks the precision needed for this type of application. Instead, we study VLM generation on two specific elements of the description: a \emph{child serve} (a child-initiated action) and \emph{adult return} (the caregiver's response). This is the same action–reaction structure \bench's interpersonal level is built around. Thus we measure zero-shot recall for serve and return in program-annotated clips (avg. $\sim 11.6$s) using Qwen3.5-\{9B, 27B\}. The input to the model is the video clip, the definition of the element required, and a description of the complementary element for that video. Appendix~\ref{app:case_study} details the preprocessing which uses the knowledge gained from NEST.

Results mirror \bench's findings. Visual-only recall is low across both roles (9B: 12.7\% / 10.5\%; 27B: 15.6\% / 12.2\%). Audio benefits adult returns substantially more than child serves (27B: +13.5\% vs. +2.9\%). This extends our main finding that audio provides limited signal for the visually-grounded behaviors \bench targets. The failure modes \bench surfaces persist in real coaching workflows.

Together, these findings suggest that a socially grounded AI system for early childhood needs a coordinated system rather than one end-to-end model. This includes generalist VLMs for contextual reasoning, specialist detectors for perception and temporal grounding, domain finetuning where descriptive language falls short, and human review for the hardest interpersonal cases. \bench provides the instrument to make these composition choices empirically by measuring which components to trust, where, and for what. We release it to support the development of reliable tools for researchers, clinicians, and practitioners who depend on understanding children's earliest social interactions.


\vspace{-0.1cm}
\section{Conclusion}
\label{sec:conclusion}
\vspace{-5pt}

While automated video analysis promises to transform developmental science, current VLMs cannot reliably interpret the non-verbal social cues defining child–caregiver interaction. This gap has broad consequences because VLMs deployed in homes, schools, pediatric care, and assistive systems will routinely encounter children and families but have been evaluated almost exclusively on adult, conversation-driven social content. \bench addresses this gap as the first naturalistic, child-centric video benchmark with a compositional schema that breaks down social reasoning into contextual, behavioral, and interpersonal cues. Our evaluation reveals a sharp drop from strong contextual recognition to weak higher-order reasoning. This is traceable to specific failure modes: poor temporal grounding, weaker consequence than cause prediction, and overlooked dyadic configurations that persist in real coaching workflows. By making these failure modes measurable, \bench turns child–caregiver interaction from a blind spot into a tractable target for future work. Understanding human social behavior includes understanding children and caregivers. In this work, we aim to provide a rigorous framework for advancing socially grounded modeling.


\begin{ack}
This work was supported by the Stanford Clinical Excellence Research Center (CERC), the Heising-Simons Foundation (Grant \#2024-5555) and the Bill and Melinda Gates Foundation (Gates Award INV-035364). We thank our annotators and collaborators, whose contributions and discussions made this work possible. The authors declare no competing interests.
\end{ack}

\bibliographystyle{abbrvnat}
\bibliography{main}

\appendix
\newpage
\addtocontents{toc}{\protect\setcounter{tocdepth}{2}}
\section*{Appendix for \bench}
\tableofcontents
\newpage

\section{Privacy and release details}
\label{app:privacy_release}

\subsection{Privacy}
\label{app:privacy}

NEST draws on existing recordings of children and caregivers, whose identities may remain recognizable in the source videos. Our secondary annotation and analysis were conducted under Stanford IRB-approved protocol 66624, which covers annotator recruitment, training, and supervision, as well as handling of the reviewed segments. Table~\ref{tab:provenance_oversight} summarizes the provenance, collection oversight, and access arrangements for each corpus.

ChildPlay comprises publicly available YouTube recordings released as a research dataset. CHILDES recordings originate from research studies whose consent procedures are documented by the contributing investigators, with video access through TalkBank registration. FIND recordings were collected under the intervention program's IRB-approved protocol, with caregiver consent at enrollment and for research purposes, and remain subject to the program's access controls. 

The public data release is limited to annotations and segment metadata for ChildPlay and CHILDES; their videos are not redistributed. FIND videos, annotations, questions, and segment metadata are excluded from the public data release. Closed-source models served through third-party APIs were not evaluated on FIND. During benchmark construction, clips containing sensitive content, including nudity, were excluded.

\begin{table}[h]
\centering
\caption{\textbf{Provenance and oversight.}}

\small
\setlength{\tabcolsep}{8pt}
\begin{tabular}{
    >{\raggedright\arraybackslash}p{\dimexpr0.22\linewidth-2\tabcolsep\relax}
    >{\raggedright\arraybackslash}p{\dimexpr0.24\linewidth-2\tabcolsep\relax}
    >{\raggedright\arraybackslash}p{\dimexpr0.27\linewidth-2\tabcolsep\relax}
    >{\raggedright\arraybackslash}p{\dimexpr0.27\linewidth-2\tabcolsep\relax}
}
\toprule
& \textbf{ChildPlay} & \textbf{CHILDES} & \textbf{\FIND} \\
\midrule
Segments / questions
& 316 / 3,837
& 736 / 8,359
& 156 / 1,828 \\

Source videos / studies
& 80 / 1
& 333 / 17
& 58 / 4 \\

Original collection
& Public YouTube video, released as a research dataset
& Research studies contributed to TalkBank
& Parent-coaching intervention program \\

Consent \& oversight at collection
& ChildPlay dataset terms
& Documented by contributing investigators; gated by TalkBank registration
& Program's IRB-approved protocol; caregiver consent at enrollment and for research purposes \\

Our secondary use
& Stanford IRB 66624
& Stanford IRB 66624
& Stanford IRB 66624; accessed under the Intervention Program protocol \\

We redistribute
& Annotations + metadata
& Annotations + metadata
& Nothing \\

Video access
& YouTube
& TalkBank registration
& Intervention program \\
\bottomrule
\end{tabular}
\label{tab:provenance_oversight}
\end{table}

\subsection{Release Manifest}

The public benchmark release comprises 12,196 questions and reference answers covering 1,052 segments: 316 from ChildPlay and 736 from CHILDES. It includes multiple-choice options and correct-answer labels, open-ended reference descriptions, the annotation schema and protocol spreadsheet, and segment metadata specifying the source video and corresponding 12-second window. We also release evaluation code, together with utilities for extracting benchmark segments from source videos obtained by the user. Table~\ref{tab:release_manifest} lists the released artifacts and their access requirements.

Annotations are released under CC BY-NC-SA 4.0. Source-video access remains subject to the original providers' terms and procedures: YouTube for ChildPlay and TalkBank registration for CHILDES. The annotation license does not alter these access arrangements. FIND materials remain accessible only through the intervention program's own authorization procedures and are not distributed with NEST.

\begin{table}[h]
\centering
\caption{\textbf{Release manifest.} Annotations: CC BY-NC-SA 4.0.}

\small
\setlength{\tabcolsep}{8pt}
\begin{tabular}{
    >{\raggedright\arraybackslash}p{\dimexpr0.13\linewidth-2\tabcolsep\relax}
    >{\raggedright\arraybackslash}p{\dimexpr0.43\linewidth-2\tabcolsep\relax}
    >{\raggedright\arraybackslash}p{\dimexpr0.20\linewidth-2\tabcolsep\relax}
    >{\raggedright\arraybackslash}p{\dimexpr0.24\linewidth-2\tabcolsep\relax}
}
\toprule
\textbf{Category} & \textbf{Artifact} & \textbf{Released} & \textbf{Access} \\
\midrule
Data
& Questions + reference answers (12,196)
& Full, public segments
& Open \\

& Multiple-choice options (generated distractors)
& Yes
& Open \\

& Annotation schema + protocol spreadsheet
& Yes
& Open \\

& Segment metadata (source video ID, 12\,s window)
& Yes
& Open \\

& ChildPlay / CHILDES videos
& No
& YouTube / TalkBank registration \\

& Intervention video, 156 segments, 1,828 questions
& No
& Intervention program \\
\midrule
Code
& Evaluation code
& Yes
& --- \\

& Segment extraction (from downloaded videos)
& Yes
& --- \\
\bottomrule
\end{tabular}
\label{tab:release_manifest}
\end{table}

Table~\ref{tab:public_reproducibility} summarizes reproducibility from the public release, which covers approximately \textbf{87\% of NEST's segments and questions}. Table~\ref{tab:main_results} can be reproduced entirely using the released ChildPlay and CHILDES annotations and code once the corresponding source videos have been obtained. Analyses combining these public subsets with FIND are partially reproducible: the same analyses can be rerun on the public subsets, but reproducing the reported full-dataset results requires authorized access to FIND. FIND-only benchmark results and the parent-coaching case study cannot be reproduced from the public release. The finetuning experiment is also partially reproducible: its CHILDES training data and 673 public held-out examples are released, while its 137 FIND validation examples used for checkpoint selection and 10 FIND held-out examples remain restricted.

\begin{table}[h]
\centering
\caption{\textbf{What is reproducible from the public release.}}

\small
\setlength{\tabcolsep}{8pt}
\begin{tabular}{
    >{\raggedright\arraybackslash}p{\dimexpr0.50\linewidth-2\tabcolsep\relax}
    >{\raggedright\arraybackslash}p{\dimexpr0.34\linewidth-2\tabcolsep\relax}
    >{\raggedright\arraybackslash}p{\dimexpr0.16\linewidth-2\tabcolsep\relax}
}
\toprule
\textbf{Result} & \textbf{Data required} & \textbf{Public-only?} \\
\midrule
Table~\ref{tab:main_results} --- model comparison, all reasoning levels
& Public subsets
& Yes \\

Fig.~\ref{fig:sankey_baselines}A --- level-wise collapse (Sankey)
& Public subsets + FIND
& Partial \\

Table~\ref{tab:blind} and Figure~\ref{fig:sankey_baselines}B --- blind ablations
& Public subsets 
& Yes \\

Figure~\ref{fig:sankey_baselines}B and Figure~\ref{fig:audio_comparison} --- audio ablations
& Public subsets + FIND
& Partial \\

Fig.~\ref{fig:diagnostics}A --- temporal grounding by action duration
& Public subsets + FIND
& Partial \\

Fig.~~\ref{fig:diagnostics}B --- no-reaction false positives, dyad bias
& Public subsets + FIND
& Partial \\

Fig.~\ref{fig:cause_conseq_init} --- cause--consequence asymmetry
& Public subsets + FIND
& Partial \\

Table~\ref{tab:specialist_vs_vlm} --- specialist vs.\ generalist models
& Public subsets 
& Yes \\

Table~\ref{tab:finetune} --- LoRA finetuning
& Train: 485 CHILDES; validation: 137 FIND; evaluation: 683 examples (673 public + 10 FIND)
& Partial \\

Table~\ref{tab:main_find} --- FIND subset results
& FIND
& No \\

App.~\ref{app:case_study} --- parent-coaching case study
& FIND
& No \\
\bottomrule
\end{tabular}
\label{tab:public_reproducibility}
\end{table}

\section{Benchmark details}

\begin{figure}[h]
    \centering
\includegraphics[width=0.6\linewidth]{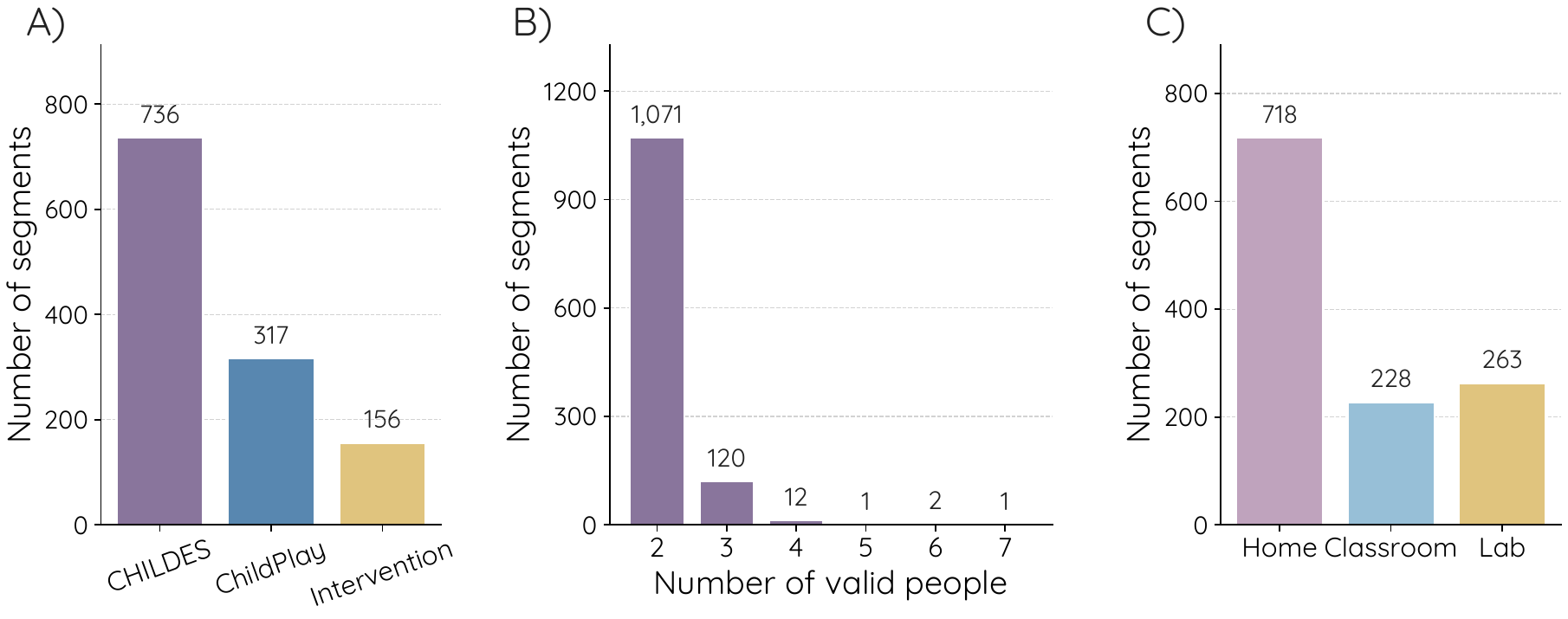}
    \caption{\textbf{Overview of dataset characteristics.} \textbf{A}) Number of segments by data source, totaling \totalvids segments across three corpora. \textbf{B}) Distribution of the number of annotated people per video segment, with most segments containing two individuals (typically a child and caregiver). \textbf{C}) Setting distribution.}
    \label{fig:people_setting_datasource}
\end{figure}

This section details the composition of NEST. The benchmark draws from three sources: CHILDES, ChildPlay, and \FIND. \FIND is a collection of videos from caregiver-child and early childhood classroom interactions collected in natural learning environments as part of the video coaching intervention program \textit{Filming Interactions to Nurture Development} (FIND) \cite{fisher2016promoting}. Table~\ref{tab:supp_question_counts} reports question counts per source and task category; Figure~\ref{fig:people_setting_datasource} summarizes segment-level characteristics including group size, recording setting, and source distribution.

\subsection{Statistics}
\label{supp:stats}

\begin{table}[ht]
\centering
\caption{Question counts per data batch and task category. Categories: \textbf{Contextual} (age, loc., main act.), \textbf{Behavioral} (shared attn., object attn., event attn.), \textbf{Interpersonal} (conseq., cause, react.), and \textbf{Open Inter.}\ (open-ended).}
\label{tab:supp_question_counts}
\resizebox{\linewidth}{!}{%
\begin{tabular}{lrrrrrrrrrrr}
\toprule
\textbf{Batch} & \multicolumn{3}{c}{\textbf{Contextual}} & \multicolumn{3}{c}{\textbf{Behavioral}} & \multicolumn{3}{c}{\textbf{Interpersonal}} & \multicolumn{1}{c}{\textbf{Open Inter.}} & \textbf{Total} \\
\cmidrule(lr){2-4} \cmidrule(lr){5-7} \cmidrule(lr){8-10} \cmidrule(lr){11-11} \cmidrule(lr){12-12}
 & age & location & main act. & shared attn. & object attn. & event attn. & conseq. & cause & react. & open inter. & Total \\
\midrule
ChildPlay & 657 & 316 & 312 & 275 & 154 & 736 & 378 & 378 & 315 & 316 & 3,837 \\
CHILDES & 1,539 & 737 & 704 & 640 & 352 & 1,592 & 649 & 681 & 729 & 736 & 8,359 \\
FIND & 388 & 154 & 134 & 139 & 56 & 346 & 150 & 150 & 155 & 156 & 1,828 \\
\midrule
\textbf{Total} & \textbf{2,584} & \textbf{1,207} & \textbf{1,150} & \textbf{1,054} & \textbf{562} & \textbf{2,674} & \textbf{1,177} & \textbf{1,209} & \textbf{1,199} & \totalvids & \totalqs \\
\bottomrule
\end{tabular}
}
\end{table}

\begin{figure}[h]
    \centering
    \includegraphics[width=0.6\linewidth]{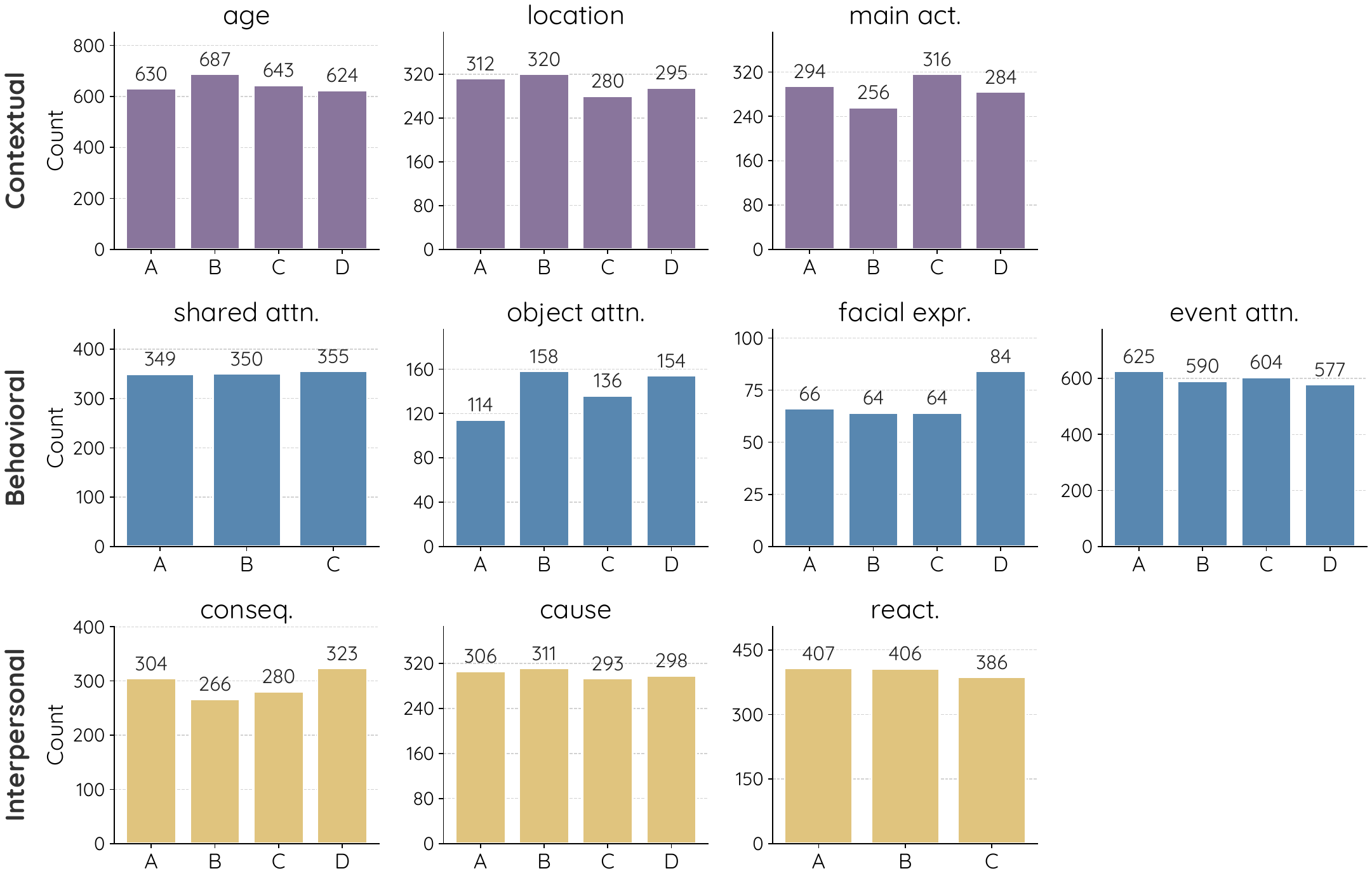}
    \caption{\textbf{Distribution of correct-answer letters per multiple-choice subtype, organized by question family.} Open-ended subtypes are excluded.}
    \label{fig:answer_distribution}
\end{figure}

To verify that the multiple-choice options are not positionally biased — a known failure mode of MCQ benchmarks — we report the distribution of the correct option letter across all multiple-choice subtypes (Figure~\ref{fig:answer_distribution}). Within every subtype the letters are approximately uniform, with no letter exceeding the expected $1/k$ frequency by more than a small amount. This holds for both 3-option subtypes (e.g., shared attention, reaction) and 4-option subtypes (e.g., age, event at time t).

\begin{figure}
    \centering
\includegraphics[width=0.9\linewidth]{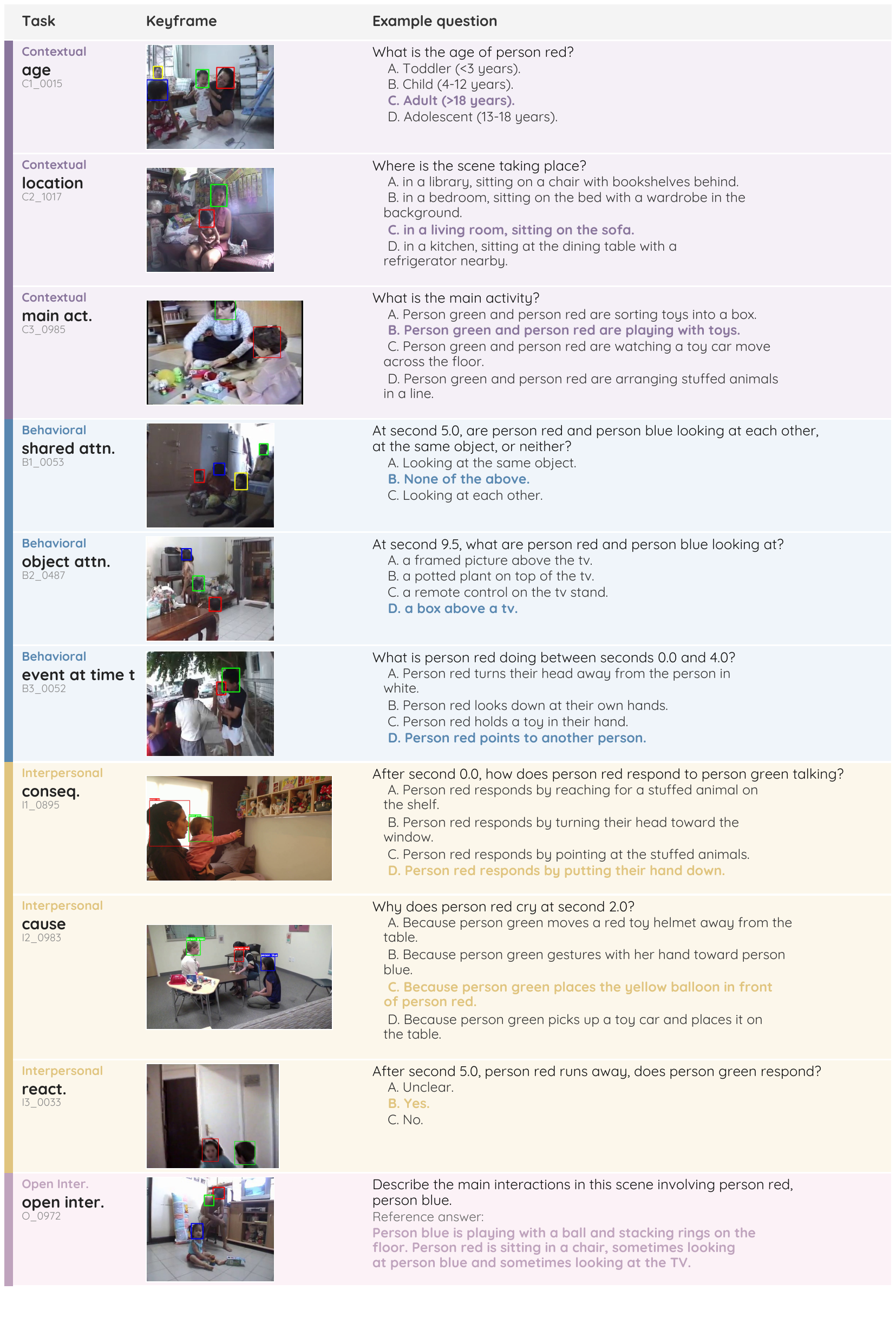}
    \caption{\textbf{Example \bench questions per task subtype.} Each row shows the keyframe with bounding boxes overlaid on the people referenced by color, the question presented to the model, and the answer choices with the correct option highlighted. Subtypes are color-banded by task family: \textbf{Contextual} (purple — age, location, main activity), \textbf{Behavioral} (blue — shared attention, object of attention, event at time t) \textbf{Interpersonal} (tan — consequence, cause, reaction), and \textbf{Open Inter.} (pink — open ended).}
    \label{fig:question_examples}
\end{figure}

To make the task taxonomy concrete, Figure~\ref{fig:question_examples} shows one
representative question per subtype, drawn directly from the released \bench split. Each row displays the keyframe with bounding boxes on the referenced people, the multiple-choice options (or free-text reference answer for the open-ended subtype), and the correct response highlighted in the family color.

\subsection{Selecting Videos}\label{supp:segment_selection}

Across our three corpora (ChildPlay, CHILDES, and \FIND), we select videos to maximize diversity of cultural contexts, recording settings, participant demographics, and interaction types. Selection was guided by metadata where available (e.g., CHILDES study tags, \FIND program labels) and by manual inspection. The final benchmark spans 22 studies across 10 countries, with participants of diverse ages (Figure~\ref{tab:data_sources}).

\begin{figure}[h]
\centering
  \caption{\textbf{Data sources and participants.}}
  \begin{tabular}{lccc}
    \toprule
    Corpora & \# Studies & \# Videos & \# Participants\\
    \midrule
    \FIND      & 4  & 58 & 43\\
    CHILDES   & 17 & 333 & 111\\
    ChildPlay & 1  & 80  & Unknown\\
    \bottomrule
  \end{tabular}
  \label{tab:data_sources}
\end{figure}

\begin{figure}[H]
    \centering
   \centering
  \resizebox{0.8\linewidth}{!}{\input{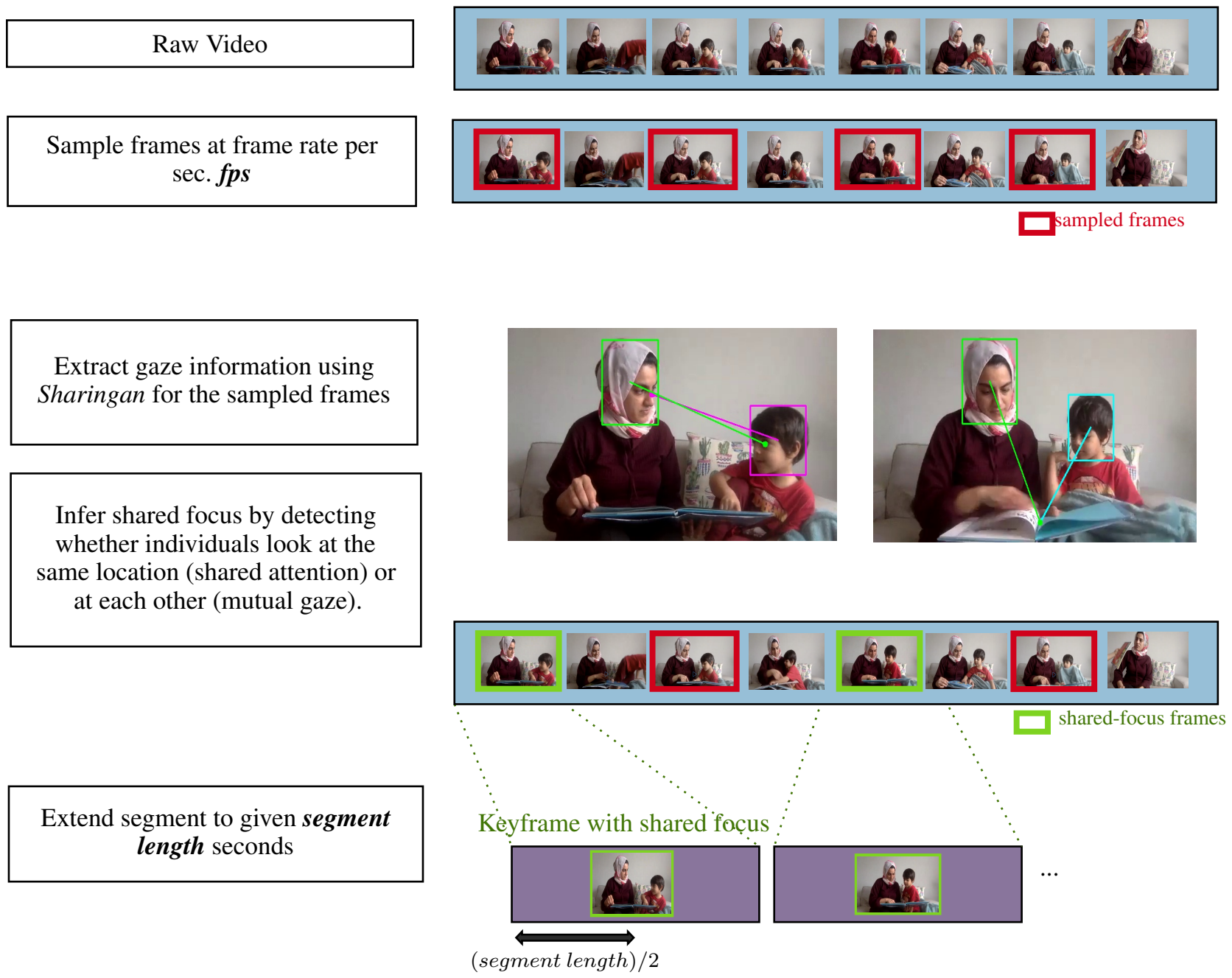}}
    \caption{\textbf{Clip extraction from long-form videos.} We use a gaze estimation model to sample candidate frames where the people are sharing each other's focus. The clip length is set to 12s and frame rate between clips to 0.2 fps.}
  \label{fig:segment_extraction}
\end{figure}

\noindent\textbf{Gaze-based clip extraction.} Figure~\ref{fig:segment_extraction} illustrates our extraction pipeline. From each long-form video, we extract short clips using gaze cues as a proxy for active social engagement. We first sample frames at 0.2 fps and run the gaze estimation model from \cite{tafasca2024sharingan} on each sampled frame, producing per-frame estimates of every detected individual's gaze target. We then identify \textit{shared focus} keyframes — frames where two or more individuals are estimated to be either looking at the same location (shared attention) or at one another (mutual gaze). Each such keyframe anchors a 12-second clip centered on the frame.

\noindent\textbf{Negative examples.} To support evaluation on cases of limited or absent social engagement, we additionally sampled 20\% of clips from windows that did not meet the gaze criterion. These segments serve as negative controls, requiring models to distinguish genuine action–reaction dynamics from common co-occurrence priors.

\subsection{Annotation Protocol}\label{sec:annotation_protocol}

\begin{figure}[ht]
    \centering
    \includegraphics[width=0.8\linewidth]{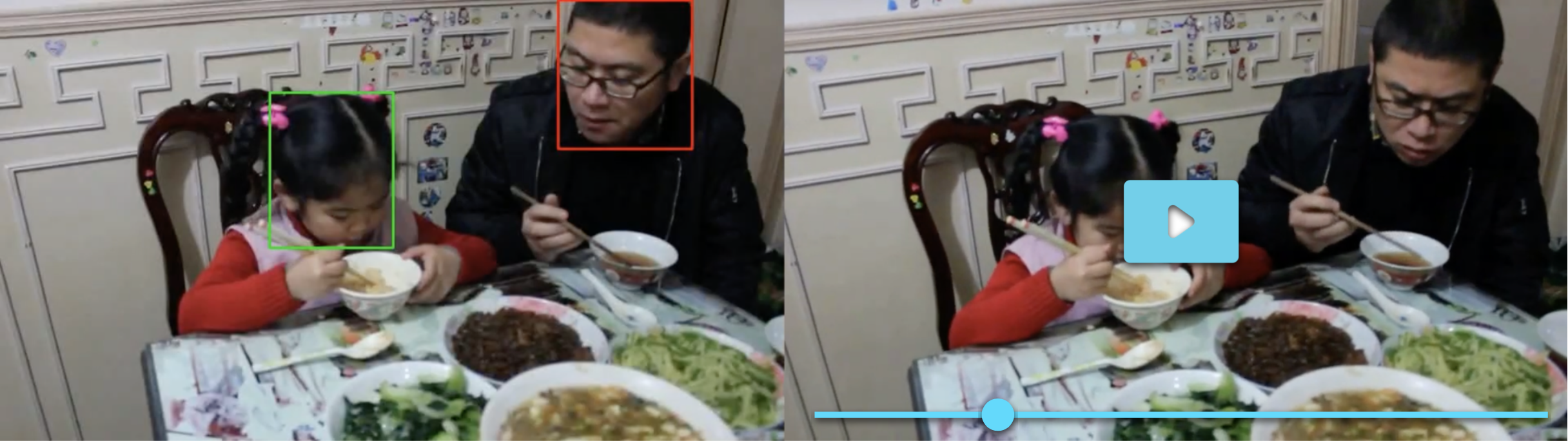} 
    \caption{\textbf{Video clip sample.} Left: Bounding boxes were overlaid on a keyframe from the video clip so annotators could refer to individuals by color. Right: The full video clip played alongside the keyframe.}
    \label{fig:clipsample}
\end{figure}

Annotators completed labels in a fixed order, each step building on the previous: (1) focus group identification and gaze validation, (2) contextual attributes, (3) action description with timing, and (4) interpersonal interaction dynamics. All annotations used the visual stream only — no audio — ensuring labels rely on observable cues. For each (keyframe, clip) pair, annotators were shown the selected keyframe (see Section~\ref{supp:segment_selection}) alongside its corresponding video clip, displayed side-by-side as in Figure~\ref{fig:clipsample}. To support consistent reference to individuals, color bounding boxes were overlaid on each detected person's head; annotators referred to individuals by bounding-box color in their natural-language responses. For each question, annotators were informed whether to refer to the keyframe, the segment, or both.

\subsubsection{Annotator recruitment and training}
Annotators were trained research assistants, onboarded with the labeling protocol and pilot examples before independent annotation. Annotation labels and protocols were co-designed with domain experts in early-childhood interaction from \FIND.

\subsubsection{Exclusion criteria}
Clips were excluded if they were too blurry to assess gaze reliably, or if the head detection model failed to identify valid individuals in the focus group. We additionally excluded clips containing sensitive content such as nudity. Excluded clips were not replaced; the final benchmark reflects only clips passing these criteria.

\subsubsection{Annotations}
Annotations were collected along three dimensions matching the benchmark's task families — contextual, behavioral, and interpersonal — using a structured schema co-designed with domain experts from \FIND. We release the full schema as a spreadsheet alongside our code for reproducibility.

\noindent\textbf{Contextual Annotations.} Contextual attributes were labeled using both the individual keyframe and the full clips. Using the keyframe, annotators were asked to identify which bounding boxes corresponded to real people rather than erroneous head detections, defining the \textit{valid group}. Then, for each individual in the valid group, annotators estimated their age range as one of: <1 (infant), 1–3 (toddler), 3–5 (preschool aged child), 6–12 (school aged), 13–17 (adolescent), or 18 (adult) years of age. When uncertain, multiple adjacent age ranges could be selected. These age bands reflect commonly used developmental stages from infancy through adolescence. Using the full clip, the annotators specified the location of the valid group (e.g. "inside a living room, sitting on a couch") and described the main activity that the group was engaged in (e.g. "eating at a table").

\noindent\textbf{Behavioral Annotations.} Given the role of gaze (Section \ref{sec:deconstructing}), annotators identified a focus group — a subset of individuals exhibiting one of three gaze conditions: (a) each looking at the other, (b) each looking at the same object, or (c) one or more looking at another individual who might be looking offscreen or at another entity. For cases (b) and (c), annotators provided a label for the object of focus (e.g., "yellow toy truck"). Annotators then grounded behaviors performed by members of the valid group, including a facial expression and three actions, each with start and end times.

\noindent\textbf{Interpersonal Annotations.} To capture the causal and reactive structure of social exchanges, annotators built on the actions and facial expressions identified above. They estimated whether the two sequential actions formed a response pair — that is, whether one was performed in reaction to the other. 
To support multiple-choice evaluation, annotators also provided a plausible distractor reaction: an alternative reaction consistent with the clip but not what actually occurred.

\noindent\textbf{Open Interaction}
 To capture details not covered by the structured annotations, annotators provided a plain-text description of interactions between individuals in the \textit{valid group} at any point in the clip. When no interaction was observed, they instead described the individual actions of \textit{valid-group} members.

\subsubsection{Quality control}
A 20\% subset of ChildPlay clips was annotated by 4 annotators and discussed before independent work began, ensuring shared interpretation of the protocol. Each completed annotation was then reviewed by a second annotator or research team member, and rejected annotations were reassigned or corrected.

\subsubsection{Inter-rater agreement}
\begin{figure}[h]
    \centering
    \includegraphics[width=\linewidth]{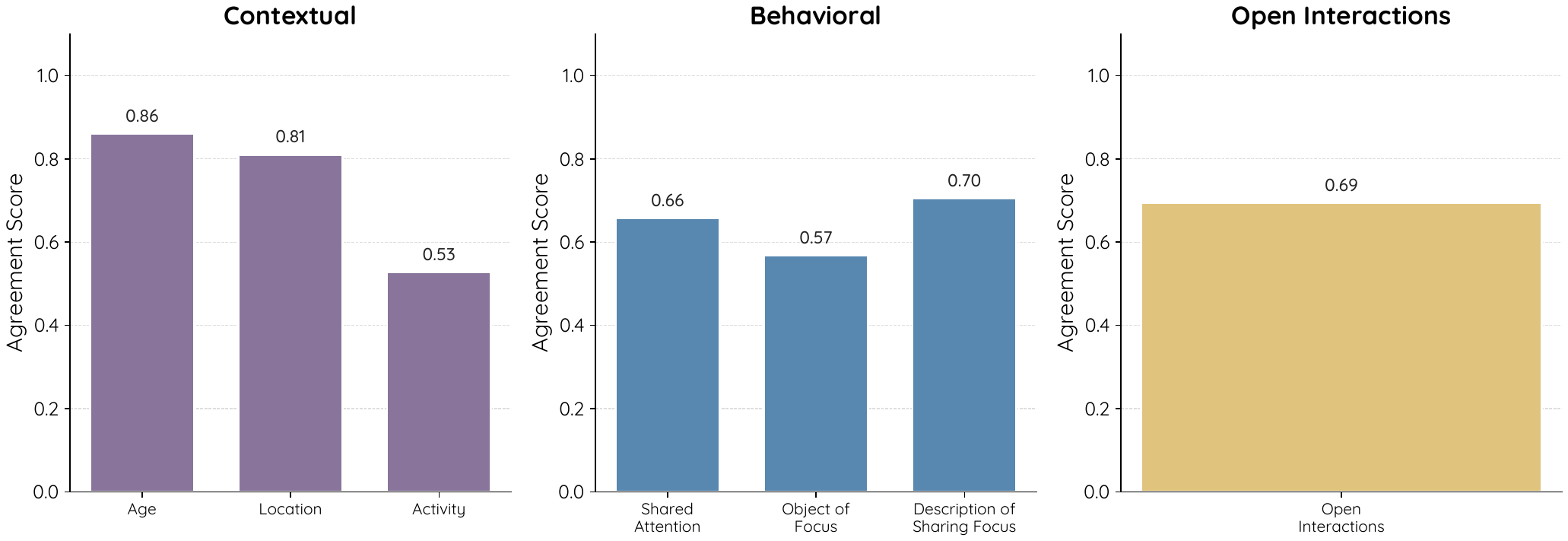}
    \caption{
        \textbf{Inter-rater agreement} for annotations in our dataset, measured on a subset of 64 video segments rated by 4 annotators. We report Krippendorff's $\alpha$ for categorical variables (Age, Shared Attention) and cosine similarity for free-text descriptions (Location, Activity, Object of Focus, Description of Sharing Focus, Open Interactions).
    }
    \label{fig:inter_rater_agreement}
\end{figure}

 On the double-annotated subset of ChildPlay clips, we computed Krippendorff's $\kappa$ for categorical labels and semantic similarity (cosine similarity over all-mpnet-base-v2 \citep{song2020mpnet} embeddings) for free-text labels. Figure~\ref{fig:inter_rater_agreement} reports per-label agreement. Structured labels such as age and shared attention reached substantial to high agreement ($\alpha$ between 0.66 and 0.86); free-text labels reached a mean cosine similarity of 0.66. The low-similarity cases were reviewed manually before proceeding. Some common sources of disagreement included borderline-ambiguous gaze direction in low-resolution frames, granularity mismatch, and stylistic differences. All disagreements were discussed before the second-pass review. Open Interactions, which involve free-form descriptions of social dynamics, show fair agreement ($0.69$), reflecting the inherent subjectivity of these annotations.

\subsection{Descriptions for identity grounding.}\label{sec:gen_descriptions}
\begin{figure}[h]
    \centering
    \includegraphics[width=\linewidth]{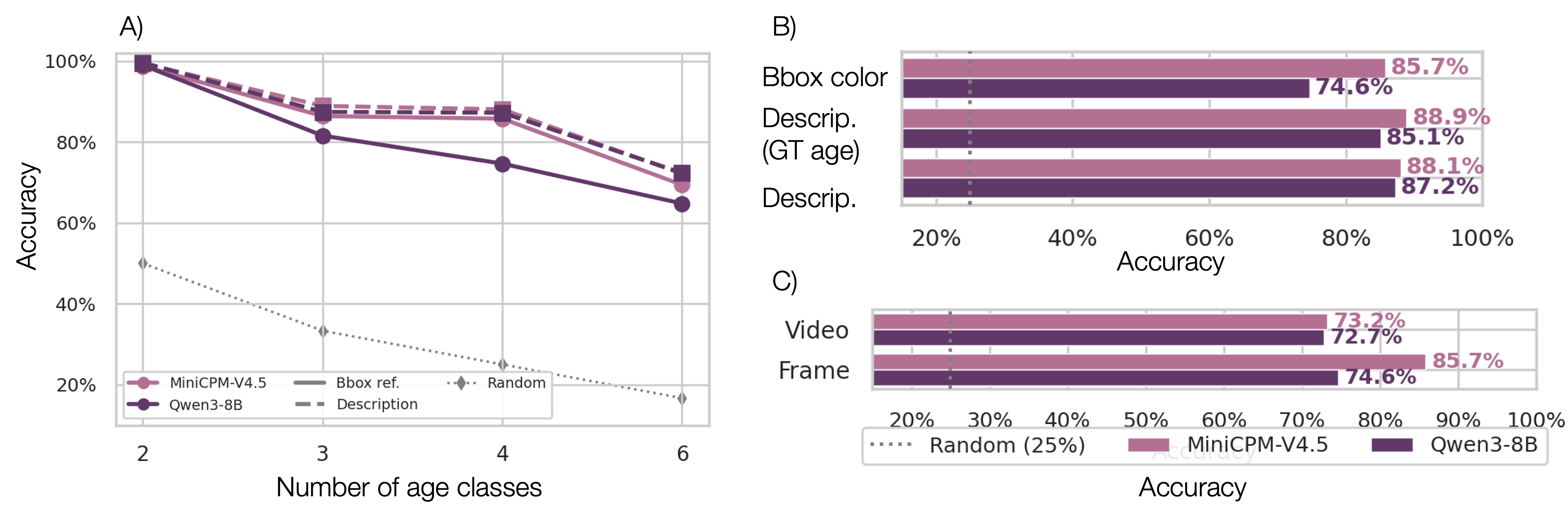}
    \caption{\textbf{Identity grounding.} We study the two open models Qwen3-VL-8B and Mini-CPM-v4.5 under three conditions: (A) age category groupings, (B) prompting strategy, and (C) input source.}
    \label{fig:prompting_exp}
\end{figure}

Establishing social interactions in video requires consistent person identification. However, generating continuous detection trajectories can be hindered by limitations in tracking itself (occlusions, video quality, detection performance). We study the utility of VLMs in maintaining identity across frames using age as a semantic proxy on the Childplay subset which has groundtruth tracklets. Fig.~\ref{fig:prompting_exp} A) shows that age serves as an effective identity marker, with task difficulty scaling with the categorization granularity. Our results (Fig.~\ref{fig:prompting_exp} B) and C) demonstrate that per-person descriptions from a single frame can effectively replace explicit tracking.  Ultimately, these single-frame VLM-derived descriptions perform comparably to ground-truth tracklets for person identification, offering an alternative for challenging videos where traditional tracking fails.

\subsection{Robustness of distractors}
\label{app:distractor_validity}

\noindent\textbf{Human vs. generated.} Consequence questions provide a within-question comparison of distractor sources: each contains one correct answer, one expert-written distractor, and two distractors generated by Qwen3-VL-32B. We analyze 15,012 responses from 17 models on 896 questions, including 7,645 incorrect responses. Conditioned on an incorrect response, uniform selection among the three distractors would assign 33.3\% of errors to the human-written option and 66.7\% to the two generated options together.

The human-written distractor accounts for 22.1\% of errors, while the two generated distractors jointly account for 77.9\%, or approximately 39.0\% per generated option on average. Thus, even after accounting for the number of options from each source, generated distractors attract a larger share of errors in this subset. This supports their effectiveness as challenging alternatives in the evaluated consequence questions.

\noindent\textbf{Generator-family comparison.} We additionally examine whether models from the generator's family tend to avoid generated distractors, which would increase the share of their errors assigned to the human-written option. The reported human-distractor error share is 21.0\% for Qwen-family models and 22.9\% for other models (Table~\ref{tab:distractor_validity}). These descriptive results do not show the hypothesized avoidance pattern, although they do not rule out other forms of model-family bias.

\begin{table}[h]
\centering
\caption{\textbf{Human-written distractor share among incorrect responses.} Each consequence question contains one human-written
and two generated distractors. Uniform selection among incorrect
options gives a human-written share of 33.3\%.}
\small
\setlength{\tabcolsep}{8pt}
\begin{tabular}{lr}
\toprule
\textbf{Model group / reference}
& \textbf{Human-written share (\%)} \\
\midrule
All models                       & 22.1 \\
Qwen-family models               & 21.0 \\
Other models                     & 22.9 \\
\midrule
Uniform choice among distractors & 33.3 \\
\bottomrule
\end{tabular}
\label{tab:distractor_validity}
\end{table}

\noindent\textbf{Scope and limitations.}
This analysis measures error attraction within questions containing both human-written and generated distractors; it does not compare separate all-human and all-generated question sets or establish semantic validity for every option. Rejection and revision counts of distractors were not logged during manual review, so these rates cannot be reported retrospectively.

\section{Experimental details}
\subsection{Compute Resources}\label{sec:compute_ressources} All experiments were run on a local Slurm cluster with NVIDIA Ampere GPUs.

\textbf{Distractor generation and description generation} We use Qwen3-VL-32B-Instruct \citep{bai2025qwen3} on a single Ampere GPU with 141 GB HBM (constraint=141G). Distractor generation additionally uses Qwen3-8B \citep{bai2025qwen3} as a lightweight grammar/coherence filter. Description generation runs at \texttt{max\_model\_len = 26{,}496}. When used as the judge model for SIM/REC scoring during evaluation, Qwen3-VL-32B \citep{bai2025qwen3} is run at \texttt{max\_model\_len = 32{,}768} with \texttt{gpu\_memory\_utilization = 0.7}.

\textbf{Finetuning.} We finetune three open-weight checkpoints — Qwen3.5-4B, Qwen3.5-9B, and Qwen3.5-27B — using Unsloth \citep{unsloth} with bf16 LoRA \citep{hu2022lora} (rank 16, $\alpha = 16$). Each run uses a single GPU: the 4B and 9B models fit on a 48 GB Ampere card with 64 GB system RAM; the 27B model requires an 80 GB A100 with 128 GB system RAM.

\textbf{Evaluation and judge.} We evaluate models and finetuned checkpoints by serving each model under vLLM \citep{kwon2023efficient} on a single 141 GB Ampere GPU and pairing it with the Qwen3-VL-32B \citep{bai2025qwen3} judge described above for SIM and REC metrics. 

\subsection{Human evaluation}

\subsubsection{Human evaluation of multiple-choice questions}

To contextualize VLM performance and verify that ground truths are recoverable from video alone, five expert raters familiar with the questions and tasks annotated a subset of the benchmark using a browser-based viewer that presented the same inputs given to the VLMs: silent video and the question text. Raters answered 664 unique (video, question) pairs spanning all nine question types: contextual and behavioral questions over 64 videos (16 per arm), interpersonal questions over 80 videos. Per-type accuracy ranged from 0.68 to 0.95, as shown in Table~\ref{tab:human_eval}. We computed inter-rater agreement only on the interpersonal questions, as they are the most subjective, and the other levels were already reviewed during benchmark construction; on the 83 pairs answered by three raters, Fleiss $\kappa$ ranged from 0.73 to 0.79, as shown in Table~\ref{tab:human_eval}.

\begin{table}[h]
\centering
\caption{\textbf{Human evaluation results.} For each (video, question) pair, we first average correctness across the raters who answered it, then average those per-question scores across questions within a task (macro-averaging at the question level), so each question contributes equally regardless of how many raters answered it. Inter-rater agreement (Fleiss' $\kappa$) is reported for the interpersonal tasks. $n$ denotes the number of unique (video, question) pairs evaluated; $n_{\text{overlap}}$ is the size of the multi-rater subset used for $\kappa$. Reaction reports balanced accuracy under a binary framing.}
\label{tab:human_eval}
\small
\setlength{\abovecaptionskip}{8pt}
\begin{tabular}{lrrrr}
\toprule
Task & $n$ & Accuracy & Fleiss' $\kappa$ & $n_{\text{overlap}}$ \\
\midrule
        \addlinespace[2pt]\multicolumn{5}{l}{\textit{Contextual}} \\
        \quad Age & 141 & 0.879 & -- & -- \\
        \quad Location & 64 & 0.906 & -- & -- \\
        \quad Main activity & 59 & 0.678 & -- & -- \\
        \addlinespace[2pt]\multicolumn{5}{l}{\textit{Behavioral}} \\
        \quad Shared attention & 59 & 0.932 & -- & -- \\
        \quad Object attention & 53 & 0.887 & -- & -- \\
        \quad Event at $t$ & 64 & 0.953 & -- & -- \\
        \addlinespace[2pt]\multicolumn{5}{l}{\textit{Interpersonal}} \\
        \quad Consequence & 108 & 0.940 & 0.789 & 41 \\
        \quad Cause & 36 & 0.861 & 0.731 & 12 \\
        \quad Reaction & 80 & 0.793 & 0.731 & 30 \\
\midrule
        \textbf{Overall} & \textbf{664} & \textbf{0.875} & \textbf{0.760} & \textbf{83} \\
\bottomrule
\end{tabular}
\end{table}

\subsubsection{Human validation of open-ended metrics}
\label{app:human_metric_validation}

We validate the automated evaluation of open-ended interaction descriptions against expert human judgments. Two expert annotators assessed 138 atomic reference claims across 31 model responses from 62 videos used in our multiple-choice human evaluation. For each claim, annotators indicated whether the model's response covered its content. We focus on Qwen3.5-27B, our strongest open model, and GPT-5 mini, our strongest closed model. Inter-rater reliability, measured over 35\% of claims coded by both annotators, is substantial (83\% agreement, Cohen’s $\kappa$ = 0.66, 95\% CI [0.46, 0.83]). All confidence intervals were estimated using 2,000 cluster bootstrap resamples of whole responses, preserving dependence among claims within each response.


Table~\ref{tab:human_metric_validation} compares human and automated judge scores for recall and similarity. Human-assessed recall was higher than the judge's for both models (Qwen3.5-27B: $+7.2$ points, 95\% CI $[-0.4, 14.6]$; GPT-5 mini: $+2.5$, $[-5.7, 11.0]$; pooled: $+4.8$, $[-0.8, 10.7]$), but every interval includes zero, so we detect no significant bias. The consistent direction suggests the judge is slightly conservative. At the claim level for Qwen3.5-27B, errors were asymmetric: the false-positive rate was 3.8\%, while the false-negative rate was 46.7\%, so the judge more often missed coverage that annotators recognized than credited coverage they did not. Human-assessed recall nonetheless remained low for Qwen3.5-27B (24.7\%; upper CI bound 33.5\%) and moderate for GPT-5 mini (49.6\%), supporting the conclusion that model descriptions omit substantial reference content. Similarity scores agreed closely, with differences of at most 0.10 on the 0-3 scale and all intervals spanning zero. We therefore find no detectable bias in the automated similarity metric and treat reported recall as a lower bound.

\begin{table}[h]
\centering
\caption{\textbf{Human validation of the open-interaction metrics.} We report human and automated judge scores for Qwen3.5-27B and GPT-5 mini ($n=31$ responses each), alongside combined results. Brackets indicate 95\% confidence intervals; $\Delta$ denotes human $-$ judge.}

\small
\setlength{\tabcolsep}{8pt}
\begin{tabular}{lrrcc}
\toprule
\textbf{Model} & \textbf{$n$} & \textbf{Judge}
& \textbf{Human [95\% CI]}
& \textbf{$\Delta$ [95\% CI]} \\
\midrule
\multicolumn{5}{l}{\textbf{Recall (\%), macro mean over responses}} \\
Qwen3.5-27B & 31 & 17.4 & 24.7 [16.5, 33.5] & $+7.2$ [$-0.4$, 14.6] \\
GPT-5 mini  & 31 & 47.2 & 49.6 [37.7, 61.8] & $+2.5$ [$-5.7$, 11.0] \\
\textbf{All} & 62 & 32.3 & 37.1 [29.2, 45.2] & $+4.8$ [$-0.8$, 10.7] \\
\midrule
\multicolumn{5}{l}{\textbf{Similarity (0--3), macro mean over responses}} \\
Qwen3.5-27B & 31 & 1.00 & 1.10 [0.89, 1.32] & $+0.10$ [$-0.21$, 0.42] \\
GPT-5 mini  & 31 & 1.68 & 1.58 [1.32, 1.84] & $-0.10$ [$-0.42$, 0.24] \\
\textbf{All} & 62 & 1.34 & 1.34 [1.16, 1.52] & $+0.00$ [$-0.24$, 0.23] \\
\bottomrule
\end{tabular}
\label{tab:human_metric_validation}
\end{table}

\subsection{Cause and consequence asymmetry}

\begin{figure}[h]
    \centering
    \includegraphics[width=0.4\linewidth]{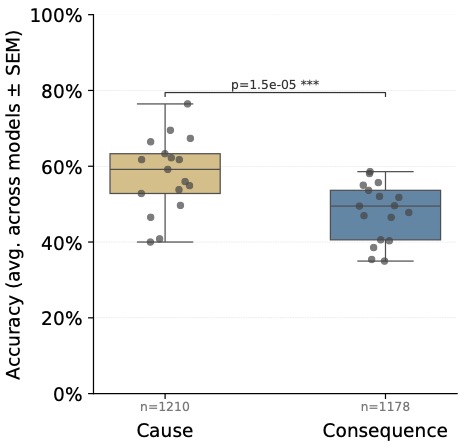}
    \caption{\textbf{Comparing performance across models for cause and consequence questions.}}
    \label{fig:cause_conseq_init}
\end{figure}

Figure~\ref{fig:cause_conseq_init} shows the asymmetry in causality direction for the interpersonal questions discussed in Section~\ref{sec:main_results}.

\subsection{Performance on \bench for the \FIND subset}
Table~\ref{tab:main_find} shows results across all open-source models on the \FIND subset of \bench. Due to the private nature of the \FIND dataset, closed-source models served by third-party companies were not evaluated on this subset. Evaluation on this held-out subset confirms the same patterns discussed for the public subsets of \bench in Section~\ref{sec:main_results}. \begin{table}[th]
\caption{\textbf{Model performance on the \FIND subset of \bench}. Open interactions are measured in similarity to ground truth (0-3) and recall (\%), all other tasks are in accuracy (\%). \textbf{Best} model and \underline{second best} model. Darker color indicates higher performance.}
\label{tab:main_find}
\resizebox{\linewidth}{!}{%
\begin{tabular}{lcccccccccccccccc}
\toprule
\textbf{Model} &  & \multicolumn{4}{c}{\textbf{Contextual}} & \multicolumn{4}{c}{\textbf{Behavioral}} & \multicolumn{4}{c}{\textbf{Interpersonal}} & \multicolumn{1}{c}{\textbf{All}} & \multicolumn{2}{c}{\textbf{Open Inter.}} \\
\cmidrule(lr){3-6} \cmidrule(lr){7-10} \cmidrule(lr){11-14} \cmidrule(lr){15-15} \cmidrule(lr){16-17}
 & Size & age & loc. & \shortstack{main\\act.} & Avg & \shortstack{object\\attn.} & \shortstack{shared\\attn.} & \shortstack{event\\at t} & Avg & \shortstack{react.\\(bal.)} & cause & conseq. & Avg & Avg & \shortstack{sim.\\(0-3)} & recall \\
\midrule
\multicolumn{17}{l}{\textit{Open-source}} \\
\midrule
LLaVA-NeXT-Video & 7B & \cellcolor[rgb]{0.852,0.707,0.776} 41.6 & \cellcolor[rgb]{0.718,0.605,0.712} 68.8 & \cellcolor[rgb]{0.852,0.707,0.776} 41.0 & \cellcolor[rgb]{0.812,0.669,0.757} 50.5 & \cellcolor[rgb]{0.852,0.707,0.776} 41.1 & \cellcolor[rgb]{0.899,0.766,0.803} 27.9 & \cellcolor[rgb]{0.910,0.785,0.812} 24.0 & \cellcolor[rgb]{0.890,0.754,0.797} 31.0 & \cellcolor[rgb]{0.841,0.696,0.770} 44.5 & \cellcolor[rgb]{0.861,0.717,0.780} 39.3 & \cellcolor[rgb]{0.867,0.724,0.783} 37.3 & \cellcolor[rgb]{0.857,0.712,0.778} 40.4 & \cellcolor[rgb]{0.857,0.712,0.778} 40.6 & \cellcolor[rgb]{0.912,0.787,0.813} 0.7 & \cellcolor[rgb]{0.955,0.882,0.873} 4.1 \\
LLaVA-OV & 7B & \cellcolor[rgb]{0.748,0.623,0.727} 63.6 & \cellcolor[rgb]{0.617,0.551,0.647} 87.0 & \cellcolor[rgb]{0.774,0.641,0.740} 59.0 & \cellcolor[rgb]{0.715,0.603,0.710} 69.9 & \cellcolor[rgb]{0.799,0.659,0.751} 53.6 & \cellcolor[rgb]{0.843,0.698,0.771} 43.4 & \cellcolor[rgb]{0.865,0.722,0.782} 38.2 & \cellcolor[rgb]{0.837,0.691,0.768} 45.0 & \cellcolor[rgb]{0.696,0.593,0.700} \textbf{72.8} & \cellcolor[rgb]{0.834,0.689,0.767} 46.0 & \cellcolor[rgb]{0.871,0.729,0.785} 36.0 & \cellcolor[rgb]{0.807,0.665,0.754} 51.6 & \cellcolor[rgb]{0.788,0.651,0.746} 55.5 & \cellcolor[rgb]{0.881,0.741,0.791} 1.0 & \cellcolor[rgb]{0.947,0.860,0.857} 8.8 \\
openPangu-VL & 7B & \cellcolor[rgb]{0.788,0.651,0.746} 55.9 & \cellcolor[rgb]{0.601,0.543,0.634} 89.6 & \cellcolor[rgb]{0.718,0.605,0.712} 69.4 & \cellcolor[rgb]{0.703,0.596,0.703} 71.6 & \cellcolor[rgb]{0.696,0.593,0.700} 73.2 & \cellcolor[rgb]{0.824,0.680,0.762} 47.7 & \cellcolor[rgb]{0.834,0.689,0.767} 46.0 & \cellcolor[rgb]{0.788,0.651,0.746} 55.6 & \cellcolor[rgb]{0.712,0.601,0.708} 69.9 & \cellcolor[rgb]{0.783,0.647,0.744} 56.7 & \cellcolor[rgb]{0.771,0.639,0.738} \textbf{59.3} & \cellcolor[rgb]{0.757,0.629,0.732} 62.0 & \cellcolor[rgb]{0.751,0.625,0.729} 63.1 & \cellcolor[rgb]{0.871,0.729,0.785} 1.1 & \cellcolor[rgb]{0.945,0.855,0.853} 10.0 \\
InternVL3.5 & 4B & \cellcolor[rgb]{0.924,0.808,0.824} 18.9 & \cellcolor[rgb]{0.739,0.618,0.723} 65.6 & \cellcolor[rgb]{0.871,0.729,0.785} 36.6 & \cellcolor[rgb]{0.857,0.712,0.778} 40.3 & \cellcolor[rgb]{0.824,0.680,0.762} 48.2 & \cellcolor[rgb]{0.887,0.749,0.794} 31.8 & \cellcolor[rgb]{0.928,0.816,0.829} 17.3 & \cellcolor[rgb]{0.885,0.746,0.793} 32.4 & \cellcolor[rgb]{0.885,0.746,0.793} 32.1 & \cellcolor[rgb]{0.863,0.719,0.781} 38.7 & \cellcolor[rgb]{0.902,0.771,0.805} 26.7 & \cellcolor[rgb]{0.885,0.746,0.793} 32.5 & \cellcolor[rgb]{0.877,0.736,0.789} 35.1 & \cellcolor[rgb]{0.912,0.787,0.813} 0.7 & \cellcolor[rgb]{0.951,0.871,0.864} 6.4 \\
InternVL3.5 & 8B & \cellcolor[rgb]{0.894,0.759,0.799} 29.6 & \cellcolor[rgb]{0.681,0.584,0.690} 76.0 & \cellcolor[rgb]{0.839,0.694,0.769} 44.8 & \cellcolor[rgb]{0.814,0.672,0.758} 50.1 & \cellcolor[rgb]{0.814,0.672,0.758} 50.0 & \cellcolor[rgb]{0.879,0.739,0.790} 34.1 & \cellcolor[rgb]{0.910,0.785,0.812} 24.6 & \cellcolor[rgb]{0.871,0.729,0.785} 36.2 & \cellcolor[rgb]{0.871,0.729,0.785} 36.3 & \cellcolor[rgb]{0.834,0.689,0.767} 46.0 & \cellcolor[rgb]{0.852,0.707,0.776} 41.3 & \cellcolor[rgb]{0.852,0.707,0.776} 41.2 & \cellcolor[rgb]{0.848,0.703,0.773} 42.5 & \cellcolor[rgb]{0.894,0.759,0.799} 0.9 & \cellcolor[rgb]{0.949,0.866,0.860} 7.7 \\
InternVL3.5 & 38B & \cellcolor[rgb]{0.832,0.687,0.766} 46.4 & \cellcolor[rgb]{0.630,0.558,0.656} 84.4 & \cellcolor[rgb]{0.769,0.637,0.737} 59.7 & \cellcolor[rgb]{0.748,0.623,0.727} 63.5 & \cellcolor[rgb]{0.774,0.641,0.740} 58.9 & \cellcolor[rgb]{0.820,0.676,0.760} 48.8 & \cellcolor[rgb]{0.867,0.724,0.783} 37.3 & \cellcolor[rgb]{0.824,0.680,0.762} 48.3 & \cellcolor[rgb]{0.822,0.678,0.761} 48.7 & \cellcolor[rgb]{0.783,0.647,0.744} 56.7 & \cellcolor[rgb]{0.829,0.685,0.765} 46.7 & \cellcolor[rgb]{0.812,0.669,0.757} 50.7 & \cellcolor[rgb]{0.796,0.657,0.750} 54.2 & \cellcolor[rgb]{0.897,0.764,0.802} 0.9 & \cellcolor[rgb]{0.949,0.866,0.860} 7.6 \\
MiniCPM-V 4.5 & 8B & \cellcolor[rgb]{0.690,0.589,0.696} 74.1 & \cellcolor[rgb]{0.564,0.525,0.601} \underline{96.8} & \cellcolor[rgb]{0.684,0.586,0.692} 75.4 & \cellcolor[rgb]{0.646,0.566,0.668} 82.1 & \cellcolor[rgb]{0.684,0.586,0.692} 75.0 & \cellcolor[rgb]{0.796,0.657,0.750} \underline{54.1} & \cellcolor[rgb]{0.820,0.676,0.760} 49.4 & \cellcolor[rgb]{0.769,0.637,0.737} 59.5 & \cellcolor[rgb]{0.718,0.605,0.712} 69.4 & \cellcolor[rgb]{0.757,0.629,0.732} 62.0 & \cellcolor[rgb]{0.827,0.682,0.764} 47.3 & \cellcolor[rgb]{0.769,0.637,0.737} 59.6 & \cellcolor[rgb]{0.730,0.612,0.718} 67.1 & \cellcolor[rgb]{0.885,0.746,0.793} 1.0 & \cellcolor[rgb]{0.942,0.849,0.849} 10.8 \\
Molmo2 & 8B & \cellcolor[rgb]{0.700,0.594,0.701} 72.3 & \cellcolor[rgb]{0.570,0.528,0.607} 95.5 & \cellcolor[rgb]{0.703,0.596,0.703} 71.6 & \cellcolor[rgb]{0.658,0.572,0.676} 79.8 & \cellcolor[rgb]{0.684,0.586,0.692} 75.0 & \cellcolor[rgb]{0.846,0.700,0.772} 43.1 & \cellcolor[rgb]{0.769,0.637,0.737} 59.5 & \cellcolor[rgb]{0.771,0.639,0.738} 59.2 & \cellcolor[rgb]{0.712,0.601,0.708} \underline{70.5} & \cellcolor[rgb]{0.739,0.618,0.723} 65.3 & \cellcolor[rgb]{0.802,0.661,0.752} 53.3 & \cellcolor[rgb]{0.751,0.625,0.729} \underline{63.0} & \cellcolor[rgb]{0.727,0.610,0.717} 67.4 & \cellcolor[rgb]{0.890,0.754,0.797} 0.9 & \cellcolor[rgb]{0.941,0.846,0.847} 11.6 \\
Keye-VL 1.5 & 8B & \cellcolor[rgb]{0.837,0.691,0.768} 45.4 & \cellcolor[rgb]{0.652,0.569,0.672} 81.2 & \cellcolor[rgb]{0.812,0.669,0.757} 50.7 & \cellcolor[rgb]{0.771,0.639,0.738} 59.1 & \cellcolor[rgb]{0.763,0.633,0.734} 60.7 & \cellcolor[rgb]{0.848,0.703,0.773} 42.3 & \cellcolor[rgb]{0.904,0.774,0.807} 26.3 & \cellcolor[rgb]{0.846,0.700,0.772} 43.1 & \cellcolor[rgb]{0.754,0.627,0.730} 62.8 & \cellcolor[rgb]{0.804,0.663,0.753} 52.7 & \cellcolor[rgb]{0.839,0.694,0.769} 44.7 & \cellcolor[rgb]{0.802,0.661,0.752} 53.4 & \cellcolor[rgb]{0.807,0.665,0.754} 51.9 & \cellcolor[rgb]{0.848,0.703,0.773} \underline{1.3} & \cellcolor[rgb]{0.934,0.830,0.837} 14.8 \\

Qwen3-VL & 4B & \cellcolor[rgb]{0.885,0.746,0.793} 32.4 & \cellcolor[rgb]{0.662,0.574,0.678} 79.2 & \cellcolor[rgb]{0.774,0.641,0.740} 58.2 & \cellcolor[rgb]{0.785,0.649,0.745} 56.6 & \cellcolor[rgb]{0.780,0.645,0.742} 57.1 & \cellcolor[rgb]{0.837,0.691,0.768} 45.2 & \cellcolor[rgb]{0.865,0.722,0.782} 38.2 & \cellcolor[rgb]{0.829,0.685,0.765} 46.8 & \cellcolor[rgb]{0.754,0.627,0.730} 62.2 & \cellcolor[rgb]{0.783,0.647,0.744} 56.7 & \cellcolor[rgb]{0.791,0.653,0.747} 55.3 & \cellcolor[rgb]{0.777,0.643,0.741} 58.1 & \cellcolor[rgb]{0.799,0.659,0.751} 53.8 & \cellcolor[rgb]{0.852,0.707,0.776} \underline{1.2} & \cellcolor[rgb]{0.935,0.833,0.839} 14.3 \\
Qwen3-VL & 8B & \cellcolor[rgb]{0.817,0.674,0.759} 49.8 & \cellcolor[rgb]{0.630,0.558,0.656} 84.4 & \cellcolor[rgb]{0.754,0.627,0.730} 62.7 & \cellcolor[rgb]{0.736,0.616,0.721} 65.6 & \cellcolor[rgb]{0.745,0.621,0.726} 64.3 & \cellcolor[rgb]{0.814,0.672,0.758} 50.2 & \cellcolor[rgb]{0.829,0.685,0.765} 47.1 & \cellcolor[rgb]{0.799,0.659,0.751} 53.9 & \cellcolor[rgb]{0.832,0.687,0.766} 46.1 & \cellcolor[rgb]{0.739,0.618,0.723} 65.3 & \cellcolor[rgb]{0.780,0.645,0.742} 57.3 & \cellcolor[rgb]{0.785,0.649,0.745} 56.3 & \cellcolor[rgb]{0.774,0.641,0.740} 58.6 & \cellcolor[rgb]{0.850,0.705,0.775} \underline{1.3} & \cellcolor[rgb]{0.933,0.827,0.835} \underline{15.4} \\
Qwen3-VL & 32B & \cellcolor[rgb]{0.748,0.623,0.727} 63.7 & \cellcolor[rgb]{0.608,0.547,0.639} 89.0 & \cellcolor[rgb]{0.681,0.584,0.690} 76.1 & \cellcolor[rgb]{0.677,0.582,0.688} 76.3 & \cellcolor[rgb]{0.696,0.593,0.700} 73.2 & \cellcolor[rgb]{0.827,0.682,0.764} 47.3 & \cellcolor[rgb]{0.774,0.641,0.740} 58.4 & \cellcolor[rgb]{0.769,0.637,0.737} 59.6 & \cellcolor[rgb]{0.804,0.663,0.753} 52.4 & \cellcolor[rgb]{0.718,0.605,0.712} 69.3 & \cellcolor[rgb]{0.774,0.641,0.740} \underline{58.7} & \cellcolor[rgb]{0.769,0.637,0.737} 60.1 & \cellcolor[rgb]{0.739,0.618,0.723} 65.3 & \cellcolor[rgb]{0.837,0.691,0.768} \textbf{1.4} & \cellcolor[rgb]{0.912,0.787,0.813} \textbf{23.8} \\

Qwen3.5 & 4B & \cellcolor[rgb]{0.627,0.556,0.654} 85.2 & \cellcolor[rgb]{0.573,0.529,0.610} 94.8 & \cellcolor[rgb]{0.658,0.572,0.676} \underline{79.9} & \cellcolor[rgb]{0.620,0.553,0.649} \underline{86.6} & \cellcolor[rgb]{0.684,0.586,0.692} 75.0 & \cellcolor[rgb]{0.827,0.682,0.764} 47.5 & \cellcolor[rgb]{0.760,0.631,0.733} \underline{61.6} & \cellcolor[rgb]{0.760,0.631,0.733} \underline{61.4} & \cellcolor[rgb]{0.804,0.663,0.753} 52.7 & \cellcolor[rgb]{0.706,0.598,0.705} \underline{71.3} & \cellcolor[rgb]{0.834,0.689,0.767} 46.0 & \cellcolor[rgb]{0.783,0.647,0.744} 56.7 & \cellcolor[rgb]{0.724,0.608,0.715} 68.2 & \cellcolor[rgb]{0.889,0.751,0.796} 0.9 & \cellcolor[rgb]{0.949,0.866,0.860} 7.2 \\
Qwen3.5 & 9B & \cellcolor[rgb]{0.620,0.553,0.649} \underline{86.4} & \cellcolor[rgb]{0.573,0.529,0.610} 94.8 & \cellcolor[rgb]{0.703,0.596,0.703} 71.6 & \cellcolor[rgb]{0.633,0.559,0.659} 84.3 & \cellcolor[rgb]{0.674,0.581,0.686} \underline{76.8} & \cellcolor[rgb]{0.814,0.672,0.758} 50.3 & \cellcolor[rgb]{0.788,0.651,0.746} 55.5 & \cellcolor[rgb]{0.763,0.633,0.734} 60.9 & \cellcolor[rgb]{0.712,0.601,0.708} 70.0 & \cellcolor[rgb]{0.724,0.608,0.715} 68.0 & \cellcolor[rgb]{0.812,0.669,0.757} 50.7 & \cellcolor[rgb]{0.751,0.625,0.729} 62.9 & \cellcolor[rgb]{0.718,0.605,0.712} \underline{69.3} & \cellcolor[rgb]{0.889,0.751,0.796} 0.9 & \cellcolor[rgb]{0.948,0.863,0.858} 8.0 \\
Qwen3.5 & 27B & \cellcolor[rgb]{0.601,0.543,0.634} \textbf{89.7} & \cellcolor[rgb]{0.558,0.521,0.595} \textbf{97.4} & \cellcolor[rgb]{0.646,0.566,0.668} \textbf{82.1} & \cellcolor[rgb]{0.601,0.543,0.634} \textbf{89.7} & \cellcolor[rgb]{0.614,0.550,0.644} \textbf{87.5} & \cellcolor[rgb]{0.785,0.649,0.745} \textbf{56.6} & \cellcolor[rgb]{0.754,0.627,0.730} \textbf{62.7} & \cellcolor[rgb]{0.718,0.605,0.712} \textbf{69.0} & \cellcolor[rgb]{0.715,0.603,0.710} 69.9 & \cellcolor[rgb]{0.687,0.587,0.694} \textbf{74.7} & \cellcolor[rgb]{0.783,0.647,0.744} 56.7 & \cellcolor[rgb]{0.730,0.612,0.718} \textbf{67.1} & \cellcolor[rgb]{0.684,0.586,0.692} \textbf{75.3} & \cellcolor[rgb]{0.859,0.714,0.779} 1.2 & \cellcolor[rgb]{0.936,0.835,0.840} 14.1 \\
\bottomrule
\end{tabular}
}
\end{table}

\subsection{Blind experiment}

To verify that \bench requires visual understanding rather than language priors, we evaluate all models on question text alone, with no video or keyframe (Tables~\ref{tab:blind} and \ref{tab:blind_find}). Performance drops substantially across all categories, providing evidence that \bench measures genuine visual social reasoning rather than language-only shortcuts. 

\begin{table}[th]
\caption{\textbf{Model performance on the public subsets of \bench without visual input}. Open interactions are measured in similarity to ground truth (0-3) and recall (\%), all other tasks are in accuracy (\%). \textbf{Best} model and \underline{second best} model. Darker color indicates higher performance.}
\label{tab:blind}
\resizebox{\linewidth}{!}{%
\begin{tabular}{lcccccccccccccccc}
\toprule
\textbf{Model} &  & \multicolumn{4}{c}{\textbf{Contextual}} & \multicolumn{4}{c}{\textbf{Behavioral}} & \multicolumn{4}{c}{\textbf{Interpersonal}} & \multicolumn{1}{c}{\textbf{All}} & \multicolumn{2}{c}{\textbf{Open Inter.}} \\
\cmidrule(lr){3-6} \cmidrule(lr){7-10} \cmidrule(lr){11-14} \cmidrule(lr){15-15} \cmidrule(lr){16-17}
 & Size & age & loc. & \shortstack{main\\act.} & Avg & \shortstack{object\\attn.} & \shortstack{shared\\attn.} & \shortstack{event\\at t} & Avg & \shortstack{react.\\(bal.)} & cause & conseq. & Avg & Avg & \shortstack{sim.\\(0-3)} & recall \\
\midrule
\multicolumn{17}{l}{\textit{Open-source}} \\
\midrule
LLaVA-NeXT-Video & 7B & \cellcolor[rgb]{0.837,0.691,0.768} 45.4 & \cellcolor[rgb]{0.909,0.782,0.810} 24.7 & \cellcolor[rgb]{0.918,0.798,0.819} 21.3 & \cellcolor[rgb]{0.890,0.754,0.797} 30.5 & \cellcolor[rgb]{0.921,0.803,0.821} 20.5 & \cellcolor[rgb]{0.873,0.731,0.787} 35.9 & \cellcolor[rgb]{0.929,0.819,0.831} 17.0 & \cellcolor[rgb]{0.910,0.785,0.812} 24.4 & \cellcolor[rgb]{0.839,0.694,0.769} 44.6 & \cellcolor[rgb]{0.863,0.719,0.781} 38.7 & \cellcolor[rgb]{0.881,0.741,0.791} 33.9 & \cellcolor[rgb]{0.861,0.717,0.780} 39.1 & \cellcolor[rgb]{0.889,0.751,0.796} 31.3 & \cellcolor[rgb]{0.919,0.800,0.820} 0.6 & \cellcolor[rgb]{0.959,0.893,0.882} 2.3 \\
LLaVA-OV & 7B & \cellcolor[rgb]{0.841,0.696,0.770} 43.8 & \cellcolor[rgb]{0.873,0.731,0.787} 35.7 & \cellcolor[rgb]{0.877,0.736,0.789} 35.0 & \cellcolor[rgb]{0.865,0.722,0.782} 38.2 & \cellcolor[rgb]{0.867,0.724,0.783} 37.3 & \cellcolor[rgb]{0.899,0.766,0.803} 27.9 & \cellcolor[rgb]{0.915,0.792,0.816} 22.4 & \cellcolor[rgb]{0.896,0.761,0.800} 29.2 & \cellcolor[rgb]{0.807,0.665,0.754} 52.2 & \cellcolor[rgb]{0.829,0.685,0.765} 47.0 & \cellcolor[rgb]{0.881,0.741,0.791} 33.8 & \cellcolor[rgb]{0.841,0.696,0.770} 44.3 & \cellcolor[rgb]{0.867,0.724,0.783} 37.2 & \cellcolor[rgb]{0.926,0.814,0.827} 0.5 & \cellcolor[rgb]{0.957,0.887,0.877} \underline{2.8} \\
openPangu-VL & 7B & \cellcolor[rgb]{0.837,0.691,0.768} 45.5 & \cellcolor[rgb]{0.774,0.641,0.740} 58.4 & \cellcolor[rgb]{0.863,0.719,0.781} 38.4 & \cellcolor[rgb]{0.827,0.682,0.764} 47.5 & \cellcolor[rgb]{0.859,0.714,0.779} 39.6 & \cellcolor[rgb]{0.885,0.746,0.793} 32.6 & \cellcolor[rgb]{0.910,0.785,0.812} 23.9 & \cellcolor[rgb]{0.885,0.746,0.793} 32.0 & \cellcolor[rgb]{0.928,0.816,0.829} 17.3 & \cellcolor[rgb]{0.834,0.689,0.767} 45.8 & \cellcolor[rgb]{0.846,0.700,0.772} 43.1 & \cellcolor[rgb]{0.875,0.734,0.788} 35.4 & \cellcolor[rgb]{0.863,0.719,0.781} 38.3 & \cellcolor[rgb]{0.907,0.779,0.809} 0.8 & \cellcolor[rgb]{0.960,0.896,0.884} 1.2 \\
InternVL3.5 & 4B & \cellcolor[rgb]{0.846,0.700,0.772} 43.1 & \cellcolor[rgb]{0.812,0.669,0.757} 50.5 & \cellcolor[rgb]{0.867,0.724,0.783} 37.3 & \cellcolor[rgb]{0.843,0.698,0.771} 43.6 & \cellcolor[rgb]{0.877,0.736,0.789} 35.1 & \cellcolor[rgb]{0.896,0.761,0.800} 28.9 & \cellcolor[rgb]{0.930,0.822,0.832} 16.1 & \cellcolor[rgb]{0.902,0.771,0.805} 26.7 & \cellcolor[rgb]{0.802,0.661,0.752} \underline{52.9} & \cellcolor[rgb]{0.857,0.712,0.778} 40.1 & \cellcolor[rgb]{0.859,0.714,0.779} 39.6 & \cellcolor[rgb]{0.841,0.696,0.770} 44.2 & \cellcolor[rgb]{0.865,0.722,0.782} 38.2 & \cellcolor[rgb]{0.890,0.754,0.797} \textbf{0.9} & \cellcolor[rgb]{0.961,0.898,0.886} 0.5 \\
InternVL3.5 & 8B & \cellcolor[rgb]{0.843,0.698,0.771} 43.6 & \cellcolor[rgb]{0.807,0.665,0.754} 51.8 & \cellcolor[rgb]{0.877,0.736,0.789} 35.0 & \cellcolor[rgb]{0.843,0.698,0.771} 43.5 & \cellcolor[rgb]{0.913,0.790,0.814} 23.3 & \cellcolor[rgb]{0.865,0.722,0.782} 38.3 & \cellcolor[rgb]{0.935,0.833,0.839} 14.1 & \cellcolor[rgb]{0.907,0.779,0.809} 25.2 & \cellcolor[rgb]{0.877,0.736,0.789} 35.0 & \cellcolor[rgb]{0.824,0.680,0.762} 47.9 & \cellcolor[rgb]{0.846,0.700,0.772} 43.3 & \cellcolor[rgb]{0.850,0.705,0.775} 42.1 & \cellcolor[rgb]{0.869,0.726,0.784} 36.9 & \cellcolor[rgb]{0.890,0.754,0.797} \textbf{0.9} & \cellcolor[rgb]{0.961,0.898,0.886} 0.6 \\
InternVL3.5 & 38B & \cellcolor[rgb]{0.848,0.703,0.773} 42.6 & \cellcolor[rgb]{0.827,0.682,0.764} 47.5 & \cellcolor[rgb]{0.875,0.734,0.788} 35.2 & \cellcolor[rgb]{0.852,0.707,0.776} 41.8 & \cellcolor[rgb]{0.871,0.729,0.785} 36.1 & \cellcolor[rgb]{0.807,0.665,0.754} \underline{52.1} & \cellcolor[rgb]{0.916,0.795,0.817} 22.1 & \cellcolor[rgb]{0.869,0.726,0.784} 36.7 & \cellcolor[rgb]{0.875,0.734,0.788} 35.5 & \cellcolor[rgb]{0.802,0.661,0.752} 53.2 & \cellcolor[rgb]{0.852,0.707,0.776} 41.2 & \cellcolor[rgb]{0.846,0.700,0.772} 43.3 & \cellcolor[rgb]{0.857,0.712,0.778} 40.6 & \cellcolor[rgb]{0.890,0.754,0.797} \textbf{0.9} & \cellcolor[rgb]{0.961,0.898,0.886} 0.6 \\

MiniCPM-V 4.5 & 8B & \cellcolor[rgb]{0.863,0.719,0.781} 38.5 & \cellcolor[rgb]{0.791,0.653,0.747} 55.4 & \cellcolor[rgb]{0.865,0.722,0.782} 38.0 & \cellcolor[rgb]{0.841,0.696,0.770} 43.9 & \cellcolor[rgb]{0.824,0.680,0.762} \underline{47.7} & \cellcolor[rgb]{0.852,0.707,0.776} 41.2 & \cellcolor[rgb]{0.901,0.769,0.804} 27.6 & \cellcolor[rgb]{0.863,0.719,0.781} \underline{38.8} & \cellcolor[rgb]{0.913,0.790,0.814} 23.2 & \cellcolor[rgb]{0.820,0.676,0.760} 48.9 & \cellcolor[rgb]{0.863,0.719,0.781} 38.8 & \cellcolor[rgb]{0.869,0.726,0.784} 36.9 & \cellcolor[rgb]{0.857,0.712,0.778} 39.9 & \cellcolor[rgb]{0.925,0.811,0.826} 0.6 & \cellcolor[rgb]{0.959,0.893,0.882} 1.8 \\
Molmo2 & 8B & \cellcolor[rgb]{0.839,0.694,0.769} 44.7 & \cellcolor[rgb]{0.834,0.689,0.767} 45.7 & \cellcolor[rgb]{0.867,0.724,0.783} 37.4 & \cellcolor[rgb]{0.848,0.703,0.773} 42.6 & \cellcolor[rgb]{0.906,0.777,0.808} 25.4 & \cellcolor[rgb]{0.809,0.667,0.756} 51.2 & \cellcolor[rgb]{0.892,0.756,0.798} 30.2 & \cellcolor[rgb]{0.873,0.731,0.787} 35.6 & \cellcolor[rgb]{0.892,0.756,0.798} 30.1 & \cellcolor[rgb]{0.796,0.657,0.750} 54.0 & \cellcolor[rgb]{0.804,0.663,0.753} \textbf{52.4} & \cellcolor[rgb]{0.837,0.691,0.768} 45.5 & \cellcolor[rgb]{0.852,0.707,0.776} 41.2 & \cellcolor[rgb]{0.929,0.819,0.831} 0.5 & \cellcolor[rgb]{0.959,0.893,0.882} 1.7 \\
Keye-VL 1.5 & 8B & \cellcolor[rgb]{0.834,0.689,0.767} 45.7 & \cellcolor[rgb]{0.788,0.651,0.746} 55.7 & \cellcolor[rgb]{0.881,0.741,0.791} 33.8 & \cellcolor[rgb]{0.837,0.691,0.768} 45.1 & \cellcolor[rgb]{0.881,0.741,0.791} 33.3 & \cellcolor[rgb]{0.791,0.653,0.747} \textbf{55.3} & \cellcolor[rgb]{0.926,0.814,0.827} 17.6 & \cellcolor[rgb]{0.875,0.734,0.788} 35.4 & \cellcolor[rgb]{0.748,0.623,0.727} \textbf{64.0} & \cellcolor[rgb]{0.829,0.685,0.765} 46.7 & \cellcolor[rgb]{0.879,0.739,0.790} 34.3 & \cellcolor[rgb]{0.824,0.680,0.762} \textbf{48.3} & \cellcolor[rgb]{0.848,0.703,0.773} 42.9 & \cellcolor[rgb]{0.924,0.808,0.824} 0.6 & \cellcolor[rgb]{0.961,0.898,0.886} 1.1 \\
Qwen3-VL & 4B & \cellcolor[rgb]{0.837,0.691,0.768} 45.0 & \cellcolor[rgb]{0.820,0.676,0.760} 49.5 & \cellcolor[rgb]{0.871,0.729,0.785} 36.6 & \cellcolor[rgb]{0.843,0.698,0.771} 43.7 & \cellcolor[rgb]{0.890,0.754,0.797} 30.6 & \cellcolor[rgb]{0.883,0.744,0.792} 33.0 & \cellcolor[rgb]{0.910,0.785,0.812} 24.1 & \cellcolor[rgb]{0.896,0.761,0.800} 29.2 & \cellcolor[rgb]{0.820,0.676,0.760} 49.0 & \cellcolor[rgb]{0.820,0.676,0.760} 49.2 & \cellcolor[rgb]{0.832,0.687,0.766} 46.4 & \cellcolor[rgb]{0.824,0.680,0.762} \underline{48.2} & \cellcolor[rgb]{0.857,0.712,0.778} 40.4 & \cellcolor[rgb]{0.926,0.814,0.827} 0.5 & \cellcolor[rgb]{0.962,0.901,0.888} 0.3 \\
Qwen3-VL & 8B & \cellcolor[rgb]{0.841,0.696,0.770} 44.5 & \cellcolor[rgb]{0.824,0.680,0.762} 48.4 & \cellcolor[rgb]{0.885,0.746,0.793} 32.3 & \cellcolor[rgb]{0.852,0.707,0.776} 41.7 & \cellcolor[rgb]{0.863,0.719,0.781} 39.1 & \cellcolor[rgb]{0.910,0.785,0.812} 24.3 & \cellcolor[rgb]{0.906,0.777,0.808} 25.7 & \cellcolor[rgb]{0.894,0.759,0.799} 29.7 & \cellcolor[rgb]{0.916,0.795,0.817} 21.5 & \cellcolor[rgb]{0.817,0.674,0.759} 49.9 & \cellcolor[rgb]{0.817,0.674,0.759} \underline{49.7} & \cellcolor[rgb]{0.857,0.712,0.778} 40.4 & \cellcolor[rgb]{0.867,0.724,0.783} 37.3 & \cellcolor[rgb]{0.918,0.798,0.819} 0.6 & \cellcolor[rgb]{0.961,0.898,0.886} 0.9 \\
Qwen3-VL & 32B & \cellcolor[rgb]{0.829,0.685,0.765} 46.8 & \cellcolor[rgb]{0.769,0.637,0.737} \underline{60.2} & \cellcolor[rgb]{0.867,0.724,0.783} 37.9 & \cellcolor[rgb]{0.824,0.680,0.762} 48.3 & \cellcolor[rgb]{0.852,0.707,0.776} 41.4 & \cellcolor[rgb]{0.894,0.759,0.799} 29.4 & \cellcolor[rgb]{0.896,0.761,0.800} 29.1 & \cellcolor[rgb]{0.881,0.741,0.791} 33.3 & \cellcolor[rgb]{0.910,0.785,0.812} 24.2 & \cellcolor[rgb]{0.785,0.649,0.745} 56.3 & \cellcolor[rgb]{0.837,0.691,0.768} 45.0 & \cellcolor[rgb]{0.850,0.705,0.775} 41.9 & \cellcolor[rgb]{0.852,0.707,0.776} 41.1 & \cellcolor[rgb]{0.910,0.785,0.812} 0.7 & \cellcolor[rgb]{0.960,0.896,0.884} 1.4 \\
Qwen3.5 & 4B & \cellcolor[rgb]{0.839,0.694,0.769} 44.8 & \cellcolor[rgb]{0.802,0.661,0.752} 52.9 & \cellcolor[rgb]{0.892,0.756,0.798} 30.4 & \cellcolor[rgb]{0.848,0.703,0.773} 42.7 & \cellcolor[rgb]{0.894,0.759,0.799} 30.0 & \cellcolor[rgb]{0.896,0.761,0.800} 29.3 & \cellcolor[rgb]{0.926,0.814,0.827} 18.3 & \cellcolor[rgb]{0.906,0.777,0.808} 25.8 & \cellcolor[rgb]{0.910,0.785,0.812} 24.4 & \cellcolor[rgb]{0.841,0.696,0.770} 44.2 & \cellcolor[rgb]{0.899,0.766,0.803} 28.5 & \cellcolor[rgb]{0.885,0.746,0.793} 32.4 & \cellcolor[rgb]{0.881,0.741,0.791} 33.6 & \cellcolor[rgb]{0.916,0.795,0.817} 0.7 & \cellcolor[rgb]{0.961,0.898,0.886} 0.4 \\
Qwen3.5 & 9B & \cellcolor[rgb]{0.837,0.691,0.768} 45.5 & \cellcolor[rgb]{0.769,0.637,0.737} 59.9 & \cellcolor[rgb]{0.867,0.724,0.783} 37.8 & \cellcolor[rgb]{0.824,0.680,0.762} 47.7 & \cellcolor[rgb]{0.857,0.712,0.778} 40.4 & \cellcolor[rgb]{0.899,0.766,0.803} 28.5 & \cellcolor[rgb]{0.916,0.795,0.817} 22.1 & \cellcolor[rgb]{0.892,0.756,0.798} 30.3 & \cellcolor[rgb]{0.863,0.719,0.781} 38.5 & \cellcolor[rgb]{0.804,0.663,0.753} 52.4 & \cellcolor[rgb]{0.867,0.724,0.783} 37.8 & \cellcolor[rgb]{0.848,0.703,0.773} 42.9 & \cellcolor[rgb]{0.857,0.712,0.778} 40.3 & \cellcolor[rgb]{0.960,0.896,0.884} 0.0 & \cellcolor[rgb]{0.962,0.901,0.888} 0.0 \\
Qwen3.5 & 27B & \cellcolor[rgb]{0.829,0.685,0.765} 47.3 & \cellcolor[rgb]{0.709,0.600,0.707} \textbf{71.0} & \cellcolor[rgb]{0.832,0.687,0.766} \textbf{46.2} & \cellcolor[rgb]{0.794,0.655,0.749} \textbf{54.8} & \cellcolor[rgb]{0.824,0.680,0.762} \textbf{48.3} & \cellcolor[rgb]{0.906,0.777,0.808} 26.1 & \cellcolor[rgb]{0.889,0.751,0.796} 31.6 & \cellcolor[rgb]{0.875,0.734,0.788} 35.4 & \cellcolor[rgb]{0.913,0.790,0.814} 23.2 & \cellcolor[rgb]{0.766,0.635,0.736} \textbf{60.5} & \cellcolor[rgb]{0.848,0.703,0.773} 42.3 & \cellcolor[rgb]{0.850,0.705,0.775} 42.0 & \cellcolor[rgb]{0.841,0.696,0.770} \underline{44.1} & \cellcolor[rgb]{0.916,0.795,0.817} 0.7 & \cellcolor[rgb]{0.960,0.896,0.884} 1.5 \\
\midrule
\multicolumn{17}{l}{\textit{Closed-source (API)}} \\
\midrule
Gemini 3.1 Flash Lite & - & \cellcolor[rgb]{0.820,0.676,0.760} \textbf{49.2} & \cellcolor[rgb]{0.799,0.659,0.751} 53.6 & \cellcolor[rgb]{0.841,0.696,0.770} \underline{43.9} & \cellcolor[rgb]{0.820,0.676,0.760} \underline{48.9} & \cellcolor[rgb]{0.837,0.691,0.768} 45.2 & \cellcolor[rgb]{0.822,0.678,0.761} 48.7 & \cellcolor[rgb]{0.869,0.726,0.784} \textbf{36.7} & \cellcolor[rgb]{0.843,0.698,0.771} \textbf{43.5} & \cellcolor[rgb]{0.883,0.744,0.792} 32.8 & \cellcolor[rgb]{0.774,0.641,0.740} \underline{58.9} & \cellcolor[rgb]{0.834,0.689,0.767} 45.8 & \cellcolor[rgb]{0.834,0.689,0.767} 45.8 & \cellcolor[rgb]{0.834,0.689,0.767} \textbf{46.1} & \cellcolor[rgb]{0.946,0.857,0.855} 0.3 & \cellcolor[rgb]{0.961,0.898,0.886} 0.6 \\
GPT-5 mini & - & \cellcolor[rgb]{0.827,0.682,0.764} \underline{47.4} & \cellcolor[rgb]{0.788,0.651,0.746} 56.1 & \cellcolor[rgb]{0.855,0.710,0.777} 40.9 & \cellcolor[rgb]{0.824,0.680,0.762} 48.1 & \cellcolor[rgb]{0.841,0.696,0.770} 44.4 & \cellcolor[rgb]{0.918,0.798,0.819} 21.2 & \cellcolor[rgb]{0.883,0.744,0.792} \underline{32.9} & \cellcolor[rgb]{0.883,0.744,0.792} 32.8 & \cellcolor[rgb]{0.909,0.782,0.810} 24.9 & \cellcolor[rgb]{0.774,0.641,0.740} 58.2 & \cellcolor[rgb]{0.824,0.680,0.762} 48.3 & \cellcolor[rgb]{0.841,0.696,0.770} 43.8 & \cellcolor[rgb]{0.852,0.707,0.776} 41.6 & \cellcolor[rgb]{0.902,0.771,0.805} 0.8 & \cellcolor[rgb]{0.952,0.874,0.866} \textbf{5.8} \\
\bottomrule
\end{tabular}
}
\end{table}

\begin{table}[th]
\caption{\textbf{Model performance on the \FIND subset of \bench without visual input}. Open interactions are measured in similarity to ground truth (0-3) and recall (\%), all other tasks are in accuracy (\%). \textbf{Best} model and \underline{second best} model. Darker color indicates higher performance.}
\label{tab:blind_find}
\resizebox{\linewidth}{!}{%
\begin{tabular}{lcccccccccccccccc}
\toprule
\textbf{Model} &  & \multicolumn{4}{c}{\textbf{Contextual}} & \multicolumn{4}{c}{\textbf{Behavioral}} & \multicolumn{4}{c}{\textbf{Interpersonal}} & \multicolumn{1}{c}{\textbf{All}} & \multicolumn{2}{c}{\textbf{Open Inter.}} \\
\cmidrule(lr){3-6} \cmidrule(lr){7-10} \cmidrule(lr){11-14} \cmidrule(lr){15-15} \cmidrule(lr){16-17}
 & Size & age & loc. & \shortstack{main\\act.} & Avg & \shortstack{object\\attn.} & \shortstack{shared\\attn.} & \shortstack{event\\at t} & Avg & \shortstack{react.\\(bal.)} & cause & conseq. & Avg & Avg & \shortstack{sim.\\(0-3)} & recall \\
\midrule
\multicolumn{17}{l}{\textit{Open-source}} \\
\midrule
LLaVA-NeXT-Video & 7B & \cellcolor[rgb]{0.837,0.691,0.768} \textbf{45.5} & \cellcolor[rgb]{0.896,0.761,0.800} 29.2 & \cellcolor[rgb]{0.921,0.803,0.821} 20.1 & \cellcolor[rgb]{0.889,0.751,0.796} 31.6 & \cellcolor[rgb]{0.918,0.798,0.819} 21.4 & \cellcolor[rgb]{0.894,0.759,0.799} 29.8 & \cellcolor[rgb]{0.926,0.814,0.827} 18.2 & \cellcolor[rgb]{0.913,0.790,0.814} 23.1 & \cellcolor[rgb]{0.829,0.685,0.765} \underline{46.9} & \cellcolor[rgb]{0.867,0.724,0.783} 37.3 & \cellcolor[rgb]{0.875,0.734,0.788} 35.3 & \cellcolor[rgb]{0.857,0.712,0.778} 39.9 & \cellcolor[rgb]{0.889,0.751,0.796} 31.5 & \cellcolor[rgb]{0.924,0.808,0.824} 0.6 & \cellcolor[rgb]{0.960,0.896,0.884} 1.4 \\
LLaVA-OV & 7B & \cellcolor[rgb]{0.841,0.696,0.770} 43.8 & \cellcolor[rgb]{0.877,0.736,0.789} 35.1 & \cellcolor[rgb]{0.889,0.751,0.796} 31.3 & \cellcolor[rgb]{0.869,0.726,0.784} 36.7 & \cellcolor[rgb]{0.881,0.741,0.791} 33.9 & \cellcolor[rgb]{0.881,0.741,0.791} 33.8 & \cellcolor[rgb]{0.918,0.798,0.819} 21.1 & \cellcolor[rgb]{0.894,0.759,0.799} 29.6 & \cellcolor[rgb]{0.829,0.685,0.765} 46.5 & \cellcolor[rgb]{0.834,0.689,0.767} 46.0 & \cellcolor[rgb]{0.875,0.734,0.788} 35.3 & \cellcolor[rgb]{0.848,0.703,0.773} 42.6 & \cellcolor[rgb]{0.871,0.729,0.785} 36.3 & \cellcolor[rgb]{0.934,0.830,0.837} 0.4 & \cellcolor[rgb]{0.959,0.893,0.882} \underline{1.9} \\
openPangu-VL & 7B & \cellcolor[rgb]{0.839,0.694,0.769} 44.7 & \cellcolor[rgb]{0.877,0.736,0.789} 35.1 & \cellcolor[rgb]{0.883,0.744,0.792} 32.8 & \cellcolor[rgb]{0.867,0.724,0.783} 37.5 & \cellcolor[rgb]{0.839,0.694,0.769} 44.6 & \cellcolor[rgb]{0.867,0.724,0.783} 37.7 & \cellcolor[rgb]{0.912,0.787,0.813} 23.7 & \cellcolor[rgb]{0.875,0.734,0.788} 35.4 & \cellcolor[rgb]{0.930,0.822,0.832} 16.3 & \cellcolor[rgb]{0.837,0.691,0.768} 45.3 & \cellcolor[rgb]{0.841,0.696,0.770} 44.0 & \cellcolor[rgb]{0.875,0.734,0.788} 35.2 & \cellcolor[rgb]{0.871,0.729,0.785} 36.0 & \cellcolor[rgb]{0.894,0.759,0.799} \textbf{0.9} & \cellcolor[rgb]{0.962,0.901,0.888} 0.1 \\
InternVL3.5 & 4B & \cellcolor[rgb]{0.852,0.707,0.776} 41.3 & \cellcolor[rgb]{0.873,0.731,0.787} 35.7 & \cellcolor[rgb]{0.885,0.746,0.793} 32.1 & \cellcolor[rgb]{0.871,0.729,0.785} 36.4 & \cellcolor[rgb]{0.892,0.756,0.798} 30.4 & \cellcolor[rgb]{0.887,0.749,0.794} 31.8 & \cellcolor[rgb]{0.936,0.835,0.840} 13.9 & \cellcolor[rgb]{0.907,0.779,0.809} 25.4 & \cellcolor[rgb]{0.834,0.689,0.767} 45.8 & \cellcolor[rgb]{0.837,0.691,0.768} 45.3 & \cellcolor[rgb]{0.881,0.741,0.791} 33.3 & \cellcolor[rgb]{0.852,0.707,0.776} 41.5 & \cellcolor[rgb]{0.877,0.736,0.789} 34.4 & \cellcolor[rgb]{0.897,0.764,0.802} \textbf{0.9} & \cellcolor[rgb]{0.962,0.901,0.888} 0.0 \\
InternVL3.5 & 8B & \cellcolor[rgb]{0.848,0.703,0.773} 42.8 & \cellcolor[rgb]{0.822,0.678,0.761} 48.7 & \cellcolor[rgb]{0.885,0.746,0.793} 32.1 & \cellcolor[rgb]{0.852,0.707,0.776} 41.2 & \cellcolor[rgb]{0.892,0.756,0.798} 30.4 & \cellcolor[rgb]{0.873,0.731,0.787} 35.8 & \cellcolor[rgb]{0.941,0.846,0.847} 11.6 & \cellcolor[rgb]{0.906,0.777,0.808} 25.9 & \cellcolor[rgb]{0.883,0.744,0.792} 32.9 & \cellcolor[rgb]{0.846,0.700,0.772} 43.3 & \cellcolor[rgb]{0.837,0.691,0.768} 45.3 & \cellcolor[rgb]{0.857,0.712,0.778} 40.5 & \cellcolor[rgb]{0.873,0.731,0.787} 35.9 & \cellcolor[rgb]{0.892,0.756,0.798} \textbf{0.9} & \cellcolor[rgb]{0.962,0.901,0.888} 0.1 \\
InternVL3.5 & 38B & \cellcolor[rgb]{0.852,0.707,0.776} 41.2 & \cellcolor[rgb]{0.857,0.712,0.778} 40.3 & \cellcolor[rgb]{0.867,0.724,0.783} 37.3 & \cellcolor[rgb]{0.859,0.714,0.779} 39.6 & \cellcolor[rgb]{0.848,0.703,0.773} 42.9 & \cellcolor[rgb]{0.852,0.707,0.776} \textbf{41.0} & \cellcolor[rgb]{0.921,0.803,0.821} 19.9 & \cellcolor[rgb]{0.877,0.736,0.789} 34.6 & \cellcolor[rgb]{0.865,0.722,0.782} 38.0 & \cellcolor[rgb]{0.812,0.669,0.757} 50.7 & \cellcolor[rgb]{0.841,0.696,0.770} 44.0 & \cellcolor[rgb]{0.841,0.696,0.770} 44.2 & \cellcolor[rgb]{0.859,0.714,0.779} 39.5 & \cellcolor[rgb]{0.894,0.759,0.799} \textbf{0.9} & \cellcolor[rgb]{0.962,0.901,0.888} 0.4 \\

MiniCPM-V 4.5 & 8B & \cellcolor[rgb]{0.875,0.734,0.788} 35.3 & \cellcolor[rgb]{0.837,0.691,0.768} 45.5 & \cellcolor[rgb]{0.877,0.736,0.789} 35.1 & \cellcolor[rgb]{0.863,0.719,0.781} 38.6 & \cellcolor[rgb]{0.839,0.694,0.769} 44.6 & \cellcolor[rgb]{0.865,0.722,0.782} 38.0 & \cellcolor[rgb]{0.899,0.766,0.803} 27.7 & \cellcolor[rgb]{0.869,0.726,0.784} \underline{36.8} & \cellcolor[rgb]{0.906,0.777,0.808} 25.4 & \cellcolor[rgb]{0.814,0.672,0.758} 50.0 & \cellcolor[rgb]{0.867,0.724,0.783} 37.3 & \cellcolor[rgb]{0.867,0.724,0.783} 37.6 & \cellcolor[rgb]{0.867,0.724,0.783} 37.7 & \cellcolor[rgb]{0.926,0.814,0.827} 0.5 & \cellcolor[rgb]{0.959,0.893,0.882} \textbf{2.3} \\
Molmo2 & 8B & \cellcolor[rgb]{0.848,0.703,0.773} 42.3 & \cellcolor[rgb]{0.841,0.696,0.770} 44.2 & \cellcolor[rgb]{0.883,0.744,0.792} 32.8 & \cellcolor[rgb]{0.859,0.714,0.779} 39.8 & \cellcolor[rgb]{0.902,0.771,0.805} 26.8 & \cellcolor[rgb]{0.859,0.714,0.779} 39.6 & \cellcolor[rgb]{0.881,0.741,0.791} \textbf{33.2} & \cellcolor[rgb]{0.881,0.741,0.791} 33.2 & \cellcolor[rgb]{0.896,0.761,0.800} 29.0 & \cellcolor[rgb]{0.769,0.637,0.737} \textbf{60.0} & \cellcolor[rgb]{0.794,0.655,0.749} \textbf{54.7} & \cellcolor[rgb]{0.824,0.680,0.762} \textbf{47.9} & \cellcolor[rgb]{0.857,0.712,0.778} 40.3 & \cellcolor[rgb]{0.916,0.795,0.817} 0.6 & \cellcolor[rgb]{0.961,0.898,0.886} 0.5 \\
Keye-VL 1.5 & 8B & \cellcolor[rgb]{0.841,0.696,0.770} 43.8 & \cellcolor[rgb]{0.832,0.687,0.766} 46.1 & \cellcolor[rgb]{0.863,0.719,0.781} 38.8 & \cellcolor[rgb]{0.848,0.703,0.773} 42.9 & \cellcolor[rgb]{0.881,0.741,0.791} 33.9 & \cellcolor[rgb]{0.855,0.710,0.777} \underline{41.0} & \cellcolor[rgb]{0.921,0.803,0.821} 20.5 & \cellcolor[rgb]{0.887,0.749,0.794} 31.8 & \cellcolor[rgb]{0.802,0.661,0.752} \textbf{52.9} & \cellcolor[rgb]{0.834,0.689,0.767} 46.0 & \cellcolor[rgb]{0.850,0.705,0.775} 42.0 & \cellcolor[rgb]{0.829,0.685,0.765} \underline{47.0} & \cellcolor[rgb]{0.857,0.712,0.778} 40.6 & \cellcolor[rgb]{0.912,0.787,0.813} 0.7 & \cellcolor[rgb]{0.961,0.898,0.886} 0.6 \\
Qwen3-VL & 4B & \cellcolor[rgb]{0.843,0.698,0.771} 43.6 & \cellcolor[rgb]{0.841,0.696,0.770} 44.2 & \cellcolor[rgb]{0.890,0.754,0.797} 30.6 & \cellcolor[rgb]{0.861,0.717,0.780} 39.4 & \cellcolor[rgb]{0.892,0.756,0.798} 30.4 & \cellcolor[rgb]{0.883,0.744,0.792} 32.9 & \cellcolor[rgb]{0.918,0.798,0.819} 21.1 & \cellcolor[rgb]{0.899,0.766,0.803} 28.1 & \cellcolor[rgb]{0.857,0.712,0.778} 39.9 & \cellcolor[rgb]{0.829,0.685,0.765} 46.7 & \cellcolor[rgb]{0.814,0.672,0.758} 50.0 & \cellcolor[rgb]{0.837,0.691,0.768} 45.5 & \cellcolor[rgb]{0.867,0.724,0.783} 37.7 & \cellcolor[rgb]{0.916,0.795,0.817} 0.7 & \cellcolor[rgb]{0.962,0.901,0.888} 0.1 \\
Qwen3-VL & 8B & \cellcolor[rgb]{0.852,0.707,0.776} 41.5 & \cellcolor[rgb]{0.841,0.696,0.770} 44.2 & \cellcolor[rgb]{0.873,0.731,0.787} 35.8 & \cellcolor[rgb]{0.857,0.712,0.778} 40.5 & \cellcolor[rgb]{0.885,0.746,0.793} 32.1 & \cellcolor[rgb]{0.879,0.739,0.790} 34.3 & \cellcolor[rgb]{0.902,0.771,0.805} 26.6 & \cellcolor[rgb]{0.890,0.754,0.797} 31.0 & \cellcolor[rgb]{0.909,0.782,0.810} 25.0 & \cellcolor[rgb]{0.809,0.667,0.756} 51.3 & \cellcolor[rgb]{0.794,0.655,0.749} \textbf{54.7} & \cellcolor[rgb]{0.843,0.698,0.771} 43.7 & \cellcolor[rgb]{0.863,0.719,0.781} 38.4 & \cellcolor[rgb]{0.910,0.785,0.812} 0.7 & \cellcolor[rgb]{0.962,0.901,0.888} 0.0 \\
Qwen3-VL & 32B & \cellcolor[rgb]{0.837,0.691,0.768} \underline{45.1} & \cellcolor[rgb]{0.855,0.710,0.777} 40.9 & \cellcolor[rgb]{0.852,0.707,0.776} \underline{41.8} & \cellcolor[rgb]{0.848,0.703,0.773} 42.6 & \cellcolor[rgb]{0.824,0.680,0.762} \textbf{48.2} & \cellcolor[rgb]{0.887,0.749,0.794} 31.8 & \cellcolor[rgb]{0.897,0.764,0.802} 28.9 & \cellcolor[rgb]{0.871,0.729,0.785} 36.3 & \cellcolor[rgb]{0.904,0.774,0.807} 26.5 & \cellcolor[rgb]{0.802,0.661,0.752} 53.3 & \cellcolor[rgb]{0.807,0.665,0.754} 52.0 & \cellcolor[rgb]{0.841,0.696,0.770} 44.0 & \cellcolor[rgb]{0.855,0.710,0.777} \underline{41.0} & \cellcolor[rgb]{0.899,0.766,0.803} 0.8 & \cellcolor[rgb]{0.961,0.898,0.886} 0.8 \\

Qwen3.5 & 4B & \cellcolor[rgb]{0.846,0.700,0.772} 43.0 & \cellcolor[rgb]{0.881,0.741,0.791} 33.8 & \cellcolor[rgb]{0.890,0.754,0.797} 30.6 & \cellcolor[rgb]{0.873,0.731,0.787} 35.8 & \cellcolor[rgb]{0.892,0.756,0.798} 30.4 & \cellcolor[rgb]{0.887,0.749,0.794} 31.8 & \cellcolor[rgb]{0.921,0.803,0.821} 20.5 & \cellcolor[rgb]{0.901,0.769,0.804} 27.6 & \cellcolor[rgb]{0.902,0.771,0.805} 26.7 & \cellcolor[rgb]{0.846,0.700,0.772} 43.3 & \cellcolor[rgb]{0.909,0.782,0.810} 24.7 & \cellcolor[rgb]{0.889,0.751,0.796} 31.6 & \cellcolor[rgb]{0.887,0.749,0.794} 31.6 & \cellcolor[rgb]{0.894,0.759,0.799} \textbf{0.9} & \cellcolor[rgb]{0.962,0.901,0.888} 0.2 \\
Qwen3.5 & 9B & \cellcolor[rgb]{0.843,0.698,0.771} 43.6 & \cellcolor[rgb]{0.794,0.655,0.749} \underline{54.5} & \cellcolor[rgb]{0.881,0.741,0.791} 33.6 & \cellcolor[rgb]{0.841,0.696,0.770} \underline{43.9} & \cellcolor[rgb]{0.881,0.741,0.791} 33.9 & \cellcolor[rgb]{0.890,0.754,0.797} 30.6 & \cellcolor[rgb]{0.916,0.795,0.817} 22.3 & \cellcolor[rgb]{0.896,0.761,0.800} 28.9 & \cellcolor[rgb]{0.871,0.729,0.785} 36.3 & \cellcolor[rgb]{0.824,0.680,0.762} 48.0 & \cellcolor[rgb]{0.848,0.703,0.773} 42.7 & \cellcolor[rgb]{0.848,0.703,0.773} 42.3 & \cellcolor[rgb]{0.863,0.719,0.781} 38.4 & \cellcolor[rgb]{0.956,0.885,0.875} 0.1 & \cellcolor[rgb]{0.962,0.901,0.888} 0.0 \\
Qwen3.5 & 27B & \cellcolor[rgb]{0.839,0.694,0.769} 44.9 & \cellcolor[rgb]{0.748,0.623,0.727} \textbf{63.6} & \cellcolor[rgb]{0.837,0.691,0.768} \textbf{45.5} & \cellcolor[rgb]{0.809,0.667,0.756} \textbf{51.3} & \cellcolor[rgb]{0.832,0.687,0.766} \underline{46.4} & \cellcolor[rgb]{0.869,0.726,0.784} 37.1 & \cellcolor[rgb]{0.890,0.754,0.797} \underline{31.2} & \cellcolor[rgb]{0.865,0.722,0.782} \textbf{38.2} & \cellcolor[rgb]{0.897,0.764,0.802} 28.6 & \cellcolor[rgb]{0.771,0.639,0.738} \underline{59.3} & \cellcolor[rgb]{0.861,0.717,0.780} 39.3 & \cellcolor[rgb]{0.848,0.703,0.773} 42.4 & \cellcolor[rgb]{0.841,0.696,0.770} \textbf{44.0} & \cellcolor[rgb]{0.907,0.779,0.809} 0.8 & \cellcolor[rgb]{0.961,0.898,0.886} 0.6 \\
\bottomrule
\end{tabular}
}
\end{table}

\subsection{Audio Experiment}

\begin{figure}[H]
    \centering
    \begin{subfigure}[h]{\linewidth}
        \centering
        \includegraphics[width=\linewidth]{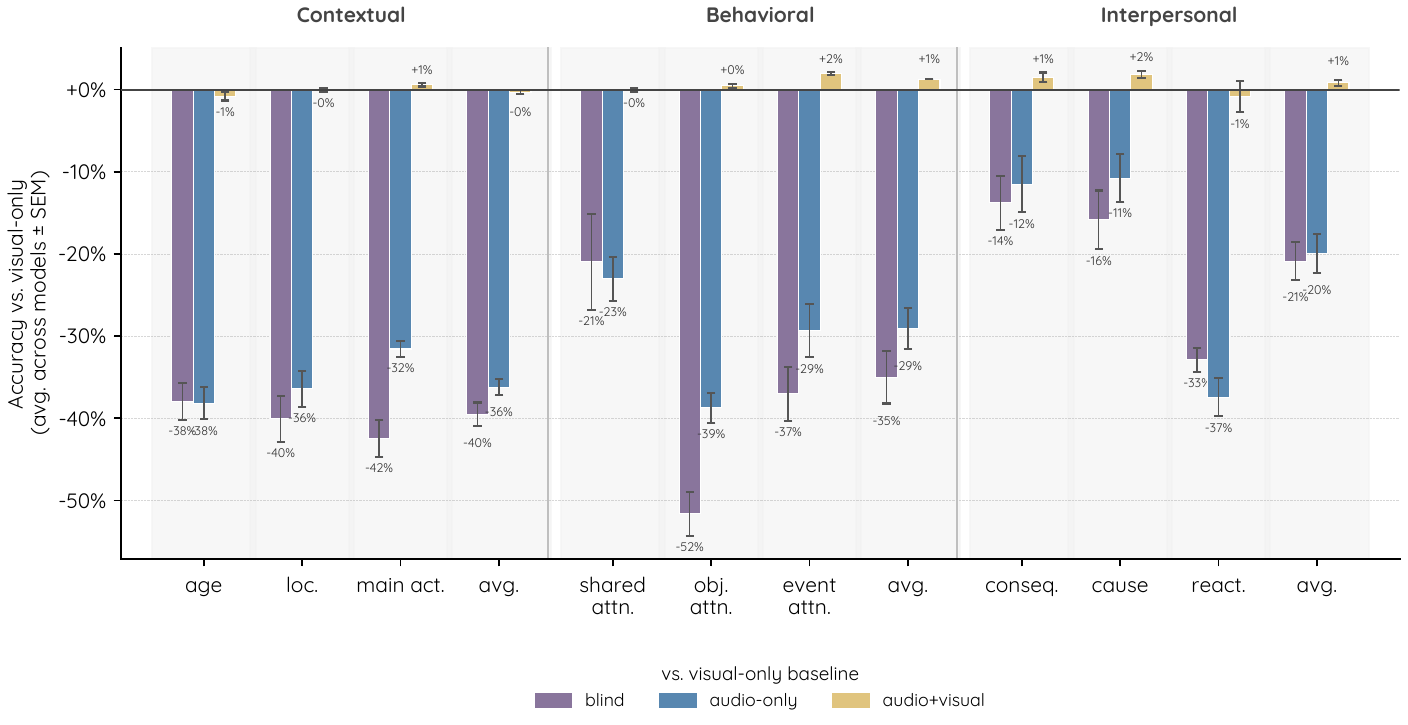}
        \caption{Accuracy gap relative to the visual-only baseline for blind, 
        audio-only, and audio+visual conditions, averaged across models ($\pm$ Standard Error of the Mean (SEM)). 
        Audio alone consistently underperforms visual input across all task 
        categories, while combining both modalities yields marginal gains.}
        \label{fig:audio_added_value}
    \end{subfigure}
    \hfill
    \begin{subfigure}[h]{\linewidth}
        \centering
        \includegraphics[width=\linewidth]{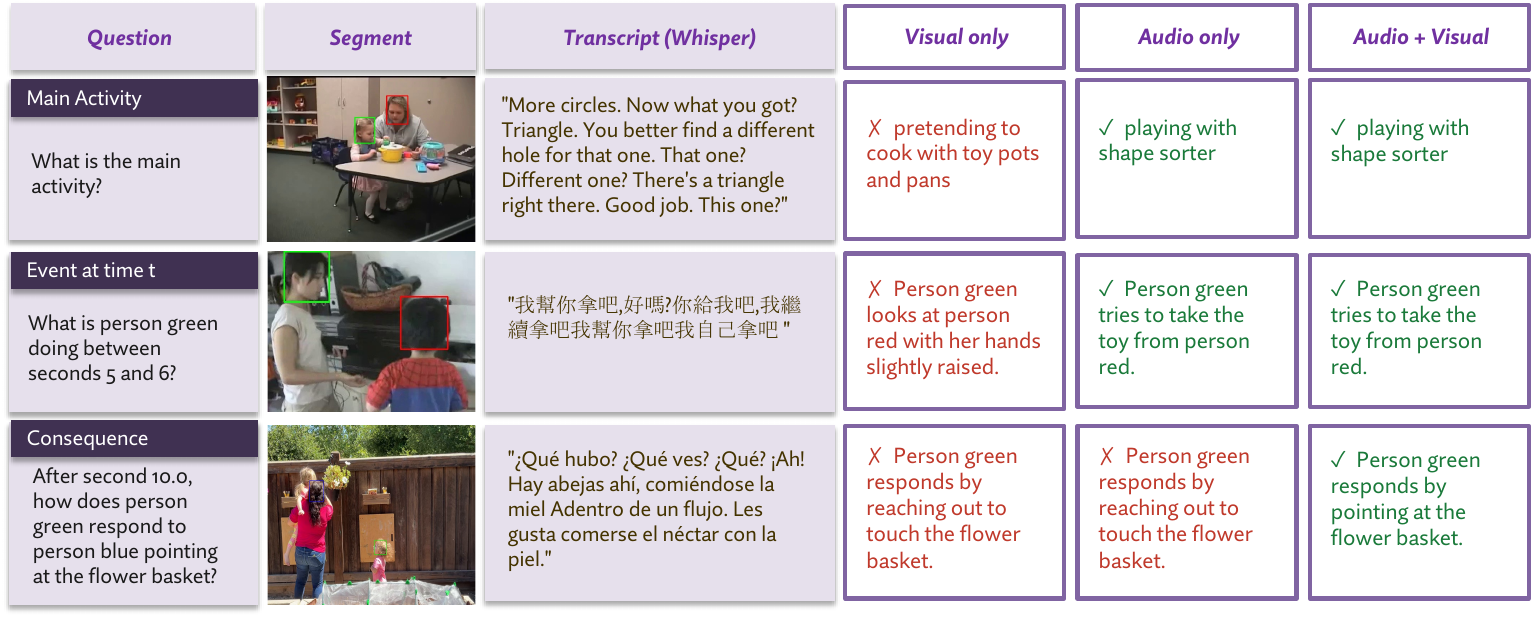}
        \caption{Qualitative examples showing cases where audio alone suffices (rows 1–2) and where combining audio and visual is necessary (row 3, consequence reasoning in Spanish).}
        \label{fig:audio_comparison}
    \end{subfigure}
    \caption{\textbf{Impact of audio on task performance.} (a) Quantitative accuracy gaps across task subtypes. (b) Qualitative examples across modality conditions.}
    \label{fig:combined}
\end{figure}

To test whether visual perception is essential for \bench or whether speech and ambient audio carry comparable signal, we evaluate models under three conditions: blind (text only), audio-only (transcribed via Whisper \citep{radford2023robust}), and audio+visual. Figure~\ref{fig:audio_added_value} reports accuracy gaps relative to the visual-only baseline. Audio alone consistently underperforms visual input across all task categories, with average drops of $-36\%$ on Contextual, $-29\%$ on Behavioral, and $-20\%$ on Interpersonal questions. Combining audio and visual yields only marginal gains over visual alone (within $±2\%$ across all subtypes), indicating that for most \bench questions the visual stream already contains the information needed to answer correctly. Qualitative examples (Figure~\ref{fig:audio_comparison}) illustrate the narrow conditions under which audio contributes meaningfully, either on its own or alongside visual input — most often when adults verbally narrate the ongoing activity. Together, these results confirm that \bench is fundamentally a visual social-reasoning benchmark.

\subsection{Reliability of LLM judge for open-ended descriptions}
\label{app:judge}
We assess sensitivity to judge choice by re-scoring all open-ended responses from four evaluated models using GPT-5, and comparing its scores with those of our primary judge, Qwen3-VL-32B. For each evaluated model, we compute Spearman correlation between the two judges' per-response scores, mean absolute difference (MAD), and the difference in mean scores ($\Delta = \text{Qwen judge} - \text{GPT-5 judge}$), separately for similarity and recall.


Table~\ref{tab:gpt_judge} shows strong overall agreement between the two judges. The relative model rankings proved stable (Spearman rank correlations = 0.60 for sim, 0.70 for recall), and the overall score deltas remained small (-0.3 for sim, 0.024 for recall), indicating that substituting the judge model preserves our findings. The results also suggest that same model bias is small, with similarity being more stringent and recall slightly higher.

\begin{table}[]
\centering
\caption{\textbf{Open question model performance by judge.} We measure inter-judge agreement for similarity and recall between Qwen3-VL-32B and GPT-5 using Spearman rank correlation, Mean Absolute Distance (MAD), and the overall delta.}
\label{tab:gpt_judge}
\resizebox{\textwidth}{!}{%
\begin{tabular}{llllllllllll}
\toprule
 &  & \multicolumn{5}{c}{\textbf{Similarity}} & \multicolumn{5}{c}{\textbf{Recall}} \\
\midrule
Model & Size (B) & Spearman & MAD & Qwen3 & GPT-5 & $\Delta$ & Spearman & MAD & Qwen3 & GPT-5 & $\Delta$ \\
\midrule
Keye-VL 1.5 & 8 & 0.6 & 0.4 & 1.3 & 1.4 & -0.1 & 0.676 & 0.104 & 0.173 & 0.151 & 0.022 \\
Qwen3-VL & 32 & 0.3 & 0.7 & 0.9 & 1.5 & -0.6 & 0.712 & 0.113 & 0.212 & 0.173 & 0.039 \\
Qwen3.5 & 27 & 0.7 & 0.4 & 1.3 & 1.5 & -0.2 & 0.727 & 0.112 & 0.219 & 0.203 & 0.016 \\
Gemini 3.1 Flash Lite & - & 0.7 & 0.4 & 1.1 & 1.4 & -0.3 & 0.683 & 0.1 & 0.171 & 0.151 & 0.02 \\
Average &  & 0.6 & 0.5 & 1.2 & 1.5 & -0.3 & 0.700 & 0.107 & 0.194 & 0.170 & 0.024 \\
\bottomrule
\end{tabular}%
}
\end{table}

\section{Additional experiments}
\label{sec:additional_exp}
\subsection{Performance stratified by initiator}
\begin{figure}[h]
    \centering
    \includegraphics[width=\linewidth]{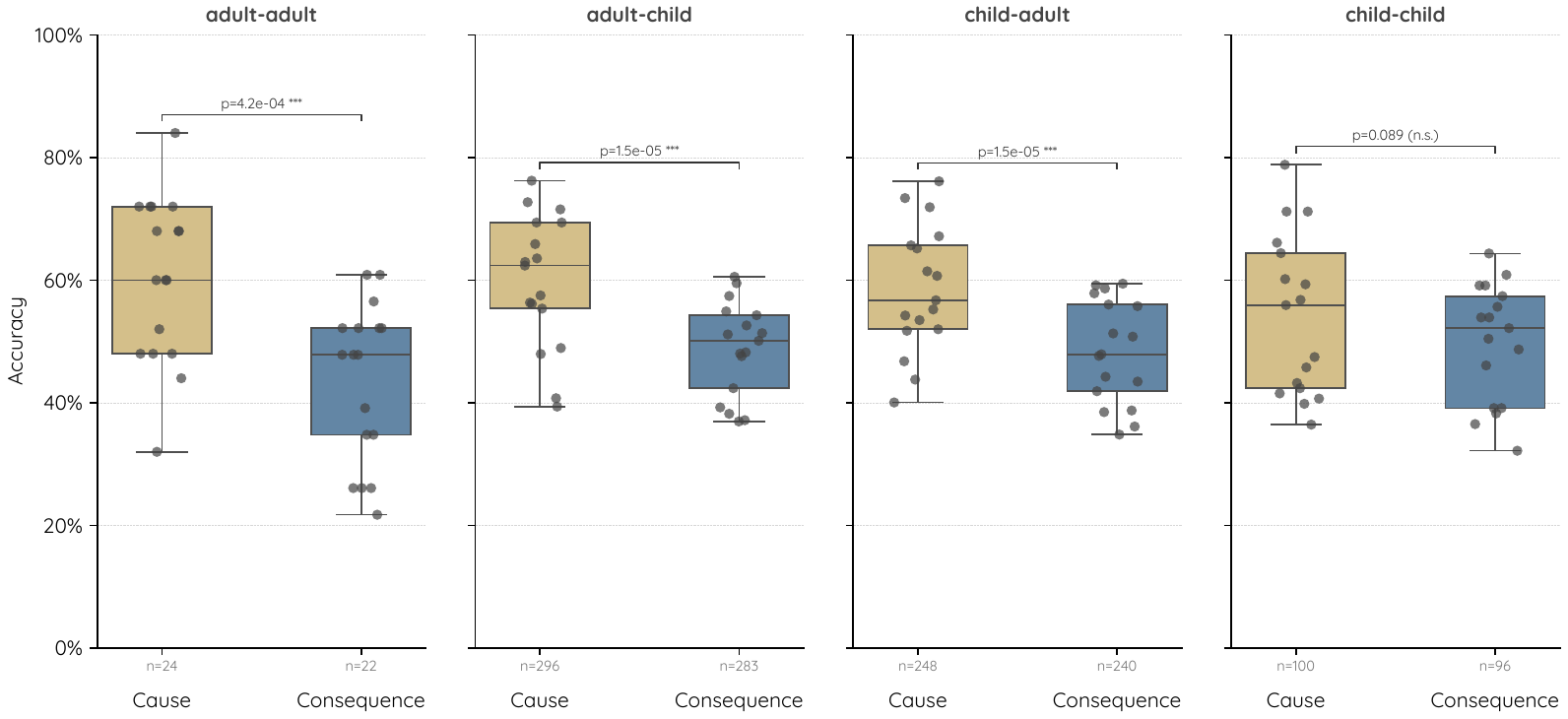}
    \caption{\textbf{Comparing performance across models by action-reaction dyad groups for cause and consequence interpersonal questions.}}
    \label{fig:cause_conseq_dyad}
\end{figure}

We further analyze model performance on Cause and Consequence questions stratified by the action-reaction dyad — i.e., whether the action–reaction pair is performed by an adult–adult, adult–child, child–adult, or child–child pair (Figure~\ref{fig:cause_conseq_dyad}). Across all four dyad groupings, models perform reliably better on Cause than on Consequence, with the gap reaching statistical significance in three of four conditions (adult–adult, adult–child, and child–adult; Wilcoxon signed-rank test, all $p<0.001$). The consistent Cause-over-Consequence gap suggests that models more reliably identify the antecedents of an observed reaction than predict the downstream effects of an observed action — a pattern consistent with the broader interpersonal reasoning weaknesses we report in the main results.

\subsection{Performance by video quality} 
\begin{figure}[h]
    \centering
    \includegraphics[width=0.5\linewidth]{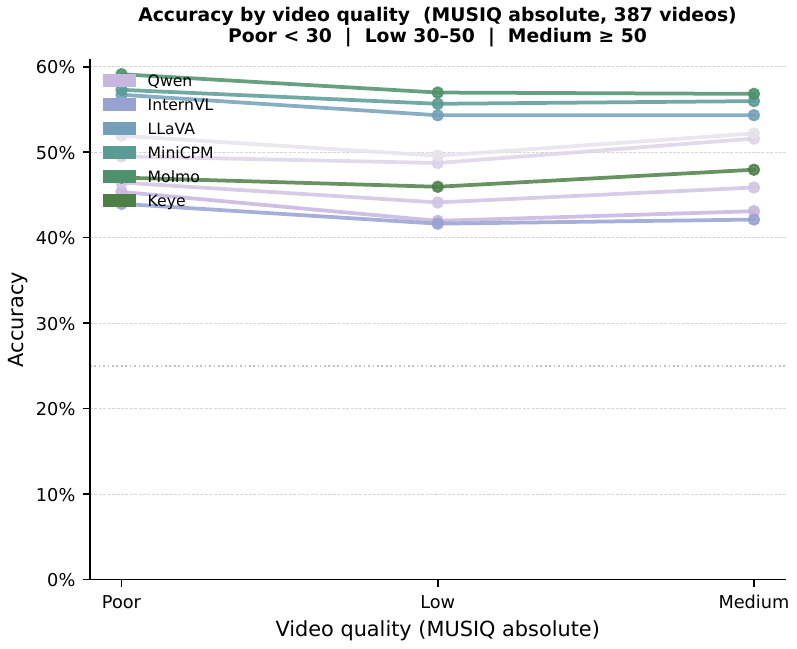}
    \caption{\textbf{Accuracy stratified by MUSIQ \cite{ke2021musiq} absolute.} A higher MUSIQ score reflects a better perceived quality. It accounts for blur, noise, compression artifacts, resolution, etc.}
    \label{fig:accuracy_quality}
\end{figure}

Because \bench aggregates videos from three corpora collected under different recording conditions, the dataset spans a wide range of perceptual quality. We assess whether this variation drives model performance using MUSIQ \citep{ke2021musiq}, a no-reference image quality metric that accounts for blur, noise, compression artifacts, and resolution. Figure~\ref{fig:accuracy_quality} shows accuracy stratified by MUSIQ quality bin (Poor $< 30$, Low 30–50, Medium $\geq$ 50). Accuracy remains relatively stable across bins for all six models, suggesting that performance limitations are not primarily driven by perceptual degradation but by challenges in structured social reasoning.


\subsection{Performance stratified by demographics}

\bench spans a broad range of participant demographics, including a wide age range of children (from infancy to adolescence) and videos collected across multiple geographic regions. To verify that model performance is not driven by, or systematically biased against, any specific demographic group, we stratify accuracy by child age and continent of recording.

\subsubsection{Performance by age}

\begin{figure}[h]
    \centering
    \includegraphics[width=0.7\linewidth]{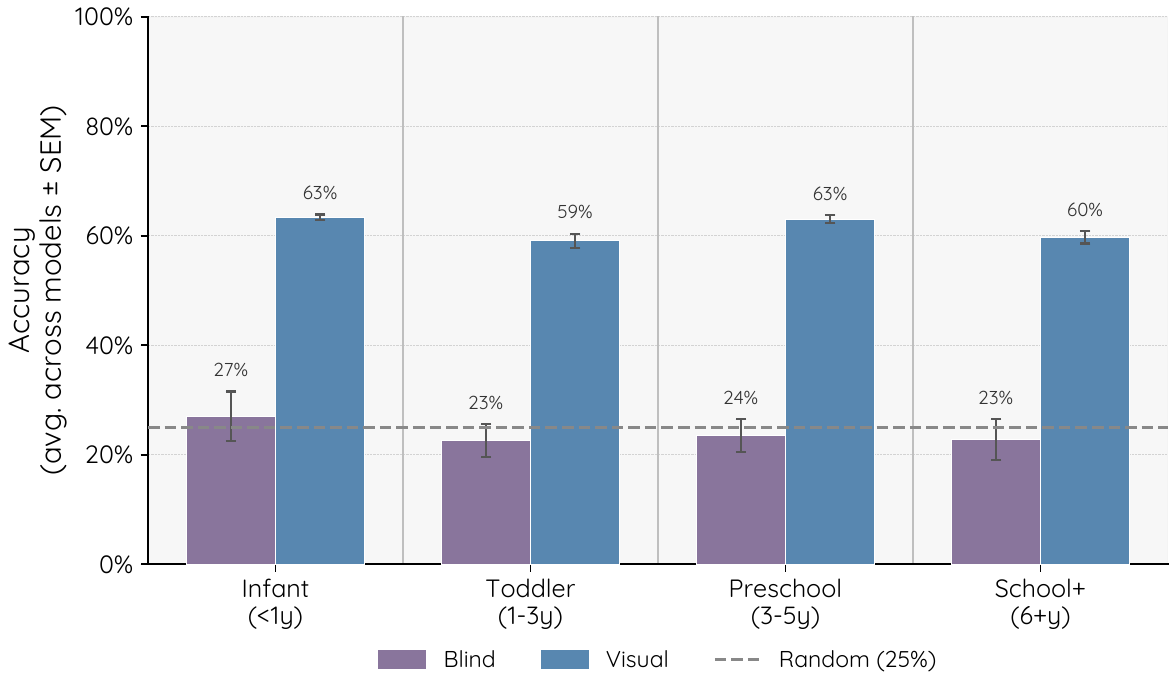}
    \caption{\textbf{Behavioral question (event at time t) accuracy by child age category.} Visual condition (blue) substantially outperforms blind (purple) and random chance (25\%) across all age groups, with no significant variation by age. Error bars show standard error of the mean (SEM) across models.}
    \label{fig:age_accuracy_event_time_t}
\end{figure}

Figure~\ref{fig:age_accuracy_event_time_t} shows model accuracy on behavioral (event-at-time-t) questions stratified by the child's age category. This indicates that model performance is broadly stable across the developmental range covered by \bench, and not concentrated in a particular age band. 

\subsubsection{Performance by continent}
\begin{figure}[h]
    \centering
    \includegraphics[width=\linewidth]{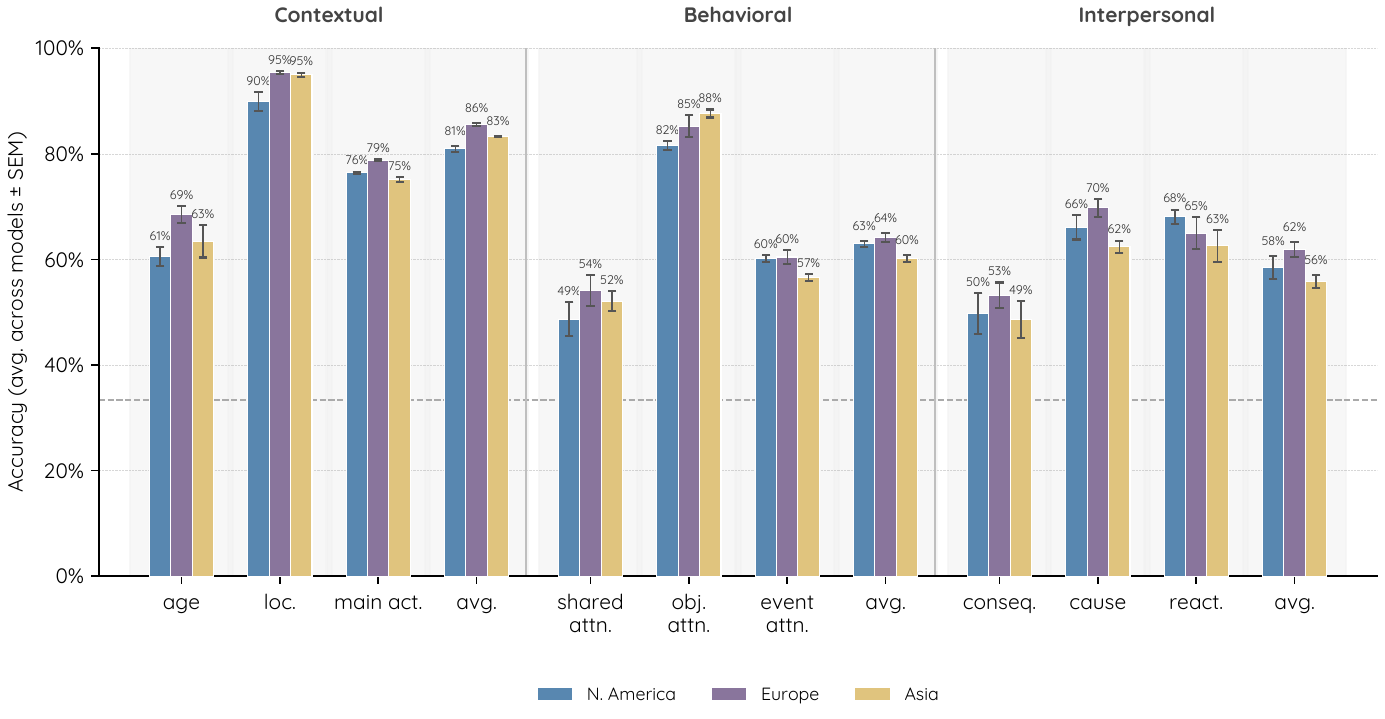}
    \caption{\textbf{Model accuracy across task subtypes and geographic regions.} Bars show average accuracy across models ($\pm$ standard error of the mean (SEM)) for North America, Europe, and Asia, grouped by task category (Contextual, Behavioral, Interpersonal). The dashed line indicates chance level.}
\label{fig:fairness_subtype_continent}
\end{figure}

Fig.~\ref{fig:fairness_subtype_continent} shows model performance across task subtypes stratified by continent. Across all three categories, geographic disparities are modest. Asia tends to show slightly lower accuracy in Interpersonal tasks (56\% vs. 58–62\% average), while North America and Europe perform comparably. 

\subsection{Performance by setting}

\begin{figure}[h]
    \centering
    \includegraphics[width=\linewidth]{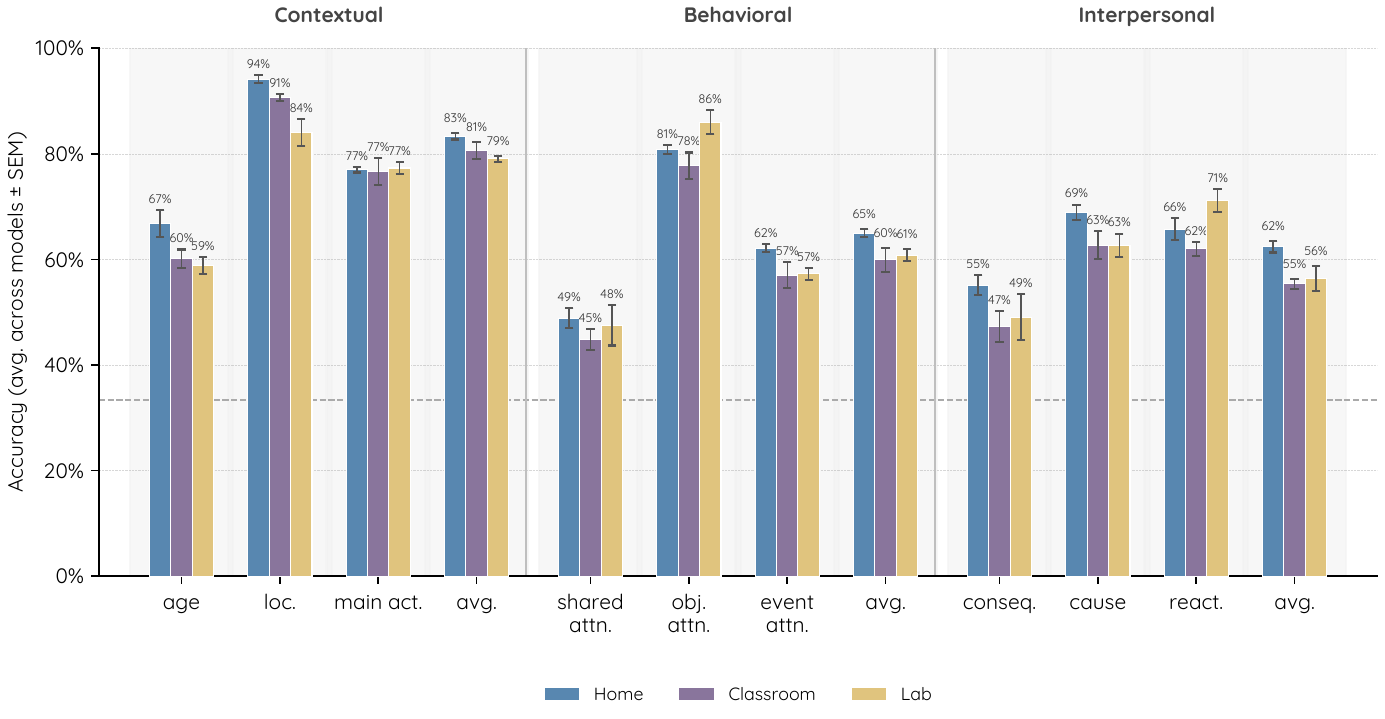}
     \caption{\textbf{Model accuracy across task subtypes and recording settings.} Bars show average accuracy across models ($\pm$SEM) for Home ($n{=}718$), Classroom ($n{=}228$), and Lab ($n{=}263$ segments), grouped by task category; the dashed line indicates chance. Setting is largely nested within data source (Lab is $237/263$ CHILDES~4) and the Classroom and Lab subsets are smaller, so differences are descriptive rather than causal.}
    \label{fig:fairness_subtype_setting}
\end{figure}

Performance is stable across homes, classrooms, and labs at the category level (largest gap: home $62\%$ vs.\ classroom $55\%$ on Interpersonal), but the ordering inverts by task type: lab segments are weakest on location ($84\%$ vs.\ $94\%$ at home) and strongest on object attention ($86\%$ vs.\ $78$--$81\%$ elsewhere) and reaction ($71\%$ vs.\ $62$--$66\%$), while main activity is unaffected ($77\%$ throughout), suggesting controlled framing trades scene context for a clearer view of the dyad. As the setting is largely nested within the data source, and the classroom ($n{=}228$) and lab ($n{=}263$) subsets are substantially smaller and drawn from fewer studies than home ($n{=}718$), we report this as a descriptive stratification rather than a causal effect of setting (Fig.~\ref{fig:fairness_subtype_setting}).

\section{Towards VLM-assisted interventions}

\subsection{Finetuning}

\label{sec:supp_finetuning}

\textbf{Setup.} We LoRA-finetune qwen3.5 at three scales (4B, 9B, 27B) on a separate pool of non-reviewed open-interaction questions, with 485 CHILDES samples for training and 137 \FIND samples for validation. We evaluate on a held-out \bench subset (n=683) with no source-video overlap with the training pool, ensuring reported gains reflect transfer to held-out scenes. Train and validation pools are split by source corpus to test cross-corpus transfer. Models are trained for 3 epochs and early-stopped at epoch 2 by validation loss. We use rank-16 LoRA ($\alpha{=}16$, no dropout) on the language layers with the vision encoder frozen, 8-bit AdamW at a learning rate of $2\times10^{-4}$ with linear decay, and an effective batch size of 4.

\textbf{Per-source results.} Table~\ref{tab:finetune_per_source} shows both the overall and per-source performance of the open interaction task before and after finetuning. The 4B and 9B models gain on every source; the 27B model regresses on similarity for the ChildPlay/CHILDES subsets, suggesting the 27B base is already close to the LoRA's reachable optimum on in-domain prompts and that more data would be needed to see consistent gains.
  
\begin{table}[h]
  \centering
  \small
  \caption{\textbf{Finetuning impact on the NEST-no-overlap open-interaction held-out set, broken out by data source.} \emph{base} is the un-finetuned base model and \emph{ft} is the finetuned model. Sim is the MMBench-Video judge score (0-3); Rec is claim-level recall (0-1).}
  \setlength{\tabcolsep}{4pt}
  \resizebox{0.8\linewidth}{!}{%
  \begin{tabular}{llrrrrrrrr}
  \toprule
  Size & Source & $n$ & Sim$_{\text{base}}$ & Sim$_{\text{ft}}$ & $\Delta$Sim & Rec$_{\text{base}}$ & Rec$_{\text{ft}}$ &
  $\Delta$Rec  \\
  \midrule
  4B  & Overall   & 683 & 1.000 & 1.113 & \dpos{+0.113} & 0.111 & 0.193 & \dpos{+0.082}  \\
      & Childplay & 316 & 1.149 & 1.253 & \dpos{+0.104} & 0.145 & 0.236 & \dpos{+0.090} \\
      & Childes   & 357 & 0.880 & 0.994 & \dpos{+0.115} & 0.081 & 0.156 & \dpos{+0.074} \\
      & \FIND      &  10 & 0.600 & 0.900 & \dpos{+0.300} & 0.073 & 0.162 & \dpos{+0.088}  \\
  \midrule
  9B  & Overall   & 683 & 1.060 & 1.211 & \dpos{+0.151} & 0.113 & 0.230 & \dpos{+0.118} \\
      & Childplay & 316 & 1.149 & 1.241 & \dpos{+0.092} & 0.130 & 0.264 & \dpos{+0.134} \\
      & Childes   & 357 & 0.986 & 1.190 & \dpos{+0.204} & 0.095 & 0.201 & \dpos{+0.106} \\
      & \FIND      &  10 & 0.900 & 1.000 & \dpos{+0.100} & 0.172 & 0.213 & \dpos{+0.042}  \\
  \midrule
  27B & Overall   & 683 & 1.262 & 1.186 & \dneg{-0.076} & 0.236 & 0.231 & \dneg{-0.005}  \\
      & Childplay & 316 & 1.342 & 1.244 & \dneg{-0.098} & 0.295 & 0.265 & \dneg{-0.030}  \\
      & childes   & 357 & 1.196 & 1.134 & \dneg{-0.062} & 0.184 & 0.199 & \dpos{+0.016} \\
      & \FIND      &  10 & 1.100 & 1.200 & \dpos{+0.100} & 0.228 & 0.260 & \dpos{+0.032}  \\
  \bottomrule
  \end{tabular}%
  }
  
  \label{tab:finetune_per_source}
  \end{table}
  
\subsection{Application: Parent-coaching videos}
\label{app:case_study}
\definecolor{prompt_bg}{gray}{0.95} 
\begin{figure*}[h]
    \centering
    \begin{subfigure}{0.48\textwidth}
        \begin{mdframed}[backgroundcolor=prompt_bg, innerleftmargin=10pt, innerrightmargin=10pt, topline=false, bottomline=false, leftline=false, rightline=false]
            \scriptsize
\begin{lstlisting}[basicstyle=\scriptsize\ttfamily, breaklines=true, columns=fullflexible]
Use the following descriptions to identify the people in the clip:
- child "{{child_ref}}": {{child_description}}
- adult "you": {{adult_description}}

Key definition: A serve is when the child initiates an interaction. A return is when the adult notices and responds.

The child's serve in this clip: {{child1}}

Describe the adult's return in one sentence written as: "You return their serve <by/with> <what you do>."

Refer to the child as "{{child_ref}}" and the adult as "you".
\end{lstlisting}
        \end{mdframed}
        \caption{Predict Return Prompt}
        \label{fig:prompt_return}
    \end{subfigure}
    \hfill
    \begin{subfigure}{0.48\textwidth}
        \begin{mdframed}[backgroundcolor=prompt_bg, innerleftmargin=10pt, innerrightmargin=10pt, topline=false, bottomline=false, leftline=false, rightline=false]
            \scriptsize
\begin{lstlisting}[basicstyle=\scriptsize\ttfamily, breaklines=true, columns=fullflexible]
Use the following descriptions to identify the people in the clip:
- child "{{child_ref}}": {{child_description}}
- adult "you": {{adult_description}}

Key definition: A serve is when the child initiates an interaction. A return is when the adult notices and responds.

The adult's return in this clip: {{adult1}}

Describe the child's serve in one sentence written as: "{{child_ref}} serves by <what they do>."

Refer to the child as "{{child_ref}}" and the adult as "you".
\end{lstlisting}
        \end{mdframed}
        \caption{Predict Serve Prompt}
        \label{fig:prompt_serve}
    \end{subfigure}
    
    \caption{\textbf{Prompts for the application of \bench on parent-coaching videos.}}
    \label{fig:app_prompts}
\end{figure*}
The parent-coaching videos for the case study are part of the \FIND program, an intervention program that aims to support parents and educators by providing feedback on how they interact with the children in their care. \FIND is based on the principle of positive feedback, where parents (and educators) are shown examples of their own developmentally supportive interactions. The examples are short ($\sim 11.6s$) clips with accompanying descriptions selected from a long-form video recorded in the caregiver-child dyad's home or other natural setting. This coaching program is designed to enhance learning and optimize engagement through the positive reinforcement. In this experiment, we build on the two main elements of the developmentally supportive interactions: the child-led serve and adult return. A child serves by doing, saying, or looking at something, while an adult returns by noticing the serve and responding. As this paired serve-return follows the same structure of \bench's action-reaction, it is a natural application setting for the benchmark.

We preprocess the 123 edited videos by extracting the interaction description and video segments. We manually review that all the clips and descriptions align. Similarly to \bench (see \ref{sec:gen_descriptions}), we produce descriptions with Qwen3-VL-32B-Instruct for grounding the identity of each main child and main adult in the scenes and manually correct if needed. We make the distinction of \textit{main} child or adult as the videos can feature siblings or other children, particularly in the classroom settings. To prompt the models for evaluation we adapt the structure of \bench's interpersonal questions with i) the people's descriptions, ii) definitions for key task specific terms (in this case \textit{child serve} and \textit{adult return}) and iii) a description of the complementary element. Figure~\ref{fig:app_prompts} shows the template for the prompts.

\section{Impact}\label{sec:impact}

\bench targets a sensitive setting--video of young children and their caregivers in naturalistic interactions--so we discuss the expected positive and negative impacts of releasing the benchmark and the associated evaluation protocol, followed by the ethical considerations that shaped its design.

\subsection{Positive impacts}\label{sec:positive_impacts}

\noindent\textbf{Reducing the manual burden of behavioral video coding.}
Research and intervention in early childhood development rely heavily on manual coding of video, a process that is slow, expensive, and a common bottleneck for both basic research and applied programs. By providing a structured, multi-dimensional benchmark we make it easier for the community to identify which off-the-shelf vision systems are appropriate for which coding sub-tasks, and where they fail. This in turn supports tools that can lower the cost of observational research and broaden the contexts in which video-based intervention is feasible.

\noindent\textbf{Improved evaluation of socially aware video models.}
Existing social-reasoning benchmarks are largely adult-centric and outcome-focused. By decomposing each interaction into contextual, behavioral, and interpersonal questions over the same segment, \bench{} surfaces \emph{at which level} a model fails. This gives developers of vision-language models a more diagnostic signal than aggregate accuracy and encourages progress on perceptual grounding and interaction-level reasoning, not only narrative inference.

\noindent\textbf{Standardization across heterogeneous corpora.}
Annotations span ChildPlay, CHILDES, and \FIND{} under a single schema, enabling like-for-like evaluation across cultural contexts, recording conditions, and interaction styles. This is unusual in the child-centric vision literature and supports more transferable conclusions about model behavior.

\subsection{Negative impacts and dual-use considerations}\label{sec:negative_impacts}

\noindent\textbf{Privacy of children depicted in source videos.}
Because \bench{} indexes video of identifiable minors, any improvement we enable in social inference also lowers the cost of unwanted analysis of similar video. We address the most acute case — the \FIND{} parent-coaching corpus — by excluding its segments from the public release entirely, and we redistribute only annotations and metadata for the remaining corpora, keeping source-video access subject to TalkBank registration and the YouTube Terms of Service. This limits, but does not eliminate, the risk that downstream users analyze child-centric video without comparable safeguards.

\noindent\textbf{Surveillance and behavioral profiling.}
Better social-reasoning models are dual-use: they can support clinical observation as easily as they can support workplace, classroom, or home surveillance. We discourage such uses through the non-commercial license on the annotations (CC BY-NC-SA 4.0) and by framing the benchmark as a diagnostic evaluation tool rather than a deployment-ready system, but we cannot enforce downstream use.

\noindent\textbf{Risk of misinterpreted model outputs in applied settings.}
Coding scales in early-childhood research feed into developmental assessments and intervention decisions. Models that achieve nontrivial accuracy on \bench{} may be tempting to plug into such pipelines, even though our results show substantial gaps in interaction-level reasoning. We therefore emphasize that \bench{} is an evaluation benchmark, not a clinical instrument, and that gains on \bench{} alone do not justify clinical or developmental use without dedicated validation.

\noindent\textbf{Demographic and cultural coverage.}
While we explicitly diversify across studies, countries, and ages, the benchmark is bounded by the demographics of the contributing corpora. Performance and failure patterns observed on \bench{} should not be assumed to generalize uniformly across populations, and downstream users should treat group-level claims with caution.

\subsection{Ethical considerations}\label{sec:ethical_considerations}

\noindent\textbf{Consent and data provenance.}
Each contributing corpus carries its own consent regime, which we respect rather than override. CHILDES  videos \citep{macwhinney2000childes} come from research studies whose consent procedures are documented by the contributing investigators and gated by TalkBank registration. ChildPlay videos \citep{tafasca2023childplay} are publicly available on YouTube; we redistribute only annotations that reference them, not the videos themselves. \FIND{} videos are governed by the parent-coaching program's consent procedures and remain entirely behind the program's access controls.

\noindent\textbf{Institutional review.}
Annotation and analysis were conducted under Stanford's IRB-approved protocol (66624). The protocol covers the recruitment, training, and supervision of human annotators, as well as the data-handling practices for the segments they reviewed.

\noindent\textbf{Annotator welfare.}
Annotators were trained, compensated above the local minimum wage, and worked from a fixed protocol with explicit guidance on which frames to attend to. Inter-annotator review was built into the workflow, both as a quality-control measure and to avoid placing the burden of ambiguous cases on a single annotator. The instructions and annotation spreadsheet are included, respectively, in the annotation-protocol subsection of this supplementary (Section~\ref{sec:annotation_protocol}) and code release.

\noindent\textbf{Mitigations carried into the release.}
The release reflects the considerations above: \FIND{} excluded; only annotations and metadata redistributed; CC BY-NC-SA 4.0 license; explicit pointers to the source corpora's access procedures; and a dataset card documenting schema, splits, license, and known limitations (Section~\ref{sec:limitations}). We expect downstream users to honor the same constraints when building on \bench{}.

\subsection{Limitations}\label{sec:limitations}
We discuss the main limitations of \bench{} along five axes, spanning data release, evaluation protocol, the role of LLMs in benchmark construction and scoring, and the heterogeneity of the underlying video material.

\noindent\textbf{Restricted release of the \FIND{} subset.}
Segments sourced from the \FIND{} parent-coaching intervention corpus identify minors and are governed by the program's consent procedures. We therefore exclude these clips entirely from the public release; the HuggingFace distribution drops 156 segments and the corresponding 1{,}828 \bench test questions. As a consequence, results reported on the \FIND{} case study (Section~\ref{sec:experiments}) cannot be reproduced from the public release alone. Researchers wishing to reproduce or extend those results must obtain access through the original program.

\noindent\textbf{Standardized evaluation context.}
To enable a fair comparison across models with very different native context windows, we cap the inference context at 32{,}768 tokens and 512$\times$512 maximum pixels per frame for all models. This standardization may not reflect each model's optimal configuration: higher-capacity models with larger native context windows or higher input resolutions may achieve better performance under their default settings, and conversely some smaller models may have been tuned for shorter contexts. Reported numbers should therefore be interpreted as performance under a shared, controlled evaluation budget rather than as an absolute upper bound on each model's capability.

\noindent\textbf{LLM-in-the-loop benchmark construction and scoring.}
NEST uses Qwen3-VL-32B-Instruct to generate multiple-choice distractors through AutoConverter~\citep{zhang2025automated} and Qwen3-VL-32B~\citep{bai2025qwen3} as the primary judge for open-ended responses. Because some evaluated models belong to the same family, these choices may introduce stylistic or family-specific biases. We manually reviewed a subset of the generated questions and examined distractor selection on 896 consequence questions containing one expert-written and two generated distractors (App.~\ref{app:distractor_validity}). Across 17 models, generated distractors attracted 77.9\% of incorrect responses, averaging 39.0\% per option compared with 22.1\% for the expert-written option. Qwen-family models showed no descriptive tendency to avoid generated distractors. These results support the generated options as challenging alternatives within this subset, although residual stylistic or family-specific biases may still exist. The analysis covers only consequence questions, does not compare all-human and all-generated question sets, and lacks rejection and revision counts from manual review. We additionally assess scoring reliability through a second judge and expert validation, as discussed below.


\noindent\textbf{Open-ended evaluation metrics.}
Our similarity and recall metrics measure semantic alignment with reference descriptions and coverage of reference claims, respectively. They capture complementary aspects of response quality, but they share known limitations of automated open-ended scoring: insensitivity to subtle factual errors, reward for surface-level fluency, and limited ability to penalize plausible but unsupported descriptions. We assess their reliability through cross-judge comparison and expert validation. Re-scoring responses from four models with GPT-5 reveals judge-dependent scores, including a 0.6-point similarity shift for one model on the 0--3 scale (App.~\ref{app:judge}). This comparison does not establish invariant model rankings and excludes GPT-5 mini, the highest-scoring model on open interactions. Human validation covers 62 responses from Qwen3.5-27B and GPT-5 mini (App.~\ref{app:human_metric_validation}). For Qwen3.5-27B, claim-level false-positive and false-negative rates are 3.8\% and 46.7\%, respectively, suggesting that automated recall can miss coverage recognized by humans. These checks also do not establish the completeness or correctness of the automatic reference decomposition. We therefore report both metrics and recommend interpreting open-interaction results as comparative rather than absolute.


\noindent\textbf{Sensitivity to source video quality.}
Source clips vary in resolution, lighting, compression artifacts, and camera stability, particularly across the three contributing corpora. We probe this directly by stratifying accuracy by MUSIQ~\cite{ke2021musiq} quality bins (Fig.~\ref{fig:accuracy_quality}) and observe that performance is largely stable across bins, suggesting that current bottlenecks are reasoning-related rather than perceptual. We nonetheless flag this as a limitation: MUSIQ is a single perceptual-quality proxy, and other axes of acquisition variability (camera angle, motion blur during fast interactions, partial occlusion of the focus group) may affect specific question types in ways our aggregate analysis does not surface.



\end{document}